\documentclass[12pt]{article}
\usepackage{amsmath}
\usepackage{graphicx}
\usepackage{natbib}
\usepackage{url} 

\usepackage{latexsym, epsfig, amssymb, amsthm, mathrsfs, bbm, enumitem}
\usepackage[OT1]{fontenc}
\usepackage{xcolor}
\usepackage[colorlinks,citecolor=blue,linkcolor=blue,urlcolor=blue]{hyperref}
\usepackage{multirow,multicol,booktabs}

\usepackage[ruled]{algorithm2e}
\usepackage{tikz}
\usepackage{float}
\usepackage{booktabs, longtable, caption, tabularx, tabulary}
\usepackage{adjustbox}

\newcommand{\e}{\mathbbm{E}}

\newcommand{\indep}{\perp \!\!\! \perp}

\DeclareMathOperator{\Var}{Var}

\newtheorem{assumption}{Condition}
\newtheorem{lemma}{Lemma}
\newtheorem{theorem}{Theorem}
\newtheorem{corollary}{Corollary}
\newtheorem{setting}{Setting}
\renewcommand{\theequation}{
	\arabic{equation}%
}

\newcommand{\ignore}[1]{}{}

\newcommand{\bx}{\mbox{\bf x}}

\newcommand{\bz}{\mbox{\bf z}}

\newcommand{\bX}{\mbox{\bf X}}

\newcommand{\bZ}{\mbox{\bf Z}}

\newcommand{\bzero}{\mbox{\bf 0}}

\newcommand{\bxi}{\mbox{\boldmath $\xi$}}

\newcommand{\var}{\mathrm{var}}

\newcommand{\tr}{ {\mathrm{tr}} }

\newcommand{\supp}{\mathrm{supp}}

\def\toD{\overset{\mathscr{D}}{\longrightarrow}}

\newcommand{\Expected}{\mathbb{E}}
\newcommand{\Char}{\mathbbm{1}}

\newcommand\independent{\protect\mathpalette{\protect\independenT}{\perp}}
\def\independenT#1#2{\mathrel{\setbox0\hbox{$#1#2$}%
		\copy0\kern-\wd0\mkern4mu\box0}}

\newcommand{\non}{\nonumber\\}
\renewcommand{\ldots}{\cdots}
\renewcommand{\e}{\mathbb{E}}
\renewcommand{\hat}{\widehat}

\usepackage{xcolor}
\usepackage[draft,inline,nomargin,index]{fixme}
\fxsetup{theme=color,mode=multiuser}
\FXRegisterAuthor{yf}{ayf}{\color{magenta}}

\usepackage{ulem}
\newcommand*{\red}{\textcolor{red}}

\begin{document}
	
	\def\spacingset#1{\renewcommand{\baselinestretch}%
		{#1}\small\normalsize} \spacingset{1}

	\title{Optimal Nonparametric Inference with Two-Scale Distributional Nearest Neighbors%
	\thanks{
	This work was partially supported by NIH Grant 1R01GM131407. 
	Send correspondence to Yingying Fan (\textit{fanyingy@usc.edu}) or Lan Gao (\textit{lgao13@utk.edu}).}
\date{June 16, 2022}
\author{Emre Demirkaya$^1$, Yingying Fan$^2$, Lan Gao$^{1,2}$, Jinchi Lv$^2$,\\ Patrick Vossler$^2$ and Jingbo Wang$^2$
	\medskip\\
	University of Tennessee Knoxville$^1$ and University of Southern California$^2$
	\\
} %
	}
	
\maketitle
	
\begin{abstract} 
The weighted nearest neighbors (WNN) estimator has been popularly used as a flexible and easy-to-implement nonparametric tool for mean regression estimation. The bagging technique is an elegant way to form WNN estimators with weights automatically generated to the nearest neighbors \citep{steele2009exact,biau2010rate}; we name the resulting estimator as the distributional nearest neighbors (DNN) for easy reference. Yet, there is a lack of distributional results for such estimator, limiting its application to statistical inference. Moreover, when the mean regression function has higher-order smoothness, DNN does not achieve the optimal nonparametric convergence rate, mainly because of the bias issue. In this work, we provide an in-depth technical analysis of the DNN, based on which we suggest a bias reduction approach for the DNN estimator by linearly combining two DNN estimators with different subsampling scales, resulting in the novel two-scale DNN (TDNN) estimator. The two-scale DNN estimator has an equivalent representation of WNN with weights admitting explicit forms and some being negative. We prove that, thanks to the use of negative weights, the two-scale DNN estimator enjoys the optimal nonparametric rate of convergence in estimating the regression function under the fourth-order smoothness condition. We further go beyond estimation and establish that the DNN and two-scale DNN are both asymptotically normal as the subsampling scales and sample size diverge to infinity. For the practical implementation, we also provide variance estimators and a distribution estimator using the jackknife and bootstrap techniques for the two-scale DNN. These estimators can be exploited for constructing valid confidence intervals for nonparametric inference of the regression function. The theoretical results and appealing finite-sample performance of the suggested two-scale DNN method are illustrated with several simulation examples and a real data application.  
\end{abstract}
	
	

\section{Introduction} \label{Sec1}
	
\spacingset{1.5} 
{\color{black}			
	
Nonparametric regression analysis is a popular and flexible statistical tool with broad applications in various scientific fields. 
Among the existing nonparametric regression methods, the $k$-nearest neighbors ($k$-NN) procedure and its extensions including the weighted nearest neighbors method, have received great popularity due to their straightforward implementation and appealing theoretical properties. For existing results and some recent developments along this direction,  see,  for example, \cite{Mack_Local_1980, Gyrfi_A_2002, biau2015lectures, Berrett_Efficient_2018, lin2021estimation}.  
 
Despite the advantage of the weighted nearest neighbors (WNN) method over the unweighted $k$-NN, selection of the adaptive weights can be challenging in implementation. To address such issue, the bagged 1-NN estimator, an ensemble learning method, has been proposed. 
Specifically, \cite{steele2009exact} and \cite{biau2010rate} proposed to estimate the mean regression function by averaging all 1-NN estimators constructed from randomly subsampling $s$ observations with or without replacement, where $s$ is required to diverge with the total sample size $n$. 
\cite{steele2009exact} showed that this procedure automatically assigns monotonic nonnegative weights to the nearest neighbors in a distributional fashion on the entire sample,  motivating us to name it as the distributional nearest neighbors (DNN) in our paper for easy presentation. The bagging technique was pioneered by the seminal work of \cite{breiman1996bagging} and has been employed to improve performance of the base estimators. For instance, see \cite{hall2005bagging} for the asymptotic properties of bagged nearest neighbor classifiers. 

\cite{biau2010rate} proved the nice results that DNN achieves the nonparametric minimax optimal convergence rate under the Lipschitz continuity assumption of the regression function. Yet, there is a lack of asymptotic distribution results for such estimator, limiting its application to statistical inference. In addition, when the mean regression function has higher order smoothness, the DNN estimator no longer achieves the {nonparametric} optimal rate. In this work, we discover through thorough investigations that the non-optimality is caused by the slow convergence rate due to the bias.  For further bias reduction, we establish the higher-order asymptotic expansion for the bias of DNN. Based on such a bias expansion, we propose to eliminate the leading order bias of DNN by linearly combining two DNN estimators with different subsampling scales, resulting in the novel two-scale DNN procedure for nonparametric estimation and inference.
 
The DNN estimator has a representation of L-statistic with weights depending only on the rank of the observations \citep{steele2009exact}, facilitating easy and fast implementation. However, such a representation does not help with establishing the sampling properties.  For the theoretical analysis, we further demonstrate that DNN estimator has an equivalent representation of U-statistic with a kernel function of diverging dimensionality equal to the subsampling scale $s$, and therefore, the two-scale DNN estimator also has a U-statistic representation with a new and carefully constructed  diverging-dimensional kernel. Despite the nice U-statistic representations, the classical theory does not apply to DNN or two-scale DNN for deriving their asymptotic properties  because of the diverging dimensionality of the kernel functions.  To overcome such a technical challenge,  we exploit Hoeffding's canonical decomposition introduced in \cite{Hoeffding_A_1948},   and carefully collect and analyze the higher-order terms in our decomposition. Our theoretical results suggest that, when the subsampling scales are appropriately chosen, two-scale DNN achieves the nonparametric optimal rate under the fourth-order smoothness assumption on the regression function 
and the density function of covariates. A larger implication of our study is that, for regression function with even higher-order smoothness, the multi-scale DNN can be constructed in the same fashion to achieve the optimal nonparametric convergence rate; we leave the detailed investigation for future study.  

By construction, some weights in the two-scale DNN take negative values. The advantage of using negative weights in the weighted nearest neighbors classifiers was formally investigated in \cite{Samworth_Optimal_2012}. For the problem of regression, although \cite{biau2015lectures}  theoretically showed that the weighted nearest neighbors estimator allowing for negative weights can improve upon that with only nonnegative weights in terms of the rate of convergence, it still remains largely unclear how to practically choose these weights. Our two-scale DNN 
provides an explicit and easy-to-implement way to assign negative weights which endorses the optimal nonparametric convergence rate under the higher-order smoothness assumption of the regression function.

We further show that DNN and two-scale DNN are asymptotically normal as the subsampling scales and sample size $n$ diverge to infinity. The asymptotic variance of the two-scale DNN estimator, however, does not admit a simple analytic form that is practically useful for statistical inference.  We exploit two methods, the jackknife and bootstrap, for asymptotic variance estimation. We formally demonstrate that both methods yield consistent estimates of the asymptotic variance.  Our proofs are more intricate than the standard technique in the literature  because of the diverging subsampling scales.  The key is to write the jackknife estimator as a weighted summation of a sequence of U-statistics and carefully analyze the higher-order terms.  Our proof for the bootstrap estimator is built on our results for the jackknife estimator.  Although both methods yield consistent variance estimates, the bootstrap estimator is much more computationally efficient.  We also provide a bootstrap method to directly estimate the distribution of the two-scale DNN estimator without estimating the asymptotic variance.
	
We then demonstrate the superior finite-sample performance of our method using simulation studies and a real data application.  
The two-scale DNN estimator has two parameters to tune -- the two subsampling scales -- and it is equivalent to tune the ratio between the two subsampling scales and one of the subsampling scales. We propose to jointly tune these two parameters using a two-dimensional grid, and choose the combination of the two parameters that minimizes the mean-squared estimation error (MSE). As an application, we discuss the usage of the two-scale DNN for the heterogeneous treatment effect (HTE) estimation and inference with theoretical guarantee under the setting of randomized experiments in Section {\color{blue} B} in the Supplementary Material.
	
The rest of the paper is organized as follows. Section \ref{sec:intro} introduces the model setting for nonparametric regression estimation and reviews the DNN estimator. We present the two-scale distributional nearest neighbors (TDNN) procedure and its sampling properties in Section \ref{Sec3}.   Section \ref{Sec4} investigates the variance estimation for the TDNN estimator.  We provide several simulation examples and a real data 
application justifying our theoretical results and illustrating the finite-sample performance of the suggested TDNN method in Sections \ref{sec:SimulationStudies} and \ref{sec:RealDataAnalysis}, respectively.  Section \ref{Sec7} discusses some implications and extensions of our work. In the Supplementary Material, we also provide a bootstrap estimator for the distribution of TDNN estimator, the application of TDNN in HTE estimation and inference, and all the proofs and technical details.
}

\section{Model Setting} \label{sec:intro}
Consider a sample of independent and identically distributed (i.i.d.) observations $\{(\bX_i, Y_i)\}_{i = 1}^n$ from the following nonparametric model
\begin{equation} \label{eq: model}
	Y= \mu(\bX) + \epsilon,
\end{equation} 
where $Y$ is the response, $\bX \in \mathbb{R}^d$ represents the vector of covariates with fixed dimensionality $d$, $\mu(\bX)$ is the unknown mean regression function, and $\epsilon$ is the model error. The goal is to estimate and infer the underlying true mean regression function $\mu(\bx)$ at some given feature vector $\bx$ in the support of $\bX$.

\subsection{Distributional nearest neighbors (DNN)} \label{Sec2.1}

Given a fixed feature vector $\bx \in \mathbb{R}^d$, we calculate the Euclidean distance of each observed feature vector $\bX_i$ to the target $\bx$ and then reorder the sample according to such distances. Denote the reordered sample as $\{(\bX_{(1)}, Y_{(1)}), \cdots, (\bX_{(n)}, Y_{(n)})\}$ with 
\begin{equation} \label{neweq.FL113}
	\|\bX_{(1)} - \bx\| \le \|\bX_{(2)} - \bx\| \le \cdots \le \|\bX_{(n)} - \bx\|,
\end{equation}
where $\| \cdot \|$ denotes the Euclidean norm of a given vector and the ties are broken by assigning the smallest rank to the observation with the smallest natural index.
Then the weighted nearest neighbors (WNN) estimate \citep{Mack_Local_1980} is defined as
\begin{align} \label{eq: wnn}
	\hat \mu_\text{WNN} (\bx)  = \sum_{i=1}^n w_{ni}Y_{(i)},
\end{align}
where  $(w_{n1}, w_{n2},\cdots,w_{nn})$ is some deterministic weight vector with all the components summing up to one. In practice, one can also use the non-Euclidean distances given by certain manifold structures.

The theoretical properties of the WNN estimator  \eqref{eq: wnn} have been studied extensively in \cite{biau2015lectures}. In particular, it has been proved therein that,  with an appropriately selected 
nonnegative 
weight vector,  $\hat \mu_\text{WNN} (\bx)$  can be consistent with the optimal rate of convergence 
$O_P(n^{-2/(d+4)})$ 
when the second-order derivative exists and can have asymptotic normality.  
Moreover, the optimal rate of convergence can be improved by allowing for negative weights under higher-order derivatives. 
These existing results provide only some general sufficient conditions on the weight vector $(w_{n1},\cdots, w_{nn})$ in order to deliver the theoretical properties. However, identifying a practical weight vector with provably appealing properties can be highly nontrivial. Furthermore, the asymptotic variance of $\hat \mu_\text{WNN} (\bx)$ can admit a rather complicated form and depend upon some \textit{unknown} population quantities that are very difficult to estimate in practice, hindering the applicability in statistical inference.  

In contrast, the bagged 1-NN estimator proposed and studied in \cite{steele2009exact} and \cite{biau2010rate} (which we refer to as the DNN estimator in this paper for the ease of presentation) automatically assigns monotonic weights to the nearest neighbors in a distributional fashion on the entire sample. 
 Denote by $s$ with $1 \leq s \leq n$ the subsampling scale. Let $\{i_1,\cdots, i_s\}$ with $i_1< i_2 <\cdots < i_s$ be a random subset of the full sample $\{1,\cdots, n\}$. Hereafter, we use $\bZ_{i}$ as a shorthand notation for $(\bX_{i}, Y_{i})$ with $1 \leq i \leq n$. Let us define $\Phi(\bx; \bZ_{i_1}, \bZ_{i_2}, \ldots, \bZ_{i_s})$ as the 1-NN estimator 
\begin{equation} \label{neweq.FL002}
	\Phi(\bx; \bZ_{i_1}, \bZ_{i_2}, \ldots, \bZ_{i_s}) = Y_{(1)}(\bZ_{i_1}, \bZ_{i_2}, \ldots, \bZ_{i_s})
\end{equation}
for estimating the true value $\mu(\bx)$ of the underlying mean function at the fixed point $\bx$ based on the given subsample $\{\bZ_{i_1}, \cdots, \bZ_{i_s}\}$. Then the DNN estimator $D_n(s)(\bx)$ with subsampling scale $s$ for estimating $\mu(\bx)$ is formally defined as a U-statistic
\begin{equation} \label{neweq.FL001}
	D_n(s)(\bx) = \binom{n}{s}^{-1} \sum _{1 \le i_1 < i_2 < \ldots < i_s \le n} \Phi(\bx; \bZ_{i_1}, \bZ_{i_2}, \ldots, \bZ_{i_s}),
\end{equation}
where the kernel function $\Phi(\bx; \cdot)$ is given in (\ref{neweq.FL002}).

The above U-statistic representation averages over all 1-NN estimators given by all possible subsamples of size $s$.  For the case of $s = 1$, the DNN estimator reduces to the simple sample average $n^{-1}\sum_{i=1}^n Y_{i}$, which admits reduced variance but inflated bias. In contrast, for the case of $s = n$, the DNN estimator reduces to the simple 1-NN estimator $Y_{(1)}$ based on the full sample of size $n$, which admits the lowest bias but inflated variance. See, e.g., \cite{Hoeffding_A_1948, Hjek_Asymptotic_1968, Korolyuk_Theory_1994} for the classical asymptotic theory of the U-statistics.  Since the computation of general U-statistics becomes more challenging when sample size $n$ grows, the following lemma in \cite{steele2009exact} shows that a different representation of the DNN estimator can be exploited for easy computation. 

\begin{lemma}[\cite{steele2009exact}]
	\label{le-L}
The DNN estimator $D_n (s) ({\bf x }) $ also admits an equivalent L-statistic \citep{Serfling_Approximation_1980}  representation as 
	\begin{equation}
		\label{eqn:dnn-lstat}
		D_n (s) ({\bf x })  =  {n \choose s}^{ - 1 }  \sum_{i = 1}^{n - s + 1} {n - i \choose s - 1} Y_{(i)},
	\end{equation}	 
	where $Y_{(i)}$'s are given by the full sample of size $n$.
\end{lemma}

One nice property of the above DNN estimator is that the distribution of weights is characterized by only two parameters of the full sample size $n$ and the subsampling scale $s$. As shown later in Section \ref{Sec3.3}, our new higher-order asymptotic expansion for the bias reveals that the distributional weights in the DNN yield the explicit constant for the leading bias term that is free of the subsampling scale $s$, which opens the door for eliminating the first-order asymptotic bias of the DNN.

\section{Two-scale distributional nearest neighbors} \label{Sec3}


\subsection{Two-scale DNN} \label{Sec3.2}

We are now ready to suggest a natural extension of the single-scale DNN procedure introduced in Section \ref{Sec2.1}. The major motivation for this extension comes from the precise higher-order asymptotic bias expansion for the single-scale DNN estimator $D_n(s) (\bx)$ unveiled in Theorem \ref{thm: bias} to be presented in Section \ref{Sec3.3}. In particular, we see that the explicit constant for the leading order term in the asymptotic expansion for the bias $B(s) = \mathbb{E} \, D_n(s) (\bx) - \mu(\bx)$ is independent of the subsampling scale $s$. Such an appealing property gives us an effective way to completely remove  the first-order asymptotic bias in the order of $s^{-2/d}$, making only the second-order asymptotic bias dominating at the finite-sample level. 

To achieve the aforementioned goal, let us consider a pair of single-scale DNN estimators $D_n(s_1)(\bx)$ and $D_n(s_2)(\bx)$ with different subsampling scales $1 \leq s_1 < s_2 \leq n$ as constructed in (\ref{neweq.FL001}). Then Theorem \ref{thm: bias} ensures that 
\begin{align} \label{neweq.FL106}
	\mathbb{E} \, D_n(s_1)(\bx) & = \mu(\bx) + c \, s_1^{-2/d} + R(s_1), \\
	\mathbb{E} \, D_n(s_2)(\bx) & = \mu(\bx) + c \, s_2^{-2/d} + R(s_2), \label{neweq.FL107}
\end{align}
where $c$ is some positive constant depending on the underlying distributions, but not on the subsampling scale parameter $s_1$ or $s_2  $, and the higher-order remainder is given by {\color{black} $ R (s) = O (s^{ - 3})$ for $  d = 1$ and $ R(s) = O(s^{ - 4/d}) $ for $ d \geq 2 $ }.

Although the specific constant $c$ in the asymptotic expansions (\ref{neweq.FL106}) and (\ref{neweq.FL107}) above is unknown to us, we can proceed with solving the following system of linear equations with respect to $w_1$ and $w_2$
$$
w_1 + w_2   = 1,  \qquad w_1 \; s_1^{-2/d} +  w_2 \; s_2^{-2/d}   = 0,
$$
whose solutions are given by the specific weights 
	\begin{align} \label{wei1}
		w_1^*& = w_1^*(s_1,s_2) = 1/(1-(s_1/s_2)^{-2/d}) \\ 
		\mbox{and} \quad 
		\label{wei2}
		w_2^*&=  w_2^*(s_1,s_2) = -(s_1/s_2)^{-2/d}/(1-(s_1/s_2)^{-2/d}).
	\end{align}
Then our two-scale distributional nearest neighbors (TDNN) estimator $D_n(s_1, s_2)(\bx)$ is formally defined as 
\begin{equation} \label{TDNN}
	D_n(s_1, s_2)(\bx) = w_1^*D_n(s_1)(\bx) + w_2^*D_n(s_2)(\bx). 
\end{equation}
We will impose the restriction that $s_1/s_2$ is bounded away from both 0 and 1 by some positive constants.   This can avoid the undesirable cases of weights being too close to 0 or having diverging magnitude as $s_1$ and $s_2$ diverge.  

Since the specific weights $w_1^*$ and $w_2^*$ depend only on subsampling scales $s_1$ and $s_2$,  we see from the asymptotic expansions (\ref{neweq.FL106}) and (\ref{neweq.FL107}) that 
\begin{equation} \label{neweq.FL110}
	\mathbb{E} \, D_n(s_1, s_2)(\bx) = \mu(\bx) + R^* ( s_1),
\end{equation}
where $ R^* (s_1) = O (s_1^{ - 4/ d}) $ for $ d  \geq 2$, $ R^* (s_1) = O (s_1^{ - 3})  $ for $ d = 1$, and we impose the constraint that $s_1\sim s_2$ with $\sim$ representing asymptotic equivalence.  
The removal of the first-order asymptotic bias as shown in (\ref{neweq.FL110}) provides the TDNN estimator appealing finite-sample performance with reduced bias and controlled variance, as demonstrated with extensive simulation examples in Section \ref{sec:SimulationStudies}.

It is worth mentioning that in view of (\ref{wei1}) and (\ref{wei2}),  weight $w_1^*$ is negative given $s_1 < s_2$. This implies that the two-scale DNN can assign negative weights to some distant nearest neighbors. In fact, the advantage of using negative weights in the $k$-NN classifier for the classification setting was discovered earlier in \cite{Samworth_Optimal_2012}.  See also the theoretical discussions in \cite{biau2015lectures} for similar advantages in the regression setting. 


{\color{black}
	 TDNN is a bias-corrected version of DNN. 
	 Bias reduction techniques have been commonly used in the literature for improved mean-squared error. For example, \cite{hall1992effect} and \cite{schucany1977improvement} discussed bias correction using the bootstrap estimator and jackknife estimator, respectively. Other works have specialized bias correction in different models; for instance, see \cite{calonico2018effect} and \cite{newey2004twicing} in kernel density estimation,  \cite{cheang2000bias} in time series models, and \cite{leblanc2010bias} in nonparametric density estimation based on the Bernstein polynomial approximations. In such an endeavor, one always needs to estimate the bias term or find its theoretical representation. The form of the bias depends on the estimator in question, and so the bias reduction techniques can differ. For example, \cite{calonico2018effect} expressed the bias of kernel density estimator in terms of the bandwidth using the Edgeworth expansion in \cite{hall2013bootstrap}. \cite{schucany1971bias} established a general bias reduction method using the ratio of bias terms of two different estimators when those estimators have different but known bias terms. In our work, Theorem \ref{thm: bias} enables us to decompose the bias of the DNN estimator in terms of the subsampling scale. By combining two DNN estimators with different subsampling scales, we are able to eliminate the first-order bias. 
}

\subsection{Accuracy and asymptotic distributions of two-scale DNN} \label{Sec3.3}

We now turn to deriving the higher-order asymptotic expansions of the DNN and TDNN estimators and their asymptotic distributions. To this end, we need to impose some necessary assumptions, which are commonly used for nonparametric regression, to facilitate our technical analysis. 

Assume that 
the distribution of $\bX$ has a density function $f(\cdot)$ with respect to the Lebesgue measure $\lambda$ on the Euclidean space $\mathbb{R}^d$. Let $\bx \in \supp(\bX) $ be a fixed feature vector. 

\begin{assumption} \label{cond:tail}
	There exists some constant $ \alpha > 0 $ such that 
	$ \mathbb{P} (\| \bX - \bx\| \ge R ) \leq e^{-\alpha R} $ for each $ R >0 $.
\end{assumption}

\begin{assumption} \label{cond2}
	The density $f(\cdot)$ is bounded away from $0$ and $\infty$, $f(\cdot)$ and $\mu(\cdot)$ are four times continuously differentiable with bounded second, third, and fourth-order partial derivatives in a neighborhood of $\bx$, and $\mathbb{E} \, Y^2 < \infty$. Moreover, the model error $\epsilon$ has zero mean and finite variance $\sigma_\epsilon^2 > 0$, and is independent of $\bX$. 
	
	%
	%
	%
	%
\end{assumption}

\begin{assumption} \label{cond3}
	We have an i.i.d. sample $\{(\bX_1, Y_1), (\bX_2, Y_2), \ldots, (\bX_n, Y_n)\}$ of size $n$ from model \eqref{eq: model}. 
\end{assumption}

We begin with presenting an asymptotic expansion of the bias of single-scale DNN estimator in the theorem below.

\begin{theorem} \label{thm: bias}
	Assume that Conditions \ref{cond:tail}--\ref{cond3} hold and $s \rightarrow \infty$. Then for any fixed $\bx \in \supp(\bX) \subset \mathbb{R}^d$, we have
	\begin{equation} \label{neweq.FL111}
		\mathbb{E} \, D_n(s) (\bx) = \mu(\bx) + B(s)
	\end{equation}
	with
	\begin{align} 
		\label{eqn:main-thm:bias}	B(s) & =  \Gamma(2/d + 1) \frac{f(\bx) \, \tr (\mu''(\bx)) + 2 \, \mu'(\bx)^Tf'(\bx)}{2 \, d \, V_d^{2/d} \, f(\bx)^{1+2/d}} \, s^{-2/d} + R(s), \\
		\label{eqn:main-thm:remainder-bias}
		R (s)  & = \left \{
		\begin{aligned}
			& O (s ^{ - 3} ) , \quad & d = 1,\\
			& O (s ^{ - 4 / d}), \quad & d \geq 2, 
		\end{aligned}
		\right.
	\end{align}
	where $	V_d = \frac{\pi^{d/2}}{\Gamma(1+d/2)}$, $\Gamma(\cdot)$ is the gamma function, $f'(\cdot)$ and $\mu'(\cdot)$ denote the first-order gradients of $f(\cdot)$ and $\mu(\cdot)$, respectively, $f''(\cdot)$ and $\mu''(\cdot)$ represent the $d\times d$ Hessian matrices of $f(\cdot)$ and $\mu(\cdot)$, respectively, and $\tr(\cdot)$ stands for the trace of a given matrix.
\end{theorem}

Theorem \ref{thm: bias} above shows that the first-order asymptotic bias of the single-scale DNN estimator $D_n(s) (\bx)$ is of order $s^{-2/d}$, and the second-order asymptotic bias is of order $s^{-4/d}$  for $ d \geq 2$ and of order $ s^{ - 3} $ for $d = 1$.  The rate of convergence for the bias term becomes slower as the feature dimensionality $d$ grows, which is common for nonparametric estimators.  It thus would be beneficial to remove the first-order asymptotic bias completely to improve the finite-sample performance.  

We relate our results to the existing literature. \cite{biau2010rate} showed that DNN  achieves the optimal convergence rate of $n^{1/(d+2)}$ under the Lipschitz continuity assumption on the regression function when $d\geq 3$. Our Theorem \ref{thm: bias} is proved assuming the fourth-order smoothness condition (see Condition \ref{cond2}). Under the fourth-order smoothness condition, DNN does not achieve the nonparametric optimal rate of $n^{4/(d+8)}$, mainly because of the bias.  Our results reveal that in such a case, bias reduction is needed for improved convergence rate. The fourth-order smoothness condition is mainly used to obtain the explicit form of the coefficient in front of the first-order bias $s^{-2/d}$ and the order of the remainder $R(s)$, which are critical for successful bias reduction and also play important roles in developing our asymptotic normality theory (which until now has been absent from literature). In addition, as mentioned in the Introduction, the de-biasing idea here can be similarly applied by constructing the multi-scale DNN to further reduce the higher-order bias under even higher-order smoothness assumption.      

\begin{corollary} \label{cor-1}
 	Assume that Conditions \ref{cond:tail}--\ref{cond3} hold and $s \rightarrow \infty$. Then for any fixed $\bx \in \supp(\bX) \subset \mathbb{R}^d$ and the two-scale DNN estimator with weights defined in \eqref{wei1}--\eqref{wei2}, we have
 	 \begin{equation}
 	     \e D_n (s_1, s_2) (\bx) = \mu (\bx) + R(s),
 	 \end{equation}
 	 where 
 	 \begin{align*}
 	    	R (s)  & = \left \{
		\begin{aligned}
			& O (s ^{ - 3} ) , \quad & d = 1,\\
			& O (s ^{ - 4 / d}), \quad & d \geq 2. 
		\end{aligned}
		\right.
 	 \end{align*}
\end{corollary}

By using the TDNN estimator, the asymptotic bias reduces to the second-order term  $O(s_1^{-4/d} + s_2^{-4/d})$ for $ d \geq 2 $ and $O(s_1^{- 3} + s_2^{- 3}) $ for $ d = 1 $. Corollary \ref{cor-1} is a direct consequence of Theorem \ref{thm: bias}. 

We further characterize the asymptotic distribution of the single-scale DNN estimator in the following theorem, which is new to the literature.

\begin{theorem} \label{thm: DNN-normality}
	Assume that Conditions \ref{cond:tail}--\ref{cond3} hold, $ s \to \infty $, and $ s = o(n) $. Then for any fixed $\bx \in \supp(\bX) \subset \mathbb{R}^d$, it holds that for some positive sequence $ \sigma_n$ of order $(s/n)^{1/2}$, 
	\begin{equation}\label{eq: DNN-normality}
		\frac{D_n(s)(\bx) - \mu(\bx) - B(s)}{\sigma_n} \toD N(0,1)
	\end{equation}
	as $n \rightarrow \infty$, where $B(s)$ is given in (\ref{eqn:main-thm:bias}).
\end{theorem}

Theorem \ref{thm: DNN-normality} requires the assumptions of $ s \to \infty $ and $ s = o(n) $, where the former leads to vanishing bias and the latter leads to controlled variance asymptotically. The technical analysis of Theorem \ref{thm: DNN-normality} exploits Hoeffding's canonical decomposition \citep{Hoeffding_A_1948} which is an extension of the H\'ajek projection. 

Despite the U-statistic representation of $D_n (s) ({\bf x })$ given in (\ref{neweq.FL001}), the classical U-statistic asymptotic theory (e.g., \cite{Serfling_Approximation_1980, Korolyuk_Theory_1994}) is not readily applicable because of the typical assumption of \textit{fixed} subsampling scale $s$. In contrast, our method requires the opposite assumption of \textit{diverging} subsampling scale $s$. 
 \color{black}{
	Such a statistic is called an infinite-order U-statistic (IOUS) and has gained more interest in the recent literature; see, e.g., \cite{borovskikh1996u, frees1989infinite, song2019approximating, athey2019generalized}. Unfortunately, the assumptions on the kernel functions of the U-statistics in most IOUS literature are not satisfied for the TDNN. For instance, \cite{frees1989infinite} assumed that the kernels are converging as the sample size grows. 
	However, in our case, the kernel $\Phi(\bx; \bZ_{i_1}, \bZ_{i_2}, \ldots, \bZ_{i_s}) = Y_{(1)}(\bZ_{i_1}, \bZ_{i_2}, \ldots, \bZ_{i_s}) $  becomes degenerate as $s$ tends to infinity. Another example is \cite{borovskikh1996u} who considered scalar-valued random variables. In our case, $\bZ_{i}$'s are vector-valued and thereby the results of \cite{borovskikh1996u} are not readily applicable. 
	
	In \cite{wager2018estimation}, the asymptotic distribution of the random forests \citep{Breiman2001,Breiman2002,ChiVosslerFanLv2020} estimator was studied via examining the asymptotic normality of the IOUS. Both their proof and ours rely on Hoeffding's decomposition of the U-statistics (and in particular IOUS) to establish the asymptotic normality. However, the main challenge in these proofs is controlling the variance of the first-order  H{\'a}jek projection. This variance term takes different forms for nearest neighbors methods and tree based methods, and thus it needs to be handled differently for each case.  For instance, Theorem 3.3 and Corollary 3 in \cite{wager2018estimation} demonstrate bounds for tree based methods, which are not directly extendable to the nearest neighbors methods. Instead, we use Lemma {\color{blue} 7} in Section {\color{blue}E.6} of the Supplementary Material to bound variance specifically for our method.

	Recently, \cite{song2019approximating} established convergence theory similar to our Theorem \ref{thm: DNN-normality} under more general setting and more complicated assumptions which also concern the kernel of the U-statistics and the H{\'a}jek projection of the kernel. In contrast, our Theorem \ref{thm: DNN-normality} is developed under simpler assumptions that are more targeted to the TDNN. It might be possible to check the conditions and then employ the results of \cite{song2019approximating} to prove our Theorem \ref{thm: DNN-normality}. However, the efforts on checking these assumptions can be rather significant and even comparable to the full development of our proof. 
	}

We proceed with characterizing the asymptotic distribution for the two-scale DNN estimator introduced in (\ref{TDNN}).

\begin{theorem} \label{thm1}
	Assume that Conditions \ref{cond:tail}--\ref{cond3} hold,  $ s_2 \to \infty$, $s_2 = o(n)$, and there exist some constants $0 < c_1 < c_2 < 1$ such that $ c_1 \leq  s_1/ s_2 \leq c_2 $. Then for any fixed $\bx \in \supp(\bX) \subset \mathbb{R}^d$, it holds that for some positive sequence $ \sigma_n$ of order $(s_2/n)^{1/2}$, 
	\begin{equation}
		\frac{D_n(s_1, s_2)(\bx) - \mu(\bx) - \Lambda}{\sigma_n} \toD N(0,1)
	\end{equation}
	as $n \rightarrow \infty$, where $ \Lambda = O(s_1^{-4/d} + s_2^{-4/d})$ for $ d \geq 2 $ and $  \Lambda = O(s_1^{- 3} + s_2^{- 3}) $ for $ d = 1 $. 
\end{theorem}

We note that the positive sequence $\sigma_n$ in Theorem \ref{thm1} is different from the sequence $\sigma_n$ in Theorem \ref{thm: DNN-normality}, with the former representing the asymptotic standard deviation of the TDNN estimator and the latter representing the asymptotic standard deviation of the single-scale DNN estimator.  We use the same generic notation for the convenience of technical presentation.  Since the explicit form of the asymptotic standard deviation will not be used, this should not cause any confusion. 
 Theorem \ref{thm1}  requires both subsampling scales $s_1$ and $s_2$ to diverge and be of smaller orders of the full sample size $n$ in order to best trade off between the squared bias and variance.  We would like to point out that Theorem \ref{thm1} is not a simple consequence of Theorem \ref{thm: DNN-normality}, since marginal asymptotic normalities do not necessarily entail joint asymptotic normality. To deal with such a technical difficulty, we have to jointly analyze the two single-scale DNN estimators. A key ingredient of our technical analysis of Theorem \ref{thm1} is to show that the TDNN estimator also admits a U-statistic representation, which enables us to exploit Hoeffding's decomposition and calculate the variances of the kernel and the associated first-order H\'ajek projection.  

{\color{black}
We also obtain the theorem below on the mean-squared error (MSE) of our TDNN estimator. Setting $ c = (s_1 / s_2)^{2/d} $, the weights of the two single-scale DNN estimators are given by $ w_1^* = c / (c - 1) $ and $ w_2^* = - 1 / (c - 1) $ according to \eqref{wei1} and \eqref{wei2}.
\begin{theorem} \label{thm-MSE}
   Assume that Conditions \ref{cond:tail}--\ref{cond3} hold, $ s_2 \to \infty$, $s_2 = o(n)$, and {\color{black}$c$ is a constant in $(0, 1)$}. Then for any fixed $\bx \in \supp (\bX) \subset \mathbb{R}^d$, we have that when $d \geq 2$,
   \begin{equation}
       \begin{split}
        & \e \big\{ D_n (s_1, s_2) (\bx) - \mu (\bx) \big\}^2 \\
        & \leq  \frac { A } {(c - 1)^2} \Big\{ R_1 (\bx, d, f, \mu) c^{-2} s_2^{ - 8/ d} +  \sigma^2_{\epsilon} \frac {s_2} {n}\Big\},
       \end{split}
   \end{equation}
   and when $d = 1$,
   \begin{equation}
       \begin{split}
        & \e \big\{ D_n (s_1, s_2) (\bx) - \mu (\bx) \big\}^2 \\
        & \leq  \frac { A } {(c - 1)^2} \Big\{ R_2 (\bx, d, f, \mu) c^{-1} s_2^{ - 6} +  \sigma^2_{\epsilon} \frac {s_2} {n}\Big\},
       \end{split}
   \end{equation}
   where $A$ is some positive constant, and $R_1 (\bx, d, f, \mu)$ and $R_2 (\bx, d, f, \mu)$ are some constants depending on the bounds of the first four derivatives of $f(\cdot)$ and $\mu(\cdot)$ in a neighborhood of $\bx$. 
\end{theorem}
 Theorem \ref{thm-MSE} provides an upper bound for the pointwise MSE and such result can be applied easily to obtain the integrated MSE under some regularity conditions. 
 The optimal choice of subsampling scale $s_2$ in terms of achieving the best bias-variance tradeoff is given by $s_2 = O(n^{d/(8+d)})$ for $d \geq 2$ and $s_2=O(n^{1/7})$ for $d = 1$,  yielding the corresponding consistency rate at the order of $O(n^{-4/(8+d)})$ for $d \geq 2$ and $O(n^{ - 3/7})$ for $d=1$.  Note that such rate of convergence is minimax optimal (see, e.g., \cite{stone1982optimal}) when $d\geq 2$ under the smoothness assumptions in Condition \ref{cond2} . Compared to the result in \cite{biau2015lectures} where the minimax optimal convergence rate for the single-scale DNN was obtained for $d\geq 3$ under the Liptchitz continuity condition, our result still remains minimax optimal when $d \geq 2$ 
 under different smoothness assumptions in Condition \ref{cond2}.   

}

	\section{Variance  estimates for two-scale DNN estimator} \label{Sec4}
	
	
	\subsection{Jackknife estimator} \label{Sec4.1}
	
	
	As unveiled in Lemma {\color{blue} 8} in Section {\color{blue}E.7} of the Supplementary Material, the two-scale DNN estimator $ D_n(s_1, s_2) ({\bf x}) $ with $s_1 < s_2$ admits the U-statistic representation
	\begin{equation} 
		D_n (s_1, s_2) ({\bf x}) = {n \choose s_2}^{-1} \sum_{1 \leq i_1 < i_2 < \ldots < i_{s_2} \leq n } \Phi^*({\bf x}; {\bf Z}_{i_1}, {\bf Z}_{i_2}, \ldots, {\bf Z}_{i_{s_2}} ),
		\label{TDNN-U}
	\end{equation}
	where the new kernel function is given by 
	\begin{equation*}
		\Phi^{*} ( {\bf x}; {\bf Z}_1, {\bf Z}_2, \ldots,  {\bf Z}_{s_2} ) = w_1^*  \Phi^{(1)} ({\bf x}; {\bf Z}_{1}, {\bf Z}_2, \ldots, {\bf Z}_{s_2} ) +  w_2^* \Phi   ({\bf x}; {\bf Z}_{1}, {\bf Z}_2, \ldots, {\bf Z}_{s_2} )
	\end{equation*}
	with {\color{black}
		$
		\Phi^{(1)} ({\bf x}; {\bf Z}_{1}, {\bf Z}_2, \ldots, {\bf Z}_{s_2} ) = {s_2 \choose s_1}^{-1} \sum_{1 \leq i_1 < i_2 < \ldots <  i_{s_1} \leq s_2} \Phi ({\bf x}; {\bf Z}_{i_1}, {\bf Z}_{i_2}, \ldots, {\bf Z}_{i_{s_1}}),  
		$}
	$\Phi(\bx;\cdot)$ the original kernel function involved in the single-scale DNN estimator introduced in (\ref{neweq.FL001}), and $ w_1^* $ and $w_2^*$ the weights defined in  equations \eqref{wei1} and \eqref{wei2}. 
	We denote by 
	\begin{equation} \label{neweq.FL032}
		\sigma_n^2 = \Var (D_n (s_1, s_2) ({\bf x}) )
	\end{equation}
	the variance of the two-scale DNN estimator $D_n (s_1, s_2) ({\bf x})$, where we drop the subscript $n$ in this population variance for notational simplicity. 
	
	For each $1 \leq i \leq n$, let us define the two-scale DNN estimator obtained after deleting the $i$th observation as in (\ref{TDNN-U})
	\begin{align} \label{neweq.FL033}
		U_{n - 1}^{(i)} = {n - 1 \choose s_2}^{-1} \sum_{\substack{ 1 \leq j_1 < j_2 < \ldots < j_{s_2} \leq n \\ j_1, j_2, \ldots, j_{s_2} \neq i }} \Phi^* ( { \bf x }; {\bf Z}_{j_1}, {\bf Z}_{j_2}, \ldots, {\bf Z}_{j_{s_2} } ) .
	\end{align}
	Then the jackknife estimator \citep{Q1949, Q1956} for $\sigma_n^2$ in (\ref{neweq.FL032}) is given by
	\begin{equation} \label{jack}
		\hat{\sigma}_{J}^2 = \frac {n - 1  } { n } \sum_{i = 1} ^n \big( U_{n- 1}^{(i )} - D_n (s_1, s_2) ({\bf x}) \big)^2. 
	\end{equation}  
	We formally establish the ratio consistency of the jackknife estimator $\hat{\sigma}_{J}^2$ introduced in (\ref{jack}) in the theorem below.
	
	\begin{theorem}  \label{thm-jack}
		Assume that Conditions \ref{cond2}--\ref{cond3} hold, $ \e [Y^4] < \infty $, $ \e [  {\epsilon}^4  ] < \infty $, $ s_1 \to \infty$, and $s_2 \to \infty  $ with some constants $ 0 < c_1  < c_2 < 1 $ such that $   c_1 \leq s_1 / s_2 \leq c_2 $. Then for any fixed $\bx \in \supp(\bX) \subset \mathbb{R}^d$, when $  s_2 = o( n^{1/3} ) $ it holds that 
		$	\hat{\sigma}_{J}^2 / \sigma_n^2 \stackrel{p}{\to} 1 $ as $n \rightarrow \infty$.
	\end{theorem}
	
	The proof of Theorem \ref{thm-jack} still builds on the U-statistic framework. 
	Similar to the discussion after Theorem \ref{thm: DNN-normality}, the conventional technical arguments in \cite{A1969} for the consistency of the jackknife estimator for the U-statistic are not applicable because of the diverging $s_1$ and $s_2$. As seen in Section {\color{blue}D.5} of Supplementary Material, our technical analysis involves rather delicate calculations of the remainders.
	We acknowledge that the assumption of $s_2 = o(n^{1/3})$ is not necessarily optimal. Moreover, the assumption on the finite fourth moments can be relaxed to finite $(2 + 2 \delta)$th moments with some $0 < \delta < 1$. Consequently, the bound on the order of $s_2$ will depend on parameter $\delta$ accordingly. 
	
	We point out that although the U-statistic representation plays a crucial role in obtaining our theoretical results, the computational cost of the jackknife estimator utilizing such a representation can become excessively prohibitive in practice. Instead, we should take advantage of the L-statistic representation revealed in Lemma \ref{le-L} to efficiently compute the U-statistics $ \{ U_{n - 1}^{(i)} \}_{1 \leq i \leq n }$ and the two-scale DNN estimator $ D_n (s_1, s_2) ({\bf x }) $ involved in the jackknife estimator $\hat{\sigma}_{J}^2$ in (\ref{jack}). When the sample size $n$ becomes large, one can speed up the implementation of jackknife using approximation with subsampling.

	\subsection{Bootstrap estimator} \label{Sec4.2}
	
	The bootstrap method \citep{E1979}  has been widely used  for estimating the parameters and the distributions of statistics of interest, empowering statistical inference. We now consider the nonparametric bootstrap for estimating the variance of the two-scale DNN estimator. Given $n$ observations $ \{{\bf Z}_1, {\bf Z}_2, \ldots, {\bf Z}_n \}$, we denote by $\{ {\bf Z}_1^*, {\bf Z}_2^*, \ldots, {\bf Z}_n^* \}$ a bootstrap sample selected independently and uniformly from the original $n$ observations with replacement. As in (\ref{TDNN-U}), let us construct the two-scale DNN estimator 
	\begin{equation} \label{boot-Dn}
		D_n^{*} (s_1, s_2) ({\bf x})  = {n \choose s_2}^{-1} \sum_{1 \leq i_1 < i_2 < \ldots < i_{s_2} \leq n } \Phi^* ( {\bf x}; {\bf Z}_{ i_1}^*, {\bf Z}_{ i_2}^*, \ldots, {\bf Z}_{ i_{s_2} }^* )
	\end{equation}
	based on the bootstrap sample $\{ {\bf Z}_1^*, {\bf Z}_2^*, \ldots, {\bf Z}_n^* \}$. 
	
	We choose the number of bootstrap samples as $ B \geq 1 $. For each $ 1 \leq b \leq B $, we independently select a bootstrap sample $ \{{\bf Z}_{b, 1}^{*} , {\bf Z}_{b, 2}^*, \ldots, {\bf Z}_{b, n}^* \}$ and calculate the corresponding bootstrap version of the two-scale DNN estimator $ D_n^{(b)} (s_1, s_2) ({\bf x}) $ as in (\ref{boot-Dn}). Observe that given the original observations $ ( {\bf Z}_1, {\bf Z}_2, \ldots, {\bf Z}_n )$, the bootstrap samples $ \{ ({\bf Z}_{b, 1}^*, {\bf Z}_{b, 2}^*, \ldots, {\bf Z}_{b, n}^* ) \}_{1 \leq b \leq B} $ are independently and identically distributed as $ ({\bf Z}_1^*, {\bf Z}_2^*, \ldots, {\bf Z}_n^*) $. 
	Then the bootstrap estimator for $\sigma_n^2$ in (\ref{neweq.FL032}) is given by 
	\begin{equation} \label{B-var}
		\hat{\sigma}_{B, n}^2 = \frac {1} { B - 1 } \sum_{b = 1}^B \big( D_n^{(b)} (s_1, s_2) ({\bf x})  - \bar{D}_{B, n} \big)^2,
	\end{equation} 
	where $ \bar{D}_{B, n} = \frac {1}{B} \sum_{b = 1}^B D_n^{(b)} (s_1, s_2) ({\bf x}) $. The ratio consistency of the bootstrap estimator $\hat{\sigma}_{B, n}^2$ introduced in (\ref{B-var}) is shown formally in the following theorem.
	
	\begin{theorem}  \label{thm-boot}
		Assume that Conditions \ref{cond2}--\ref{cond3} hold, $ \e [Y^4] < \infty $, $ \e [  {\epsilon}^4  ] < \infty $, $ s_1 \to \infty$, and $s_2 \to \infty  $ with some constants $ 0 < c_1  < c_2 < 1 $ such that $   c_1 \leq s_1 / s_2 \leq c_2 $. Then for any fixed $\bx \in \supp(\bX) \subset \mathbb{R}^d$, when $  s_2 = o( n^{1/3} ) $ and $ B \to \infty $, it holds that $ \hat{\sigma}_{B, n}^2  / \sigma_n^2  \stackrel{p}{\to} 1 $ as $n \rightarrow \infty$.
	\end{theorem}
	
	Let us gain some insights into the technical analysis for the consistency of the bootstrap estimator  established in Theorem \ref{thm-boot}. First, we observe that conditional on $ ({\bf Z}_1, {\bf Z}_2, \ldots, {\bf Z}_n) $,  the bootstrap versions of the TDNN estimator $ {D_n^{(b)}} (s_1, s_2) ({\bf x}) $ are i.i.d. random variables and thus the law of large numbers entails that $ \hat{\sigma}_{B, n}^2 $ is asymptotically close to the conditional variance $ \Var ( D_n^*(s_1, s_2) ({\bf x}) | {\bf Z}_1, {\bf Z}_2, \ldots, {\bf Z}_n ) $ as $B \rightarrow \infty$. Second, since the bootstrap samples are independently drawn from the empirical distribution based on $ ({\bf Z}_1, {\bf Z}_2, \ldots, {\bf Z}_n) $ and the empirical distribution converges to the underlying distribution of $ {\bf Z} $ asymptotically,  the bootstrap version $ \Var ( D_n^*(s_1, s_2) ({\bf x}) | {\bf Z}_1, {\bf Z}_2, \ldots, {\bf Z}_n ) $ of the variance will converge to the population quantity $ \sigma^2 $ as $n \rightarrow \infty$. It is worth mentioning that for the second part of our technical analysis, we resort to the consistency result of the jackknife estimator established in Theorem \ref{thm-jack}. In particular, we see that the jackknife and the bootstrap are asymptotically equivalent in the variance estimation for the TDNN estimator. Indeed, \cite{E1979} showed that the jackknife can be viewed as a linear approximation method for the bootstrap. It was also pointed out in \cite{E1979} that the jackknife can fail for certain nonsmooth functionals, while the bootstrap can still work.

\ignore{	
	\red{
		Now, we introduce an alternative version bootstrap algorithm called divide and conquer algorithm that is borrowed from \cite{song2019approximating}.
		Let $ S_1 \subset \{1, \dots, n\}$ and $ |S_1| = n_1$. Let $ K = \lfloor (n-1)/(s-1) \rfloor$. For any $i_1 \in S_1$, we form disjoint subsets $S_{2,k}^{(i_1)} $ of size $s-1$ for $ k = 1, \dots K$ such that their union is $ S_1 $. Define $\bar{S}_{2,k}^{(i_1)} = S_{2,k}^{(i_1)} \cup \{ i_1\}$, 
		$ G_{i_1} = K^{-1} \sum_{k=1}^{K} \phi (Z_{\bar{S}_{2,k}^{(i_1)}} )$ , and 
		$ \bar{G} = n_1 ^{-1} \sum_{i_1 \in S_1} G_{i_1} $. Finally, the bootstrap sample is given by
		\begin{equation}
			U^{\#}_{n_1} = n_1 ^ {-1/2} \sum_{i_1 \in S_1} \xi_{i_1} (G_{i_1} - \bar{G} )
		\end{equation}
		where $ \{ \xi_{i_1}, i_1 \in S_1 \} $  is an i.i.d. standard Gaussian random variables independent of $\bZ$.
	}
	\red{
		\begin{theorem}[Semi-parametric Bootstrap] \label{thm-boot-semiparam}
			Assume that
			(i) $\eta_1 > 0 $ \\
			(ii) For some $C_n$, $\mathbb{E} g_1(Z_1) ^4 \leq \eta_1 C_n$ \\
			(iii) $ \| \phi( Z_1, \dots, Z_s) - \mathbb{E} \phi \| _{\Psi_q} \leq C_n $ \\
			(iv) $ \frac{C_n \log^{q_2}(n)}{\eta_1 n_1} \leq C n ^{-\zeta} $
			where $ q_2 = \max \{(4/q + 1), 5\}$, $ \| \cdot \| _{\Psi_q} $ is the Orlicz norm, and $ \phi _q (x) = e^{x^q} - 1$.
			Then,
			\begin{equation}
				sup_{a<b} | \mathbb{P}( a \leq U^{\#}_{n_1} \leq b ) - \mathbb{P} (a \leq Y \leq b) | \leq C n ^{-\zeta/6}
			\end{equation}
			where Y is normal random variable with variance $\sigma_n^2$
		\end{theorem}
	}

	\red{
		This theorem and hence the conditions are borrowed from \cite{song2019approximating}. The condition (i) is relatively mild condition on the variance of the first H{\'a}jek projection. Indeed, we can see that this condition is satisfied by \ref{l-bdd}. Section 4.1 of \cite{song2019approximating} provides more cases where this condition is satisfied even beyond our setting. On the other hand, as mentioned before, \cite{song2019approximating} requires stronger assumptions such as (ii)-(iv). These assumptions are not only depend on the distribution of the covariates but also the kernel function of the U-statistics, and its H{\'a}jek projection. For instance, the condition (ii) is on the moment of the first H{\'a}jek projection projection and the condition (iii) is on the tail behaviour of the 
		Condition (iv) is just controlling relative sizes of the constants in conditions (ii) and (iii). Even though, it is not direct consequence of Lemma \ref{new.lem.3}, the conditions (ii) and (iii) above borrowed from \cite{song2019approximating} and our Lemma \ref{new.lem.3} is closely related. 
		We refer to \cite{song2019approximating} for the proof of the Theorem and discussion of the assumptions. 
		Finally, if we replace $\phi, s, g_1 $, and $\eta_1$ with $\phi^{\ast}, s_2, g_1 ^{\ast}$, and $\eta_1^{\ast}$ (defined in the proof of Theorem \ref{thm1}), respectively, we obtain the bootstrap approximation for the two-scale DNN.
	}
}

\section{Simulation studies}%
\label{sec:SimulationStudies}
	
In this section, we investigate the finite-sample performance of the TDNN estimator for nonparametric estimation and inference in comparison to the DNN and $k$-NN.
	We use equations \eqref{wei1} and \eqref{wei2} to construct weights for the TDNN estimator. Specifically, we choose $w_{1}^{*} = -1 /(c^{2/d} - 1)$ and $w_{2}^{*} = c^{2/d} / (c^{2/d} - 1)$ with $ c = s_2/s_1$. Without loss of generality, we set $c > 1$.
	It is seen that $w_1^* < 0$ and hence the TDNN estimator assigns negative weights to some nearest neighbors which manages to reduce the bias for DNN. However, to control the variance of TDNN, the ratio $c = s_2/s_1$ should be chosen appropriately away from one. 
	With the above choice of weights, there are two parameters to tune for the TDNN: 
 subsampling scale	$s_1$ and the ratio $c = s_2/s_1$. To tune the parameters for  prediction of a given feature vector $\bx$, we perform a weighted leave-one-out cross-validation (LOOCV) procedure using each of the $B$ nearest neighbors to $\bx$ as a single left-out observation. Specifically, we set aside each of the $B$ nearest neighbors to $\bx$ and make prediction for it using the TDNN estimator with all the remaining $n - 1$ observations and the given combination $(c, s_1)$. Then the tuned $(c, s_1)$ is obtained by minimizing a weighted sum of the squared error over those $B$ left-out nearest neighbors, where the weights are defined by the standard Gaussian kernel distances of the nearest neighbors to the given feature vector $\bx$. Finally, we calculate our
    TDNN estimate $D_n(s_1, cs_1)(\bx)$ for the given point $\bx$ using the $s_1$ and $c$ selected by our weighted LOOCV tuning procedure.  
    
	In our analysis, we always select the ratio $c$ from a set of values. Then for a given value of $c$, we provide our choices of the subsampling scale $s_1$ through a sign-change tuning method. 
	Specifically, for the prediction of a feature vector $\bx$, we compute the TDNN estimator $D_{n}(s_1, cs_{1})(\bx)$ for each consecutive $s_1$ starting from $1$. We continue this process until the difference in the absolute differences of consecutive TDNN estimators changes sign. Intuitively, the sign change represents the value of $s_1$ where the curvature of the  TDNN estimator as a function of $s_1$ changes. We denote the subsampling scale chosen by the sign-change tuning process as $s_{\text{sign}}$. This process is motivated by the curve structure in Figure~\ref{fig:biasMSEDNN1} from the simulation example in Section~\ref{subsec:TDNN vs. DNN}. One issue with the simple sign-change tuning method is that we may risk selecting a value of $s_1$ that corresponds to a local minimum for the MSE of TDNN as a function of $s_1$. To mitigate such concern, we consider a sequence of subsampling scales in the next step of our tuning process with $s_{\text{sign}}$ as our lower limit and $2s_{\text{sign}}$ as our upper limit, where the initial value $s_{\text{sign}}$ given by the sign-change tuning method provides a warm start for and specifies the order of tuning parameter $s_1$.



	\subsection{Two-scale DNN versus DNN}%
	\label{subsec:TDNN vs. DNN}
	
	To illustrate the effectiveness of the two-scale framework compared to the single-scale DNN, we simulate $n=1000$ data points from the following model. 
	\begin{setting} \label{setting_1}
	Assume that $ Y = \mu(\mathbf{X}) + \epsilon$, where $\mu(\mathbf{X} ) = (x_{1}-1)^{2} + (x_{2} +1)^{3} -3x_{3}$ with $\mathbf{X} = (x_1, x_2, x_3)^T$ and $(\bX^T, \epsilon)^{T} \sim N(\mathbf{0},I_{4})$.
	\end{setting} 
	 Our goal here is to  compare the mean-squared error (MSE) of the TDNN estimator with those of the DNN and $k$-NN estimators at a fixed test point chosen to be $(0.5,-0.5,0.5)^{T}$. 
	 For the implementation of the DNN, we estimate the regression function at this test point and calculate the MSE while varying the subsampling scale $s$ from 1 to 250.  
	{For the TDNN, we estimate the regression function with fixed $c = 2$ for simplicity and $s_1$ varying from 1 to 250. 
	}
	
			\begin{figure}[htp]
		\centering
		\includegraphics[width=3.5in, height=3in]{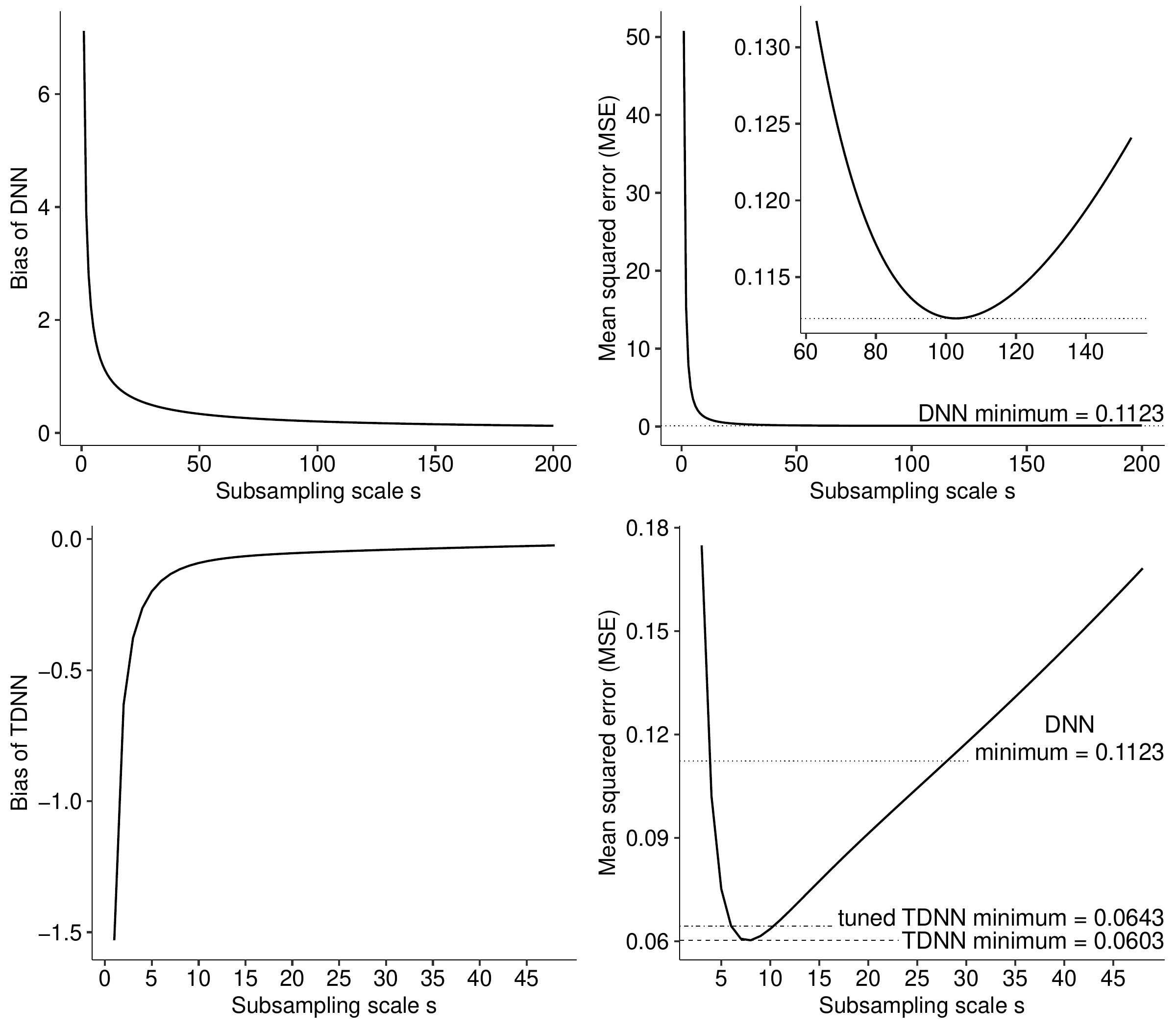}
		\caption{
			The results of simulation setting \ref{setting_1} described in Section~\ref{subsec:TDNN vs. DNN} for DNN and TDNN. The rows show the bias and MSE as  functions of the subsampling scale $s$ for DNN and TDNN, respectively. The top right panel also depicts a zoomed-in plot where the U-shaped pattern is more apparent. The dashed lines in the MSE plots are labeled with the minimum MSE value for each of the methods. The tuned TDNN MSE minimum corresponds to the weighted LOOCV tuning method described at the beginning of Section~\ref{sec:SimulationStudies}.  
		}
		\label{fig:biasMSEDNN1}
	\end{figure}

	Figure~\ref{fig:biasMSEDNN1} presents the simulation results for DNN and TDNN in terms of both the bias and the MSE. 
	A first observation is that as the subsampling scale $s$ increases, the bias of the DNN estimator shrinks toward zero, which is intuitive from a geometric perspective since larger subsampling scale $s$ leads to the use of the information in the sample more concentrated around  the fixed test point. From the MSE plot for DNN, we observe the classical U-shaped pattern of the bias-variance tradeoff. Thanks to the higher-order asymptotic expansions, the two-scale procedure of TDNN is completely free of the first-order asymptotic bias. The substantial difference between the dominating first-order asymptotic bias in DNN and the second-order asymptotic bias in TDNN at the finite-sample level is evident in the left panel of Figure~\ref{fig:biasMSEDNN1}.

	From the MSE plot for TDNN, we also see a similar bias-variance tradeoff. An interesting phenomenon by comparing the two smooth U-shaped curves in the right panel of Figure~\ref{fig:biasMSEDNN1} is that the minimum of the MSE for TDNN is attained at a much smaller subsampling scale $s$ than that for DNN. Furthermore, we observe that because of the reduced finite-sample bias, TDNN attains a more than 45\% reduction of minimum MSE compared to the single-scale DNN. We also show in the bottom right panel of Figure~\ref{fig:biasMSEDNN1} the MSE obtained by our weighted LOOCV tuning procedure for TDNN described at the beginning of Section~\ref{sec:SimulationStudies}, \textit{without} using any knowledge of the underlying true regression function. We see that our tuning procedure provides a good approximation to the true MSE despite considering a smaller range of subsampling scales in a data-adaptive way. Finally, an additional comparison of TDNN and $k$-NN is included in Section {\color{blue}C.1} of the Supplementary Material.
	

	\subsection{Comparisons with DNN and $k$-NN for nonparametric inference}%
	\label{subsec:comparisonswithcausalforest}
	
	We further compare TDNN with DNN and $k$-NN over two simulation examples in terms of the estimation accuracy in nonparametric regression settings.
	
	
	For each simulation setting, we use a training sample size of $n = 1000$ and the summary statistics are calculated based on 1000 simulation replications. Throughout our simulations, we estimate the variance of the TDNN estimator and DNN estimator using the bootstrap method that has been theoretically justified in Section \ref{Sec4.2}. As for the inference by the $k$-NN estimator, we adopt the modeling strategy in \cite{wager2018estimation} and model $\hat{\mu}_{kNN}$ as Gaussian with mean $\mu(\bx)$ and variance $\hat{\sigma}^2_{kNN} / (k-1)$, where $\hat{\sigma}^2_{kNN}$ is the sample variance over the $k$ nearest neighbors. We tune our TDNN estimator using the weighted LOOCV tuning method by leaving out each of the $B$ nearest neighbors of a given feature vector $\bx$ to predict, which has been described at the beginning of Section~\ref{subsec:TDNN vs. DNN}. We also adopt the same weighted LOOCV tuning strategy for the DNN estimator. We employ the \texttt{kknn} R package \citep{knnRpackage} to tune the neighborhood size $k$ for the $k$-NN estimator using the leave-one-out cross-validation. In our simulation studies, $B$ for the weighted LOOCV tuning procedure is always chosen as $20$, the subsampling scales $s$ for DNN varies from 1 to 250, {and the neighborhood size $k$ for $k$-NN varies from 1 to 200.}

	The first simulation setting in this section also uses Setting \ref{setting_1} described in Section~\ref{subsec:TDNN vs. DNN}. We evaluate the performance of TDNN, DNN, and $k$-NN in terms of the bias, variance, and MSE at a fixed test point $(0.5, -0.5, 0.5)^T$ as well as for a set of 100 random test points drawn from the distribution of the covariates $\bX \sim N({\bf 0}, I_3)$. The MSE, bias, and variance for the set of random test points are obtained by averaging over all the random test points. For the TDNN estimator, the ratio $c = s_2/s_1$ is chosen from the sequence $\{2,4,6,8,10,15,20,25,30\}$ for the random test points and we fix $c = 2$ for the fixed test point for simplicity. The subsampling scale $s_1$ is chosen from the interval $[s_{\text{sign}}, 2s_{\text{sign}}]$ for each given $c$, where $s_{\text{sign}}$ is given by the sign-change tuning process (related to the curvature) introduced at the beginning of Section \ref{sec:SimulationStudies}. 

	\begin{table}[tp]
 	     \resizebox{0.8\textwidth}{!}{\begin{minipage}{\textwidth}
        \centering
        
\begin{tabular}{lcccccc}
\toprule
\multicolumn{1}{c}{ } & \multicolumn{3}{c}{Fixed Test Point} & \multicolumn{3}{c}{Random Test Points} \\
\cmidrule(l{3pt}r{3pt}){2-4} \cmidrule(l{3pt}r{3pt}){5-7}
Method & MSE & Bias$^2$ & Variance & MSE & Bias$^2$ & Variance\\
\midrule
DNN & 0.1249 & 0.0556 & 0.0623 & 15.0989 & 14.4701 & 0.5968\\
$k$-NN & 0.3207 & 0.0062 & 0.3114 & 9.4558 & 6.8510 & 2.2138\\
TDNN & 0.0576 & 0.0082 & 0.0464 & 7.3142 & 5.3296 & 1.5235\\
\bottomrule
\end{tabular}

        \end{minipage}}
       \centering
	    \caption{Comparison of DNN, $k$-NN, and TDNN in simulation setting \ref{setting_1} described in Section~\ref{subsec:TDNN vs. DNN}.}
	    \label{table:simsetting_1_normal}
	\end{table}

		\begin{table}[tp]
		\resizebox{0.8\textwidth}{!}{\begin{minipage}{\textwidth}
				\centering
				
\begin{tabular}{lccccccc}
\toprule
\multicolumn{2}{c}{ } & \multicolumn{3}{c}{Fixed Test Point} & \multicolumn{3}{c}{Random Test Points} \\
\cmidrule(l{3pt}r{3pt}){3-5} \cmidrule(l{3pt}r{3pt}){6-8}
Method & p & MSE & Bias$^2$ & Variance & MSE & Bias$^2$ & Variance\\
\midrule
DNN & 3 & 0.7266 & 0.5159 & 0.2379 & 5.6775 & 4.5131 & 1.2315\\
$k$-NN & 3 & 0.6630 & 0.1453 & 0.5158 & 5.1323 & 2.2046 & 2.8698\\
TDNN & 3 & 0.2594 & 0.0535 & 0.2707 & 3.4788 & 1.6190 & 2.0933\\
\hline
DNN & 5 & 0.7271 & 0.5176 & 0.2390 & 5.9057 & 4.6419 & 1.2638\\
$k$-NN & 5 & 0.6699 & 0.1664 & 0.5272 & 5.3712 & 2.3072 & 2.9450\\
TDNN & 5 & 0.2579 & 0.0699 & 0.2789 & 3.6883 & 1.6990 & 2.1640\\
\hline
DNN & 10 & 0.8756 & 0.5822 & 0.2602 & 6.2705 & 4.8632 & 1.3218\\
$k$-NN & 10 & 0.8297 & 0.1992 & 0.5643 & 5.7003 & 2.4493 & 3.0715\\
TDNN & 10 & 0.2867 & 0.0503 & 0.2987 & 3.9243 & 1.7798 & 2.3084\\
\hline
DNN & 15 & 0.9376 & 0.6434 & 0.2685 & 6.4583 & 4.9885 & 1.3466\\
$k$-NN & 15 & 0.7919 & 0.2213 & 0.5693 & 5.8418 & 2.5189 & 3.1275\\
TDNN & 15 & 0.2823 & 0.0439 & 0.3083 & 4.0509 & 1.8257 & 2.3743\\
\hline
DNN & 20 & 0.9653 & 0.6341 & 0.2735 & 6.8909 & 5.2649 & 1.4073\\
$k$-NN & 20 & 0.8174 & 0.1868 & 0.5729 & 6.2427 & 2.6994 & 3.2703\\
TDNN & 20 & 0.3298 & 0.0530 & 0.3276 & 4.4064 & 1.9583 & 2.5184\\
\bottomrule
\end{tabular}

		\end{minipage}}
		\centering

		\caption{Comparison of DNN, $k$-NN, and TDNN in  simulation setting \ref{setting_2} described in Section~\ref{subsec:comparisonswithcausalforest}. 
		}
		\label{table:simsetting2}
	\end{table}
	
	We observe from Table \ref{table:simsetting_1_normal} that for both fixed test point and random test points, the TDNN estimator significantly outperforms the DNN and $k$-NN estimators in terms of MSE. In addition, the improvement over the DNN is mainly due to the largely reduced bias, which is in line with our theory. In contrast, the TDNN has reduced variance compared to the $k$-NN, because TDNN is a bagged statistic and the bagging technique is known to be successful in variance reduction. We can see that the average MSE over a set of random test points is much larger than the MSE at the fixed test point (0.5, -0.5, 0.5). The main reason is that the covariate vector $\bX$ is generated from a normal distribution and the density function at extreme values is close to zero, and thus the theoretical MSE can be very large for those extreme points. As a comparison, we also present the simulating results under the same setting except that $\bX \sim U([0, 1 ]^3)$ in Section {\color{blue}C.2} of the Supplementary Material. It is seen from Table {\color{blue}4} in Section {\color{blue}C.2} of Supplementary Material that under the uniform distribution setting, the MSE for random test points is only slightly larger than the MSE at the fixed test point.

	{For the second simulation setting, we investigate the performance of TDNN, DNN, and $k$-NN in the setting below, which is a modified version of a simulation setting first considered in \cite{DettePepe2010}.}
	\begin{setting} \label{setting_2}
	{Assume that $ Y = \mu(\mathbf{X}) + \epsilon$, where $\mu(\mathbf{X} ) = 4(4x_{1} - 2 + 8x_{2}^{2})^{2} + (3-4x_{2})^{2} + 16\sqrt{x_{3} + 1}(2x_{3}-1)^{2}$} with $\mathbf{X} = (x_1, \cdots, x_p)^T$, $\bX \sim U([0,1]^{p})$, and $\epsilon \sim N(\mathbf{0}, 1)$ independent of $\bX$. We increase the ambient dimensionality $p$ along the sequence $\{ 3,5,10,15,20 \}$.
	\end{setting}

	Since the theoretical properties of TDNN established in this paper rely on the assumption of fixed dimensionality, it is natural to expect that the performance of TDNN can deteriorate as the dimensionality grows. To alleviate such difficulty, we exploit the feature screening idea \citep{Fan_Sure_2008,FanFan2008,Fan_Sure_2018} for dimension reduction to accompany the implementation of TDNN. For the screening step, we test the null hypothesis of independence between the response and each feature using the nonparametric tool of distance correlation statistic \citep{Szekely2007,Gao2020} and calculate the corresponding p-value. Then we select features with p-values less than $\alpha/p$ with some significance level $\alpha \in (0, 1)$ and make prediction by using these selected features. For our simulation studies, we fix {$\alpha = 0.001$}. For the TDNN estimator, the ratio $c = s_2/s_1$ is chosen from the sequence $\{2,4,6,8,10,15,20,25,30\}$ for random test points and we fix $c = 2$ for the fixed test point for simplicity. The subsampling scale $s_1$ is chosen from the interval $[s_{\text{sign}}, 2s_{\text{sign}}]$ for each given $c$, where $s_{\text{sign}}$ is given by the sign-change tuning process introduced at the beginning of Section \ref{sec:SimulationStudies}.

	We again evaluate the performance of the three estimators at a fixed test point chosen as $x_{1} = 0.2$, $x_{2} = 0.4$, $x_{3} = 0.6$, and $x_{j} = 0.5$ for $j > 3$ as well as for a set of 100 test points randomly drawn from the hypercube $[0,1]^p$. The simulation results in Table \ref{table:simsetting2} show that the screening technique works well and the TDNN estimator has significantly reduced MSEs compared to the single-scale DNN and $k$-NN estimators. Observe that although the density function of the covariates is uniform, the average MSE for random test points is larger than the MSE for the fixed test point because the MSE also depends on the values of the regression function and its derivatives. 

\section{Real data application} %
\label{sec:RealDataAnalysis}

In this section, we demonstrate the practical performance of the suggested TDNN procedure for nonparametric learning on the Abalone data set, which is available at the UCI repository (\url{https://archive.ics.uci.edu/ml/datasets/abalone}). The Abalone data set has been widely investigated in the literature for the illustration of various nonparametric regression methods; see, e.g., \cite{breiman1999using, Breiman2001} and \cite{steele2009exact}. This data set contains $4177$ observations on $8$ input variables and a response that represents the number of rings indicating the age of an abalone. The major goal of this real data application is to predict the response based on the information of the $8$ input variables. Since the first input variable is categorical and consists of three categories indicating the sex (Male, Female, and Infant), we only search nearest neighbors restrictively in each category. 
Consequently, there are $7$ features after splitting the data set into three categories. 
Because the nonparametric rate of convergence for the nearest neighbors methods becomes slower as the feature dimensionality grows, we exploit the popular tool of principal component analysis (PCA) to reduce the dimensionality of the feature space and employ the first $m$ principal components for nonparametric learning. In our analysis, we choose $m = 3$ since the first three principal components account for more than $99\%$ of the variation in the response. 
	
Specifically, we randomly set aside $25\%$ of the $4177$ observations as a test set and train the TDNN estimator based on the remaining $75\%$ of the observations. As mentioned in Section \ref{sec:SimulationStudies}, the tuning of the two subsampling scales $s_1$ and $s_2$ is equivalent to that of the subsampling scale $s_1$ and their ratio $c = s_2/s_1$. We adopt the same strategy as described in Section \ref{sec:SimulationStudies} to tune both parameters $s_1$ and $c$ for the TDNN in a data-adaptive fashion. For a given feature vector $\bx$ in the test set, each of the $B$ nearest neighbors to $\bx$ is chosen as the left-out observation in the weighted LOOCV tuning procedure. Then the tuned $(c, s_1)$ is obtained by minimizing the weighted squared error over those $B$ left-out observations with the weights defined by the corresponding standard Gaussian kernel distance to the given feature vector $\bx$. Finally, we apply the TDNN estimator constructed with the tuned $(c, s_1)$ to the test set and calculate the prediction error in terms of the MSE. The above procedure involving random data splitting is repeated $50$ times and the prediction errors are averaged over those $50$ random splits.

\begin{table}[tp]
	    \centering
	    \begin{tabular}{l c c c c}
	    \toprule
	        Method & $k$-NN & DNN & TDNN & RF  \\
	         MSE & 4.99 & 4.553 &  4.512 & 4.60 \\
	    \bottomrule
	    \end{tabular}
	    \caption{The MSEs of different nonparametric learning methods on the real data application in Section \ref{sec:RealDataAnalysis}.}
	    \label{table_realdata}
\end{table}
	
In particular, we tune $(c, s_1)$ from $c \in \{1.2, 1.5, 2, 3, 4, 5, 6, 7,  8, 9, 10, 15, 20\}$ and $s_1 \in [s_{\text{sign}},\, 2 s_{\text{sign}}]$ with $s_{\text{sign}}$ obtained by the sign-change tuning process (related to the curvature) introduced in Section \ref{sec:SimulationStudies}. The subsamping size $s$ for the DNN estimator is chosen from the sequence starting from 50 to 250 with an increment of 5. We set the neighborhood size of $B = 50$ for the implementation of the weighted LOOCV tuning procedure. We compare the prediction performance of the TDNN to that of the $k$-NN, DNN, and random forests (RF) in terms of the MSE evaluated on the test data. Table \ref{table_realdata} summarizes the results of all the nonparametric learning methods on this real data application. In particular, the results for the $k$-NN and RF are extracted from \cite{steele2009exact}. Indeed, from Table \ref{table_realdata} we see that TDNN improves  
over both $k$-NN and DNN at the finite-sample level, which is in line with our theoretical results and simulation examples. Moreover, the TDNN also outperforms the RF. In contrast, there still lack optimality results for the tool of the RF.

\ignore{
	The major goal of this real data application is to characterize the heterogeneity of a treatment effect across subpopulations defined by the values of some continuous covariate. In particular, we aim to study the effect of smoking on a child's birth weight across mothers' ages. We utilize the data set studied originally in \cite{Abrevaya2015} with a kernel-based estimator\footnote{The data set used in our analysis can be found on the research web page of Robert P. Lieli, https://sites.google.com/site/robertplieli/research.}.
	
	\begin{figure}[h]
		\centering
		\includegraphics[width=5.5in,  height=2.8in]{real_data_combined_plot_update.png}
		\caption{The effect of mother's smoking on a child's birth weight as a function of mother's age. The error bars correspond to 95\% confidence intervals. 
		}
		\label{fig:real_data_figure}
	\end{figure}
	
	
	For our analysis, the feature vector $\mathbf{x}$ includes mother's age, mother's education, father's education, gestation length in weeks, and the number of prenatal visits.
	We use this subset of covariates because each of these covariates has been shown to have some association with low birth weight \citep{Silvestrin2013}. 
	The response $Y$ is the child's birth weight. The binary treatment indicator $T$ is whether or not the mother smoked during pregnancy. In particular, the data set consists of 591,547 observations in total, 85,976 of whom smoked during pregnancy.  
	We estimate the heterogeneous treatment effects with all the features fixed at the average levels of the corresponding treatment group, except for mother's age which we vary from 16 to 35. This allows us to see how the effect of smoking changes with age, while controlling for all other observed confounders. Similarly as in the simulation examples in Section \ref{sec:SimulationStudies}, we also conduct the same analysis using the causal forests (CF) 
	as a popular benchmark for nonparametric heterogeneous treatment effect inference. \textcolor{red}{For the TDNN estimator, we choose the subsampling scales $s_{1}$ and $s_{2}$ using the data-driven tuning strategy described in Section~\ref{sec:SimulationStudies}.}
	
	
	Figure~\ref{fig:real_data_figure} presents the heterogeneous treatment effect estimates for both TDNN and CF along with 95\% confidence intervals. 
	The confidence intervals for TDNN are generated by bootstrapping the difference between the treatment and control groups, while the confidence intervals for CF are generated using the variance estimation method in the R package \texttt{grf}. The results from both procedures suggest that as age increases, the decrease in a newborn's weight associated with a mother's smoking behavior becomes larger. In particular, we see from Figure~\ref{fig:real_data_figure} that the TDNN estimates exhibit a monotone decreasing relationship between mother's age and the heterogeneous treatment effect of smoking with tight confidence intervals. In contrast, the CF estimates show a much more irregular relationship with wide confidence intervals.
	
}

\section{Discussion} \label{Sec7}

In this paper, we have investigated the problems of estimation and inference for nonparametric mean regression function using the two-scale DNN (TDNN), a bias reduced estimator based on the distributional nearest neighbors (DNN). Our suggested method of TDNN alleviates the finite-sample bias issue of the classical $k$-nearest neighbors and admits easy implementation with simple tuning under the assumption of the fourth-order smoothness on the mean regression function. We have provided theoretical justifications for the proposed estimator and established the asymptotic normality theory for practical use of TDNN in nonparametric statistical inference with optimality.  The new TDNN tool can be exploited for the heterogeneous treatment effect (HTE) estimation and inference that is key to identifying individualized treatment effects. 
	
	Our bias reduction idea can be generalized to construct the multi-scale DNN  when the mean regression function has even higher-order smoothness. In such case, DNN or TDNN no longer enjoys the nonparametric minimax optimal convergence rate. By exploiting higher-order asymptotic bias expansion, a multi-scale DNN can be constructed in the same fashion for achieving the nonparametric optimal convergence rate. We leave the detailed investigations for future study. 
	
	It would also be interesting to extend the idea of TDNN to the settings of  diverging or high feature dimensionality and consider the non-i.i.d. data settings such as time series, panel, and survival data. Since the distance function plays a natural role in identifying the nearest neighbors, it would be interesting to investigate the choice of different distance metrics, aside from the Euclidean distance, that are pertinent to specific manifold structures intrinsic to data. These problems are beyond the scope of the current paper and will be interesting topics for future research. 
       	

\bibliographystyle{chicago}
\bibliography{references}

\begin{thebibliography}{}

\bibitem[\protect\citeauthoryear{Arvesen}{Arvesen}{1969}]{A1969}
Arvesen, J.~N. (1969).
\newblock Jackknifing {$U$}-statistics.
\newblock {\em Ann. Math. Statist.\/}~{\em 40}, 2076--2100.

\bibitem[\protect\citeauthoryear{Athey, Tibshirani, Wager, et~al.}{Athey
  et~al.}{2019}]{athey2019generalized}
Athey, S., J.~Tibshirani, S.~Wager, et~al. (2019).
\newblock Generalized random forests.
\newblock {\em Annals of Statistics\/}~{\em 47\/}(2), 1148--1178.

\bibitem[\protect\citeauthoryear{Berrett, Samworth, and Yuan}{Berrett
  et~al.}{2019}]{Berrett_Efficient_2018}
Berrett, T.~B., R.~J. Samworth, and M.~Yuan (2019).
\newblock Efficient multivariate entropy estimation via $k$-nearest neighbour
  distances.
\newblock {\em The Annals of Statistics\/}~{\em 47}, 288--318.

\bibitem[\protect\citeauthoryear{Berry}{Berry}{1941}]{berry1941}
Berry, A.~C. (1941).
\newblock The accuracy of the {G}aussian approximation to the sum of
  independent variates.
\newblock {\em Trans. Amer. Math. Soc.\/}~{\em 49}, 122--136.

\bibitem[\protect\citeauthoryear{Biau, C\'{e}rou, and Guyader}{Biau
  et~al.}{2010}]{biau2010rate}
Biau, G., F.~C\'{e}rou, and A.~Guyader (2010).
\newblock On the rate of convergence of the bagged nearest neighbor estimate.
\newblock {\em Journal of Machine Learning Research\/}~{\em 11}, 687--712.

\bibitem[\protect\citeauthoryear{Biau and Devroye}{Biau and
  Devroye}{2015}]{biau2015lectures}
Biau, G. and L.~Devroye (2015).
\newblock {\em Lectures on the nearest neighbor method}.
\newblock Springer.

\bibitem[\protect\citeauthoryear{Borovkov}{Borovkov}{2013}]{Borovkov_Probability_2013}
Borovkov, A.~A. (2013).
\newblock {\em Probability Theory}.
\newblock Springer.

\bibitem[\protect\citeauthoryear{Borovskikh}{Borovskikh}{1996}]{borovskikh1996u}
Borovskikh, I. I.~V. (1996).
\newblock {\em U-statistics in Banach Spaces}.
\newblock VSP.

\bibitem[\protect\citeauthoryear{Breiman}{Breiman}{1996}]{breiman1996bagging}
Breiman, L. (1996).
\newblock Bagging predictors.
\newblock {\em Machine learning\/}~{\em 24\/}(2), 123--140.

\bibitem[\protect\citeauthoryear{Breiman}{Breiman}{1999}]{breiman1999using}
Breiman, L. (1999).
\newblock Using adaptive bagging to debias regressions.
\newblock Technical report, Technical Report 547, Statistics Dept. UCB.

\bibitem[\protect\citeauthoryear{Breiman}{Breiman}{2001}]{Breiman2001}
Breiman, L. (2001).
\newblock Random forests.
\newblock {\em Machine Learning\/}~{\em 45}, 5--32.

\bibitem[\protect\citeauthoryear{Breiman}{Breiman}{2002}]{Breiman2002}
Breiman, L. (2002).
\newblock Manual on setting up, using, and understanding random forests v3.1.
\newblock {\em Statistics Department University of California Berkeley, CA,
  USA\/}~{\em 1}, 58.

\bibitem[\protect\citeauthoryear{Calonico, Cattaneo, and Farrell}{Calonico
  et~al.}{2018}]{calonico2018effect}
Calonico, S., M.~D. Cattaneo, and M.~H. Farrell (2018).
\newblock On the effect of bias estimation on coverage accuracy in
  nonparametric inference.
\newblock {\em Journal of the American Statistical Association\/}~{\em
  113\/}(522), 767--779.

\bibitem[\protect\citeauthoryear{Cheang and Reinsel}{Cheang and
  Reinsel}{2000}]{cheang2000bias}
Cheang, W.-K. and G.~C. Reinsel (2000).
\newblock Bias reduction of autoregressive estimates in time series regression
  model through restricted maximum likelihood.
\newblock {\em Journal of the American Statistical Association\/}~{\em
  95\/}(452), 1173--1184.

\bibitem[\protect\citeauthoryear{Chi, Vossler, Fan, and Lv}{Chi
  et~al.}{2020}]{ChiVosslerFanLv2020}
Chi, C.-M., P.~Vossler, Y.~Fan, and J.~Lv (2020).
\newblock Asymptotic properties of high-dimensional random forests.
\newblock {\em arXiv preprint arXiv:2004.13953\/}.

\bibitem[\protect\citeauthoryear{Crump, Hotz, Imbens, and Mitnik}{Crump
  et~al.}{2008}]{crump2008nonparametric}
Crump, R.~K., V.~J. Hotz, G.~W. Imbens, and O.~A. Mitnik (2008).
\newblock Nonparametric tests for treatment effect heterogeneity.
\newblock {\em Review of Economics and Statistics\/}~{\em 90}, 389--405.

\bibitem[\protect\citeauthoryear{Dette and Pepelyshev}{Dette and
  Pepelyshev}{2010}]{DettePepe2010}
Dette, H. and A.~Pepelyshev (2010).
\newblock Generalized latin hypercube design for computer experiments.
\newblock {\em Technometrics\/}~{\em 52\/}(4), 421--429.

\bibitem[\protect\citeauthoryear{Efron}{Efron}{1979}]{E1979}
Efron, B. (1979).
\newblock Bootstrap methods: another look at the jackknife.
\newblock {\em Ann. Statist.\/}~{\em 7}, 1--26.

\bibitem[\protect\citeauthoryear{Fan and Fan}{Fan and Fan}{2008}]{FanFan2008}
Fan, J. and Y.~Fan (2008).
\newblock High dimensional classification using features annealed independence
  rules.
\newblock {\em Annals of statistics\/}~{\em 36}, 2605.

\bibitem[\protect\citeauthoryear{Fan and Lv}{Fan and Lv}{2008}]{Fan_Sure_2008}
Fan, J. and J.~Lv (2008).
\newblock Sure independence screening for ultrahigh dimensional feature space
  (with discussion).
\newblock {\em Journal of the Royal Statistical Society: Series B
  {(Statistical} Methodology)\/}~{\em 70}, 849--911.

\bibitem[\protect\citeauthoryear{Fan and Lv}{Fan and Lv}{2018}]{Fan_Sure_2018}
Fan, J. and J.~Lv (2018).
\newblock Sure independence screening (invited review article).
\newblock {\em Wiley {StatsRef:} Statistics Reference Online\/}.

\bibitem[\protect\citeauthoryear{Frees}{Frees}{1989}]{frees1989infinite}
Frees, E.~W. (1989).
\newblock Infinite order u-statistics.
\newblock {\em Scandinavian Journal of Statistics\/}, 29--45.

\bibitem[\protect\citeauthoryear{Gao, Fan, Lv, and Shao}{Gao
  et~al.}{2021}]{Gao2020}
Gao, L., Y.~Fan, J.~Lv, and Q.~Shao (2021).
\newblock Asymptotic distributions of high-dimensional distance correlation
  inference.
\newblock {\em The Annals of Statistics\/}~{\em 49}, 1999--2020.

\bibitem[\protect\citeauthoryear{Gy{\"o}rfi, Kohler, Krzyżak, and
  Walk}{Gy{\"o}rfi et~al.}{2002}]{Gyrfi_A_2002}
Gy{\"o}rfi, L., M.~Kohler, A.~Krzyżak, and H.~Walk (2002).
\newblock {\em A {Distribution-Free} Theory of Nonparametric Regression}.
\newblock Springer.

\bibitem[\protect\citeauthoryear{Hahn, Murray, Carvalho, et~al.}{Hahn
  et~al.}{2020}]{hahn2020bayesian}
Hahn, P.~R., J.~S. Murray, C.~M. Carvalho, et~al. (2020).
\newblock Bayesian regression tree models for causal inference: regularization,
  confounding, and heterogeneous effects.
\newblock {\em Bayesian Analysis\/}.

\bibitem[\protect\citeauthoryear{H{\'a}jek}{H{\'a}jek}{1968}]{Hjek_Asymptotic_1968}
H{\'a}jek, J. (1968).
\newblock Asymptotic normality of simple linear rank statistics under
  alternatives.
\newblock {\em The Annals of Mathematical Statistics\/}~{\em 39}, 325--346.

\bibitem[\protect\citeauthoryear{Hall}{Hall}{1992}]{hall1992effect}
Hall, P. (1992).
\newblock Effect of bias estimation on coverage accuracy of bootstrap
  confidence intervals for a probability density.
\newblock {\em The Annals of Statistics\/}, 675--694.

\bibitem[\protect\citeauthoryear{Hall}{Hall}{2013}]{hall2013bootstrap}
Hall, P. (2013).
\newblock {\em The bootstrap and Edgeworth expansion}.
\newblock Springer Science \& Business Media.

\bibitem[\protect\citeauthoryear{Hall and Samworth}{Hall and
  Samworth}{2005}]{hall2005bagging}
Hall, P. and R.~J. Samworth (2005).
\newblock Properties of bagged nearest neighbour classifiers.
\newblock {\em J. R. Stat. Soc. Ser. B Stat. Methodol.\/}~{\em 67\/}(3),
  363--379.

\bibitem[\protect\citeauthoryear{Hechenbichler and Schliep}{Hechenbichler and
  Schliep}{2004}]{knnRpackage}
Hechenbichler, K. and K.~Schliep (2004).
\newblock Weighted k-nearest-neighbor techniques and ordinal classification.

\bibitem[\protect\citeauthoryear{Hitsch and Misra}{Hitsch and
  Misra}{2018}]{hitsch2018heterogeneous}
Hitsch, G.~J. and S.~Misra (2018).
\newblock Heterogeneous treatment effects and optimal targeting policy
  evaluation.
\newblock {\em Available at SSRN 3111957\/}.

\bibitem[\protect\citeauthoryear{Hoeffding}{Hoeffding}{1948}]{Hoeffding_A_1948}
Hoeffding, W. (1948).
\newblock A class of statistics with asymptotically normal distribution.
\newblock {\em The Annals of Mathematical Statistics\/}~{\em 19}, 293--325.

\bibitem[\protect\citeauthoryear{Imbens and Rubin}{Imbens and
  Rubin}{2015}]{Imbens_Causal_2015}
Imbens, G.~W. and D.~B. Rubin (2015).
\newblock {\em Causal Inference in Statistics, Social, and Biomedical
  Sciences}.
\newblock Cambridge University Press.

\bibitem[\protect\citeauthoryear{Korolyuk and Borovskich}{Korolyuk and
  Borovskich}{1994}]{Korolyuk_Theory_1994}
Korolyuk, V.~S. and Y.~V. Borovskich (1994).
\newblock {\em Theory of U-statistics}.
\newblock Springer.

\bibitem[\protect\citeauthoryear{Leblanc}{Leblanc}{2010}]{leblanc2010bias}
Leblanc, A. (2010).
\newblock A bias-reduced approach to density estimation using bernstein
  polynomials.
\newblock {\em Journal of Nonparametric Statistics\/}~{\em 22\/}(4), 459--475.

\bibitem[\protect\citeauthoryear{Lee}{Lee}{2009}]{lee2009non}
Lee, M.-j. (2009).
\newblock Non-parametric tests for distributional treatment effect for randomly
  censored responses.
\newblock {\em Journal of the Royal Statistical Society: Series B (Statistical
  Methodology)\/}~{\em 71}, 243--264.

\bibitem[\protect\citeauthoryear{Lin, Ding, and Han}{Lin
  et~al.}{2021}]{lin2021estimation}
Lin, Z., P.~Ding, and F.~Han (2021).
\newblock Estimation based on nearest neighbor matching: from density ratio to
  average treatment effect.
\newblock {\em arXiv preprint arXiv:2112.13506\/}.

\bibitem[\protect\citeauthoryear{Mack}{Mack}{1980}]{Mack_Local_1980}
Mack, Y. (1980).
\newblock Local properties of {k-NN} regression estimates.
\newblock {\em {SIAM} Journal on Algebraic Discrete Methods\/}~{\em 2},
  311--323.

\bibitem[\protect\citeauthoryear{Newey, Hsieh, and Robins}{Newey
  et~al.}{2004}]{newey2004twicing}
Newey, W.~K., F.~Hsieh, and J.~M. Robins (2004).
\newblock Twicing kernels and a small bias property of semiparametric
  estimators.
\newblock {\em Econometrica\/}~{\em 72\/}(3), 947--962.

\bibitem[\protect\citeauthoryear{Peng, Coleman, and Mentch}{Peng
  et~al.}{2019}]{PCM2019}
Peng, W., T.~Coleman, and L.~Mentch (2019).
\newblock Asymptotic distributions and rates of convergence for random forests
  via generalized {$U$}-statistics.
\newblock {\em arXiv preprint arXiv:1905.10651\/}.

\bibitem[\protect\citeauthoryear{Powers, Qian, Jung, Schuler, Shah, Hastie, and
  Tibshirani}{Powers et~al.}{2017}]{powers2017some}
Powers, S., J.~Qian, K.~Jung, A.~Schuler, N.~H. Shah, T.~Hastie, and
  R.~Tibshirani (2017).
\newblock Some methods for heterogeneous treatment effect estimation in
  high-dimensions.
\newblock {\em arXiv preprint arXiv:1707.00102\/}.

\bibitem[\protect\citeauthoryear{Quenouille}{Quenouille}{1949}]{Q1949}
Quenouille, M.~H. (1949).
\newblock Approximate tests of correlation in time-series.
\newblock {\em J. Roy. Statist. Soc. Ser. B\/}~{\em 11}, 68--84.

\bibitem[\protect\citeauthoryear{Quenouille}{Quenouille}{1956}]{Q1956}
Quenouille, M.~H. (1956).
\newblock Notes on bias in estimation.
\newblock {\em Biometrika\/}~{\em 43}, 353--360.

\bibitem[\protect\citeauthoryear{Rubin}{Rubin}{1974}]{Rubin_Estimating_1974}
Rubin, D.~B. (1974).
\newblock Estimating causal effects of treatments in randomized and
  nonrandomized studies.
\newblock {\em Journal of Educational Psychology\/}~{\em 66}, 688--701.

\bibitem[\protect\citeauthoryear{Samworth}{Samworth}{2012}]{Samworth_Optimal_2012}
Samworth, R.~J. (2012).
\newblock Optimal weighted nearest neighbour classifiers.
\newblock {\em The Annals of Statistics\/}~{\em 40}, 2733--2763.

\bibitem[\protect\citeauthoryear{Schucany, Gray, and Owen}{Schucany
  et~al.}{1971}]{schucany1971bias}
Schucany, W., H.~Gray, and D.~Owen (1971).
\newblock On bias reduction in estimation.
\newblock {\em Journal of the American Statistical Association\/}~{\em
  66\/}(335), 524--533.

\bibitem[\protect\citeauthoryear{Schucany and Sommers}{Schucany and
  Sommers}{1977}]{schucany1977improvement}
Schucany, W. and J.~P. Sommers (1977).
\newblock Improvement of kernel type density estimators.
\newblock {\em Journal of the American Statistical Association\/}~{\em
  72\/}(358), 420--423.

\bibitem[\protect\citeauthoryear{Serfling}{Serfling}{1980}]{Serfling_Approximation_1980}
Serfling, R.~J. (1980).
\newblock {\em Approximation Theorems of Mathematical Statistics}.
\newblock Wiley Series in Probability and Statistics.

\bibitem[\protect\citeauthoryear{Shalit, Johansson, and Sontag}{Shalit
  et~al.}{2017}]{shalit2017estimating}
Shalit, U., F.~D. Johansson, and D.~Sontag (2017).
\newblock Estimating individual treatment effect: generalization bounds and
  algorithms.
\newblock In {\em Proceedings of the 34th International Conference on Machine
  Learning-Volume 70}, pp.\  3076--3085. JMLR. org.

\bibitem[\protect\citeauthoryear{Song, Chen, Kato, et~al.}{Song
  et~al.}{2019}]{song2019approximating}
Song, Y., X.~Chen, K.~Kato, et~al. (2019).
\newblock Approximating high-dimensional infinite-order $ u $-statistics:
  Statistical and computational guarantees.
\newblock {\em Electronic Journal of Statistics\/}~{\em 13\/}(2), 4794--4848.

\bibitem[\protect\citeauthoryear{Steele}{Steele}{2009}]{steele2009exact}
Steele, B.~M. (2009).
\newblock Exact bootstrap k-nearest neighbor learners.
\newblock {\em Machine Learning\/}~{\em 74\/}(3), 235--255.

\bibitem[\protect\citeauthoryear{Stone}{Stone}{1982}]{stone1982optimal}
Stone, C.~J. (1982).
\newblock Optimal global rates of convergence for nonparametric regression.
\newblock {\em The annals of statistics\/}, 1040--1053.

\bibitem[\protect\citeauthoryear{Sz{\'e}kely, Rizzo, and Bakirov}{Sz{\'e}kely
  et~al.}{2007}]{Szekely2007}
Sz{\'e}kely, G.~J., M.~L. Rizzo, and N.~K. Bakirov (2007).
\newblock Measuring and testing dependence by correlation of distances.
\newblock {\em The Annals of Statistics\/}~{\em 35}, 2769--2794.

\bibitem[\protect\citeauthoryear{Wager and Athey}{Wager and
  Athey}{2018}]{wager2018estimation}
Wager, S. and S.~Athey (2018).
\newblock Estimation and inference of heterogeneous treatment effects using
  random forests.
\newblock {\em Journal of the American Statistical Association\/}~{\em 113},
  1228--1242.

\bibitem[\protect\citeauthoryear{Wager, Hastie, and Efron}{Wager
  et~al.}{2014}]{wager2014confidence}
Wager, S., T.~Hastie, and B.~Efron (2014).
\newblock Confidence intervals for random forests: The jackknife and the
  infinitesimal jackknife.
\newblock {\em Journal of Machine Learning Research\/}~{\em 15}, 1625--1651.

\bibitem[\protect\citeauthoryear{Zaidi and Mukherjee}{Zaidi and
  Mukherjee}{2018}]{zaidi2018gaussian}
Zaidi, A. and S.~Mukherjee (2018).
\newblock Gaussian process mixtures for estimating heterogeneous treatment
  effects.
\newblock {\em arXiv preprint arXiv:1812.07153\/}.

\end{thebibliography}

	
\newpage
\appendix
\setcounter{page}{1}
\setcounter{section}{0}
\setcounter{equation}{0}

\renewcommand{\theequation}{A.\arabic{equation}}
\renewcommand{\thesubsection}{A.\arabic{subsection}}

	\begin{center}{\bf \Large Supplementary Material to ``Optimal Nonparametric Inference with Two-Scale Distributional Nearest Neighbors"}

	\bigskip
	
	Emre Demirkaya, Yingying Fan, Lan Gao, Jinchi Lv,\\ Patrick Vossler and Jingbo Wang
	\end{center}
	
	\noindent This Supplementary Material contains {a bootstrap estimator for the distribution of TDNN}, an application of TDNN in heterogeneous treatment effect estimation and inference, some additional simulation results,  and the proofs of all main results and key lemmas, as well as some additional technical details.
	
	\renewcommand{\thesubsection}{A.\arabic{subsection}}

	\section{Bootstrap estimator for distribution of TDNN}

	
	We now provide an alternative bootstrap method for directly estimating the distribution of  the TDNN estimator. Denote by $\mathbb{P}^*$ the distribution of a bootstrap sample $(\bZ_1^*, \ldots, \bZ_n^*)$ with replacement conditional on the original $n$ observations $(\bZ_1, \ldots, \bZ_n)$. Let us define $\theta^* = \mathbb{E}\big[ \Phi^* (\bx; \bZ_1^*, \ldots, \bZ_{s_2}^*)) \big| \bZ_1, \ldots, \bZ_n \big]$, where the expectation is taken with respect to the resampling distribution $\mathbb{P}^*$. Recall that $\{(\bX_{(1)}, Y_{(1)}), \ldots, (\bX_{(n)}, Y_{(n)})\}$ is the ascendingly ordered sample by the distance of $\bX_i$ to the given point $\bx$.
	Using the result in \cite{biau2010rate}, we can show that 
	\begin{equation}
		\theta^* = \sum_{i = 1}^n \bigg\{ w_1^*\Big[ \Big( 1 - \frac {i-1} {n} \Big)^{s_1} - \Big(1 - \frac {i} {n} \Big)^{s_1} \Big] + w_2^*\Big[ \Big( 1 - \frac {i-1} {n} \Big)^{s_2} - \Big(1 - \frac {i} {n} \Big)^{s_2} \Big] \bigg\} Y_{(i)}.
	\end{equation}

	The theorem below shows that the conditional distribution of the bootstrapped TDNN estimator is asymptotically equivalent to the distribution of the TDNN estimator.
	\begin{theorem} \label{thm-boot-dist}
		Assume that all the conditions of Theorem \ref{thm-boot} are satisfied. Then we have that as $n \to \infty$,
		\begin{equation}
			\begin{split}
				&\sup_{u \in \mathbb{R}} \Big| \mathbb{P}^* \Big\{ (s_2/n)^{-1/2} [ D_n^*(s_1, s_2) (\bx) - \theta^* ] \leq u \Big\} \\
				& \quad - \mathbb{P} \Big\{ (s_2/n)^{-1/2} [D_n(s_1, s_2) (\bx)  - \mu(\bx) - \Lambda] \leq u \Big\}
				\Big| = o_p(1).
			\end{split}
		\end{equation}
	\end{theorem}
	
	Theorem \ref{thm-boot-dist} lays the theoretical foundation for directly estimating the distribution of the TDNN estimator with the bootstrap. The Glivenko--Cantelli theorem implies that the empirical distribution of i.i.d. observations converges uniformly to the underlying true distribution almost surely as the number of observations grows to infinity. 
	Therefore, practically, we can generate $B$ i.i.d. bootstrap samples $\{(\bZ_{b, 1}^*, \ldots, \bZ_{b, n}^*)\}_{1 \leq b \leq B}$ from $(\bZ_1, \ldots, \bZ_n)$ with replacement for a relatively large value of $B$. Then we can approximate the distribution $\mathbb{P} \{D_n(s_1, s_2) (\bx) - \mu (\bx) - \Lambda \leq u \}$ using $ B^{-1} \sum_{b = 1}^B \mathbbm{1} \{ D_n^{(b)}(s_1, s_2) (\bx) - \theta^* \leq u\}  $ with $\mathbbm{1} \{\cdot\}$ representing the indicator function, which is the empirical distribution of $D_n^* (s_1, s_2) (\bx)$ based on the $B$ bootstrap samples. As a consequence, any quantile of the distribution of $D_n(s_1, s_2) (\bx) - \mu (\bx) - \Lambda$ can be approximated by that of the empirical bootstrap distribution $B^{-1} \sum_{b = 1}^B \mathbbm{1} \{ D_n^*(s_1, s_2) (\bx) - \theta^* \leq u\} $. Accordingly, for each given $\alpha \in (0, 1)$, the two-sided $(1 - \alpha)$-level confidence interval for the mean regression function $\mu(\bx)$ can be constructed as $[D_n(s_1, s_2) (\bx) - (\hat\xi_{1 - \alpha/2} - \theta^*), D_n(s_1, s_2) (\bx) - (\hat\xi_{\alpha/2} - \theta^*)]$ , where $\hat{\xi}_{\alpha/2}$ and $\hat{\xi}_{1 - \alpha/2}$ denote the $\alpha$th and $(1 - \alpha)$th sample quantiles of the bootstrap samples $\{D_n^{(b)}(s_1, s_2) (\bx)\}_{1 \leq b \leq B}$, respectively.

	\renewcommand{\thesubsection}{B.\arabic{subsection}}
\section{Application to heterogeneous treatment effect estimation and inference} \label{sec:HTE}

As an application, we discuss in this section how to exploit the suggested TDNN method to estimate and infer the treatment effects in the potential outcomes model framework \citep{Rubin_Estimating_1974, Imbens_Causal_2015}. The problems of treatment effect estimation and inference have broad applications in a wide variety of scientific areas,  ranging from economics to medical studies. In particular, the estimation and inference of the heterogeneous treatment effect (HTE) which focuses on the unit level effect by considering the treatment effect conditional on the pre-treatment covariates have received rapidly growing attention in recent years because of their ability to provide information that the average treatment effect (ATE) cannot provide. For some recent developments, see, e.g.,  \cite{crump2008nonparametric,lee2009non, wager2018estimation, wager2014confidence, shalit2017estimating,hahn2020bayesian,powers2017some,zaidi2018gaussian}.  

Among the existing literature, the causal $k$-NN (\cite{hitsch2018heterogeneous}) is most closely related to our approach. This method estimates the treatment effect function by taking the difference of two separate  $k$-NN regression function estimates for the treatment group and control group, respectively.  
The tuning parameter of neighborhood size $k$ was chosen by minimizing the squared difference between the estimated treatment effect function and the propensity score weighted response.  However, there lacks theoretical justification for the causal $k$-NN estimator.


Let $Y_{T=1} \in \mathbb R$ and $Y_{T=0}\in \mathbb R$ represent the potential outcomes for the treatment and control groups, respectively, where $T$ denotes the treatment indicator with $T=1$ representing treated and $T=0$ being untreated. Then the observed scalar response can be written as 
\[
Y = T\,Y_{T=1}+(1-T)\,Y_{T=0}.
\]
Denote by $\bX \in \mathbb{R}^d$ the random feature vector for an individual. We consider the randomized experiment setting  which amounts to the choice of constant treatment propensity $\mathbb{P}(T = 1 | \bX, Y_{T=1}, Y_{T=0}) = 1/2$.  Here, $1/2$ can be replaced with any other constant in $(0,1)$. 
Given a fixed feature vector $\bx \in \mathbb{R}^d$, the heterogeneous treatment effect (HTE) of treatment $T$ on response $Y$ is defined as 
\begin{equation} \label{neweq.FL029}
	\tau (\bx) = \mathbb{E} \, [Y_{T=1} - Y_{T=0} | \bX = \bx].
\end{equation}

Since the setting of randomized experiments entails the unconfoundedness given by  $
(Y_{T=0}, Y_{T=1} )\independent	T \; | \; \bX,
$
our goal of HTE estimation and inference for (\ref{neweq.FL029}) reduces to the problem of nonparametric regression applied separately to the treatment and control groups, giving rise to 
\begin{align}\label{define: taux}
	\tau(\bx) &= \mathbb{E}[Y_{T=1}|\bX = \bx] - \mathbb{E}[Y_{T=0}|\bX = \bx] \nonumber \\ 
	&=\mathbb E[Y|\bX=\bx, T=1]-\mathbb E[Y|\bX=\bx, T=0].
\end{align} Specifically, let us consider the nonparametric regression model for the treatment group 
\[
Y_{T=1} = \mu(\bX) + \epsilon
,
\]
where $\mu(\bX) = \mathbb{E}[Y_{T=1}|\bX]$ denotes the true mean regression function and the model error $\epsilon$ with zero mean and finite variance is independent of  $d$-dimensional random feature vector $\bX$. Similarly, we can introduce the corresponding nonparametric regression model for the control group; see Section \ref{Sec3.3} for more detailed technical descriptions.   We will separately apply TDNN to the control and treatment groups and then combine the resulting estimators together using \eqref{define: taux} to estimate the heterogeneous treatment effect.

To formally present the asymptotic theory, let us first introduce some necessary notation. Denote by $n_1$ and $n_0$ the sizes of the i.i.d. samples from the treatment and control groups, respectively.  The assumption of completely randomized experiments entails that  $n_0/n_1 \stackrel{p}{\longrightarrow} 1$ as $n\rightarrow \infty$ and the two samples for the treatment and control groups are independent of each other. Let $\bx \in \supp(\bX_1) \cap \supp(\bX_0)$ be a fixed feature vector, where $\supp(\bX_1)$ and $\supp(\bX_0)$ stand for the supports of the corresponding feature distributions for the treatment and control groups, respectively. Similarly, denote by $\mu_1(\cdot)$ and $\mu_0(\cdot)$ the true mean regression functions corresponding to responses $Y_{T=1}$ and $Y_{T=0}$, respectively, and $\epsilon_1$ and $\epsilon_0$ the model errors, with the subscript indicating the treatment and control groups, respectively. Then we can construct two individual two-scale DNN estimators $D_{n_1}^{(1)}(s_1^{(1)}, s_2^{(1)})(\bx)$ and $D_{n_0}^{(0)}(s_1^{(0)}, s_2^{(0)})(\bx)$ separately based on the treatment and control samples with pairs of subsampling scales $(s_1^{(1)}, s_2^{(1)})$ and $(s_1^{(0)}, s_2^{(0)})$, respectively. 

In view of \eqref{define: taux}, the population version of the heterogeneous treatment effect at the fixed vector $\bx$ is given by 
\begin{equation} \label{neweq.FL112}
	\tau(\bx)  = \mu_1(\bx) - \mu_0(\bx).
\end{equation}
We estimate $\tau(\bx)$  using the following TDNN heterogeneous treatment effect estimator
\begin{equation} \label{neweq.FL031}
	\widehat{\tau}(\bx) = D_{n_1}^{(1)} ( s_1^{(1)}, s_2^{(1)} ) ( {\bf x} ) \\- D_{n_0}^{(0)} ( s_1^{(0)}, s_2^{(0)} ) ( {\bf x} ).
\end{equation}
The theorem below characterizes the asymptotic distribution of the TDNN HTE estimator $\widehat{\tau}(\bx)$. 

\begin{theorem} \label{thm2}
	Assume that Conditions \ref{cond:tail}--\ref{cond3} with the subscripts attached hold for both treatment and control groups.  Further assume that $ s_2^{(i)} \to \infty$, $s_2^{(i)} = o (n) $, and there exist some constants $ 0 < c_1 < c_2 < 1$ such that $ c_1 \leq s_1^{(i)} / s_2^{(i)} \leq c_2$  for $i = 0, 1$. Then for any fixed $\bx \in \supp(\bX_1) \cap \supp(\bX_0) \subset \mathbb{R}^d$, it holds that for some positive sequence $ \sigma_n$ of order $\{(s_2^{(1)}+s_2^{(0)})/n\}^{1/2}$, 
	\begin{equation}
		\frac{[D_{n_1}^{(1)}(s_1^{(1)}, s_2^{(1)})(\bx) - D_{n_0}^{(0)}(s_1^{(0)}, s_2^{(0)})(\bx)]- \tau(\bx) - \Lambda}{\sigma_n} \toD N(0,1) \label{te-normal}
	\end{equation}
	as $n \rightarrow \infty$, where {\color{black} $\Lambda = O\{(s_1^{(1)})^{-4/d}+(s_2^{(1)})^{-4/d}+(s_1^{(0)})^{-4/d}+(s_2^{(0)})^{-4/d}\}$ for $d \geq 2$ and $ \Lambda = O\{(s_1^{(1)})^{-3}+(s_2^{(1)})^{-3}+(s_1^{(0)})^{-3}+(s_2^{(0)})^{-3}\} $ for $d = 1$}.
\end{theorem}

As explained before, the sequence $\sigma_n$ in Theorem \ref{thm2} above is a generic notation representing the asymptotic standard deviation of the TDNN heterogeneous treatment effect estimator.   We see that the subsampling scales need to satisfy that $ s_2^{(i)} \to \infty$ and $s_2^{(i)} = o (n) $ for $i = 0, 1$. The asymptotic bias of the TDNN estimator $\widehat{\tau}(\bx)$ is only of the second order {\color{black} $O\{(s_1^{(1)})^{-4/d}+(s_2^{(1)})^{-4/d}+(s_1^{(0)})^{-4/d}+(s_2^{(0)})^{-4/d}\}$ for $d \geq 2$ and $  O\{(s_1^{(1)})^{-3}+(s_2^{(1)})^{-3}+(s_1^{(0)})^{-3}+(s_2^{(0)})^{-3}\} $ for $d = 1$}. The asymptotic variance  identified in Theorems \ref{thm1} and \ref{thm2} depends generally on the underlying distributions and the fixed vector $\bx$, whose complicated form calls for a need to develop practical approaches to the estimation of the asymptotic variance for the TDNN estimator.

For the practical implementation of TDNN for the HTE inference, we advocate the use of the L-statistic representation. Since the single-scale DNN estimator is an L-statistic as shown in Lemma \ref{le-L}, the two-scale DNN estimator, which is a linear combination of a pair of single-scale DNN estimators, is still an L-statistic. We thus can construct a pair of two-scale DNN estimators separately based on the treatment and control subsamples and then take a difference. As suggested by Theorems \ref{thm-boot} and \ref{thm-boot-dist}, we can further bootstrap such difference by resampling within each group to provide tight heterogeneous treatment effect inference. Therefore, the two-scale procedure of TDNN coupled with the bootstrap enjoys both theoretical justifications and computational scalability.


	\renewcommand{\thesubsection}{C.\arabic{subsection}}
	\section{Additional simulation results} \label{new.secA}
	
	\subsection{Comparison with $k$-NN} \label{new.secA.1}
	
\begin{figure}[htp]
		\centering
		\includegraphics[width=0.8\textwidth]{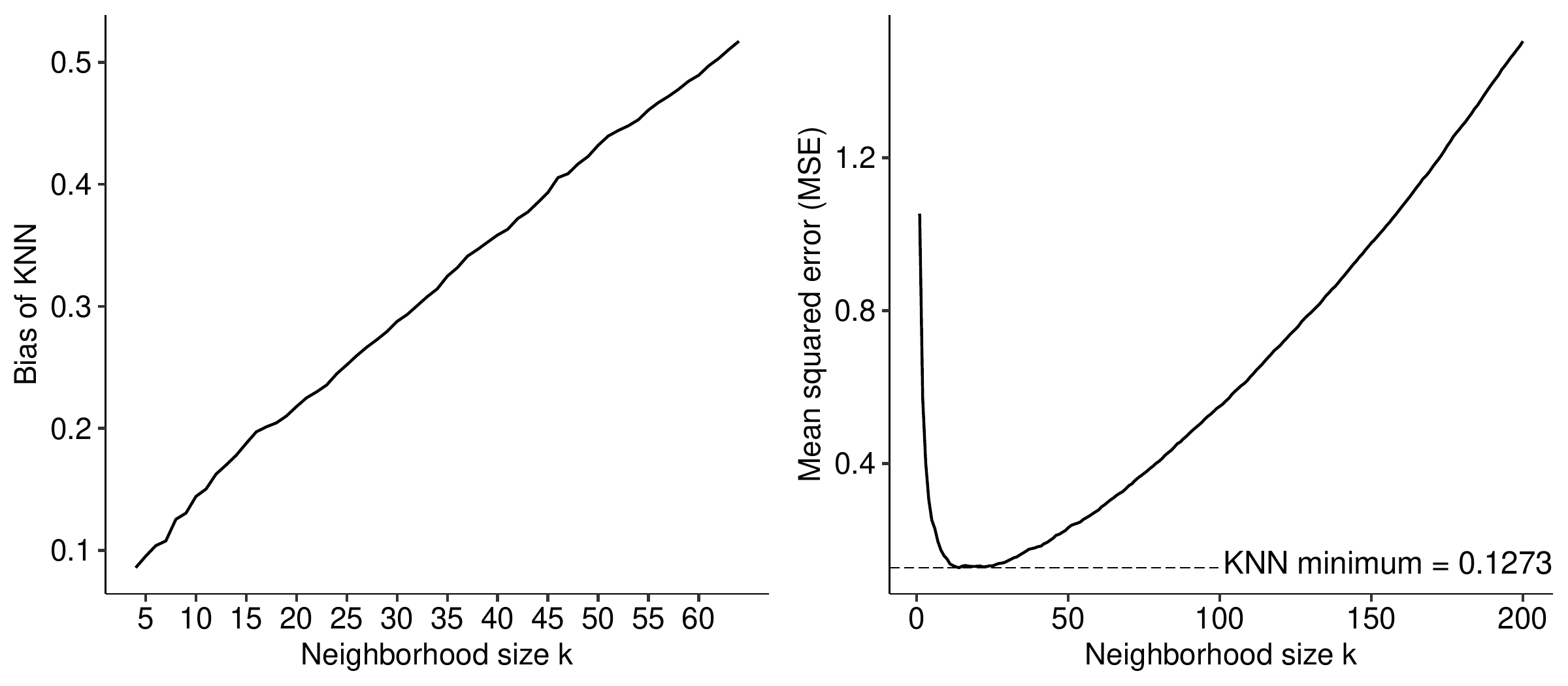}
		\caption{The bias and MSE results for $k$-NN 
			in Section~\ref{subsec:TDNN vs. DNN}.  
		}
		\label{fig:biasMSEKNN}
	\end{figure}

	We repeat the same simulation study as in Section \ref{subsec:TDNN vs. DNN} of the main text using the $k$-NN estimator by varying the neighborhood size $k$ from 1 to 200. The performance of the $k$-NN estimator is shown in Figure~\ref{fig:biasMSEKNN}.
	From 
	Figure~\ref{fig:biasMSEKNN}, we see that the finite-sample bias of $k$-NN tends to increase with the neighborhood size $k$, which is sensible since moving further away from the fixed test point incurs naturally inflated bias. 
	The MSE plot in Figure~\ref{fig:biasMSEKNN} shows a similar U-shaped pattern of the bias-variance tradeoff. 
	In contrast, the minimum value of the MSE attained by $k$-NN is 0.1273, which is outperformed by both the single-scale DNN and TDNN.  
	
	\begin{table}[htp]
 	    \resizebox{0.9\textwidth}{!}{\begin{minipage}{\textwidth}
 			\centering
			\vskip0.4cm
			\tabcolsep 0.1cm
 			
\begin{tabular}{lcccccc}
\toprule
\multicolumn{1}{c}{ } & \multicolumn{3}{c}{Fixed Test Point} & \multicolumn{3}{c}{Random Test Points} \\
\cmidrule(l{3pt}r{3pt}){2-4} \cmidrule(l{3pt}r{3pt}){5-7}
Method & MSE & Bias$^2$ & Variance & MSE & Bias$^2$ & Variance\\
\midrule
DNN & 0.0402 & 0.0038 & 0.0270 & 0.1337 & 0.0815 & 0.0404\\
$k$-NN & 0.0488 & 0.0024 & 0.0470 & 0.1826 & 0.1305 & 0.0499\\
TDNN & 0.0259 & 0.0005 & 0.0252 & 0.1284 & 0.0388 & 0.0649\\
\bottomrule
\end{tabular}

 		\end{minipage}}
	    \caption{A modified version of the comparison of DNN, $k$-NN, and TDNN in simulation setting \ref{setting_1} as described in Section~\ref{subsec:comparisonswithcausalforest}, but with the random feature vector $\bX$ drawn from $U([0, 1]^3)$ instead of $N(0, I_3)$.
	    }
	    \label{table:simsetting_1_unif}
	\end{table}
	
	\subsection{Simulation setting~\ref{setting_1} with uniform design}\label{sup:sec:CF}
	
	We repeat simulation setting \ref{setting_1} as described in Section \ref{subsec:comparisonswithcausalforest}, but now with random feature vector $\bX \sim U([0,1]^{p})$ as opposed to the Gaussian design used in the original model setting. All the parameter settings stay the same as in Section \ref{subsec:comparisonswithcausalforest}. From the results in Table \ref{table:simsetting_1_unif}, we can see that TDNN improve substantially over both DNN and $k$-NN. Moreover, compared to the results in Table \ref{table:simsetting_1_normal} under the Gaussian design, the average MSEs for random test points under the uniform design are now much smaller and closer to the MSE for the fixed test point.
	
\subsection{Simulation setting \ref{setting_3} for HTE estimation and inference} \label{sec:setting3}
The first two simulation examples in Section \ref{subsec:comparisonswithcausalforest} demonstrate the estimation accuracy of TDNN for general nonparametric regression and the third one will focus on the heterogeneous treatment effect (HTE) estimation and inference with the confidence interval coverage. 
We use a modified version of the second simulation setting for causal inference in \cite{wager2018estimation}. 

\begin{setting} \label{setting_3}
	Assume that the treatment propensity $e(\mathbf{x}) = 0.5$, the main effect $m(\mathbf{x}) = \frac{1}{8}(x_{1} - 1)$ for the control group, and the treatment effect $\tau(\mathbf{x}) = \varsigma(x_{1}) \varsigma(x_{2}) \varsigma(x_{3})$  with  $ \varsigma(x) = 1 + \{1 + \exp(-20(x - \frac{1}{3}))\}^{-1}$ for the treatment group, where $\mathbf{x} = (x_1, \cdots, x_p)^T$. Further assume that the feature vector  $\bX \sim U([0,1]^{p})$ and the regression error $\epsilon \sim N(0, 1)$ independent of $\bX$ for both groups. We increase the ambient dimensionality $p$ along the sequence $\{ 3,5,10,15,20 \}$.
\end{setting} 

\begin{table}[htp]
	\resizebox{0.9\textwidth}{!}{\begin{minipage}{\textwidth}
			\centering
			\vskip 0.4cm
			\tabcolsep 0.1cm
			
\begin{tabular}{lc@{\hskip 0.1in}ccccc@{\hskip 0.2in}ccccc}
\toprule
\multicolumn{2}{c}{ } & \multicolumn{5}{c}{Fixed Test Point} & \multicolumn{5}{c}{Random Test Points} \\
\cmidrule(l{3pt}r{3pt}){3-7} \cmidrule(l{3pt}r{3pt}){8-12}
Method & p & MSE & Bias$^2$ & Variance & Coverage & Width & MSE & Bias$^2$ & Variance & Coverage & Width\\
\midrule
DNN & 3 & 0.1511 & 0.0414 & 0.0977 & 0.816 & 1.1541 & 0.3152 & 0.1580 & 0.1066 & 0.6727 & 1.2215\\
$k$-NN & 3 & 0.1269 & 0.0517 & 0.0756 & 0.856 & 1.0702 & 0.3916 & 0.3130 & 0.0733 & 0.5340 & 1.0511\\
TDNN & 3 & 0.0899 & 0.0145 & 0.0836 & 0.948 & 1.1236 & 0.3022 & 0.0672 & 0.1670 & 0.8196 & 1.5124\\
\hline
DNN & 5 & 0.1706 & 0.0430 & 0.0967 & 0.801 & 1.1551 & 0.3204 & 0.1612 & 0.1061 & 0.6707 & 1.2188\\
$k$-NN & 5 & 0.1320 & 0.0560 & 0.0752 & 0.852 & 1.0676 & 0.4013 & 0.3208 & 0.0731 & 0.5262 & 1.0499\\
TDNN & 5 & 0.1008 & 0.0168 & 0.0833 & 0.915 & 1.1209 & 0.3063 & 0.0704 & 0.1668 & 0.8162 & 1.5112\\
\hline
DNN & 10 & 0.1600 & 0.0364 & 0.0987 & 0.833 & 1.1647 & 0.3337 & 0.1718 & 0.1083 & 0.6635 & 1.2305\\
$k$-NN & 10 & 0.1302 & 0.0489 & 0.0780 & 0.869 & 1.0866 & 0.4154 & 0.3325 & 0.0750 & 0.5251 & 1.0627\\
TDNN & 10 & 0.1014 & 0.0113 & 0.0852 & 0.934 & 1.1318 & 0.3174 & 0.0764 & 0.1722 & 0.8143 & 1.5336\\
\hline
DNN & 15 & 0.1687 & 0.0313 & 0.1019 & 0.825 & 1.1808 & 0.3428 & 0.1782 & 0.1093 & 0.6608 & 1.2361\\
$k$-NN & 15 & 0.1287 & 0.0500 & 0.0782 & 0.872 & 1.0868 & 0.4291 & 0.3427 & 0.0759 & 0.5201 & 1.0682\\
TDNN & 15 & 0.1021 & 0.0109 & 0.0888 & 0.923 & 1.1536 & 0.3237 & 0.0791 & 0.1746 & 0.8124 & 1.5445\\
\hline
DNN & 20 & 0.1628 & 0.0382 & 0.0985 & 0.820 & 1.1669 & 0.3394 & 0.1757 & 0.1094 & 0.6642 & 1.2368\\
$k$-NN & 20 & 0.1330 & 0.0497 & 0.0798 & 0.877 & 1.0981 & 0.4232 & 0.3366 & 0.0764 & 0.5248 & 1.0721\\
TDNN & 20 & 0.1061 & 0.0125 & 0.0892 & 0.927 & 1.1564 & 0.3215 & 0.0772 & 0.1748 & 0.8144 & 1.5464\\
\bottomrule
\end{tabular}

	\end{minipage}}
	
	\caption{Comparison of DNN, $k$-NN, and TDNN in  simulation setting \ref{setting_3} described in Section~\ref{sec:setting3}. 
	}
	\label{table:simsetting3}
\end{table}

As with simulation setting~\ref{setting_2}, we evaluate the performance of the three nonparametric learning and inference methods at a fixed test point chosen as $x_{1} = 0.2$, $x_{2} = 0.4$, $x_{3} = 0.6$, and $x_{j} = 0.5$ for $j > 3$ as well as for a set of 100 test points randomly drawn from the hypercube $[0,1]^p$.  For the TDNN estimator, the ratio $c = s_{2}/s_{1}$ { is chosen from the sequence $\{2,4,6,8,10,15,20,25,30\}$ for random test points and we fix $c = 2$ for the fixed test point for simplicity. The subsampling scale $s_1$ is chosen from the interval $[s_{\text{sign}}, 2s_{\text{sign}}]$ for each given $c$, where $s_{\text{sign}}$ is given by the sign-change tuning process introduced at the beginning of Section \ref{sec:SimulationStudies}.} We apply the TDNN estimator to the treatment group and control group separately, and then take the difference between the TDNN estimators for the two groups to estimate the HTE. In addition, we also report the coverage probability of $95\%$ confidence intervals for the HTE constructed based on the asymptotic normality results established in Section \ref{sec:HTE}. The DNN and $k$-NN estimators are similarly applied for estimation and inference of the HTE. In particular, we see from the results in Table \ref{table:simsetting3} that the TDNN estimator indeed provides lower MSEs for HTE estimation and valid confidence intervals for HTE inference with higher coverage compared to the DNN and $k$-NN estimators.

	\renewcommand{\thesubsection}{D.\arabic{subsection}}
	
	\section{Proofs of main results} \label{Appen} 
	
	\subsection{Proof of Theorem \ref{thm: bias}} \label{SecA.1}
	
	Let us investigate the higher-order asymptotic expansion for the bias term of the single-scale distributional nearest neighbors (DNN) estimator $D_n (s) ({\bf x })$ introduced in (\ref{neweq.FL001}) under the asymptotic setting when the subsampling scale $s \rightarrow \infty$ as the sample size $n$ increases. Recall that the target point $\bx$ is a given vector inside the domain $\supp(\bX) \subset \mathbb{R}^d$ of the covariate distribution, where the feature dimensionality $d$ is assumed to be fixed for simplifying the technical presentation of our work. The main idea of the proof is to first consider the specific case of $s = n$ in Lemma \ref{lemma1} in Section \ref{SecB.5}, and then analyze the general case of $s \rightarrow \infty$ by exploiting the projection of the mean function $\mu(\bX) = \mathbb{E}(Y | \bX)$ onto the positive half line $\mathbb{R}_+ = [0, \infty)$ given by $\|\bX - \bx\|$ in Lemma \ref{lem2} in Section \ref{SecB.6}.
	
	Since $\{i_1,\cdots, i_s\}$ is a random subsample of $\{1,\cdots, n\}$ with subsampling scale $s$, in view of (\ref{neweq.FL002}) and (\ref{neweq.FL001}) we have
	\begin{align} \label{neweq.FL003}
		\mathbb{E} \, D_n(s) (\bx) 
		& = \mathbb{E} \, \Phi(\bx; \bZ_{i_1}, \bZ_{i_2}, \ldots, \bZ_{i_s}) \nonumber\\
		& = \mathbb{E} \, [Y_{(1)}(\bZ_{i_1}, \bZ_{i_2}, \ldots, \bZ_{i_s})] \nonumber \\
		& = \mathbb{E} \, [m(r_{(1)})(\bZ_{i_1}, \bZ_{i_2}, \ldots, \bZ_{i_s})],
	\end{align}
	where the kernel $\Phi(\bx; \cdot)$ in the U-statistic representation of the DNN estimator is simply the 1-nearest neighbor (1NN) estimator $Y_{(1)}(\cdot)$ given by the response for the closest neighbor $\bX_{i_{(1)}}$ of $\bx$ in the random subsample $\{\bX_{i_1},\cdots, \bX_{i_s}\}$ with $\bZ_{i_j}$ denoting $(\bX_{i_j}, Y_{i_j})$, $m(r) = \mathbb{E}(Y \, | \, \|\bX - \bx\| = r)$ is the projection of the mean function $\mu(\bX)$ onto the positive half line introduced in (\ref{neweq.FL004}) in Lemma \ref{lem2}, and $r_{(1)} = \|\bX_{i_{(1)}} - \bx\|$. The representation in (\ref{neweq.FL003}) provides a useful starting point for our technical analysis.
	
	From (\ref{neweq.FL003}) above, we see that it is necessary to first study the asymptotic behavior of term $r_{(1)}$. Without loss of generality, for this step we can simply replace parameter $s$ with parameter $n$ since both subsample size $s$ and full sample size $n$ are assumed to diverge simultaneously. With such a notational simplification, the 1NN $\bX_{i_{(1)}}$ of $\bx$ in the subsample becomes the 1NN $\bX_{(1)}$ of $\bx$ in the full sample and thus $r_{(1)} = \|\bX_{(1)} - \bx\|$.  We see from Lemma \ref{lemma1} that $\mathbb{E} r_{(1)}^2 = \mathbb{E} \|\bX_{(1)} - \bx\|^2$ admits a higher-order asymptotic expansion with explicit constants provided for the first two leading orders,  which are  $n^{-2/d}$ and $n^{-4/d}$, respectively, for $d \geq 2$ as shown in \eqref{neweq.L4} and (\ref{eqn:lemma1-extra-order}),  and $n^{-2}$ and $n^{-3}$, respectively,  for $d = 1$ as shown in \eqref{eqn:lemma1-extra-order}.  To apply such an asymptotic expansion in Lemma \ref{lemma1} to the term $r_{(1)} = \|\bX_{i_{(1)}} - \bx\|$ in (\ref{neweq.FL003}), we now need to replace parameter $n$ back with parameter $s$, which also diverges by assumption. 
	
	A natural next step is to consider the expectation on the right-hand side of (\ref{neweq.FL003}) by conditioning on $r_{(1)} = \|\bX_{i_{(1)}} - \bx\|$. Indeed, this motivates us to investigate the higher-order asymptotic expansion of the projected mean function $m(r) = \mathbb{E}(Y \, | \, \|\bX - \bx\| = r)$ in Lemma \ref{lem2}, where $r \rightarrow 0$ and some constants are given for the first two leading orders $r^2$ and $r^4$ in (\ref{neweq.FL005}). {\color{black} Observe that the asymptotic regime of $r \rightarrow 0$ is reasonable since it has been shown by Lemma 2.2 in \cite{biau2015lectures} that $r_{(1)} = \|\bX_{i_{(1)}} - \bx\| \rightarrow 0 $ almost surely as $s \rightarrow \infty$. 
		
		Based on the expansion of $  \mathbb{E} \Vert \bX_{(1)} - \bx \Vert^2  $ under different regimes of $d$ provided in \eqref{neweq.L3}--\eqref{eqn:lemma1-extra-order} in Lemma \ref{lemma1}, we can see that there are two cases for the expansion of $ \mathbb{E} \Vert \bX_{(1)} - \bx \Vert^2  $. Specifically, the first two leading  orders are $ n ^{ - 2 } $ and $ n^{ - 3} $ for $ d = 1$, while the first two leading orders are $ n^{ - 2/d} $ and $ n^{ - 4/d} $ for $ d \geq 2 $. 
		Thus, we calculate $  \mathbb{E}  \, D_n(s) (\bx) $ for $ d \geq 2  $ and $ d = 1 $, separately. 
		
		First, for the case of $d = 1$, combining the arguments above using \eqref{neweq.L3} and Lemma \ref{lem2}, from (\ref{neweq.FL003}) we can deduce that 
		\begin{align} \label{neweq.L14}
			\mathbb{E} & \, D_n(s) (\bx)		 = \mu(\bx) + \frac{f(\bx)\tr(\mu''(\bx)) + 2 \, \mu'(\bx)^Tf'(\bx)}{2 \, d \, f(\bx)} \;  \mathbb{E} r_{(1)}^2 + O_4 \mathbb{E} r_{(1)}^4   \non
			& = \mu(\bx) + \frac{ f(\bx)\tr(\mu''(\bx)) + 2 \, \mu'(\bx)^Tf'(\bx)} {2 \, d \, f(\bx)} \non
			& \quad  \times \left(  \frac{\Gamma(2/d + 1)}{(f(\bx) V_d)^{2/d}} s^{-2/d} -  \left(  \frac { \Gamma (2/d + 2) } {d  (f ({\bf x}) V_d)^{2/d} }  \right) s^{ - ( 1 + 2/ d )  }  \right) \non
			& \quad+ O_4 \frac{\Gamma(4/d + 1)}{(f(\bx) V_d)^{4/d}} s^{-4/d} + o(s^{- ( 1 + 2 / d) }) \non
			& = \mu(\bx) 
			+ \Gamma(2/d + 1) \frac{f(\bx)\tr(\mu''(\bx)) + 2 \, \mu'(\bx)^Tf'(\bx)}{2 \, d \, V_d^{2/d} \, f(\bx)^{2/d + 1}}    s^{-2/d}  + R (s) , \end{align}
		where $ R (s) =  O(s^{- 3})  $. In addition, $\Gamma(\cdot)$ denotes the gamma function, $V_d = \frac{\pi^{d/2}}{\Gamma(1+d/2)}$,  $f'(\bx)$ and $\mu'(\bx)$ represent the first-order gradients of $f(\bx)$ and $\mu(\bx)$ at $\bx$, respectively, $\mu''(\bx)$ denotes the Hessian matrix of $\mu(\cdot)$ at $\bx$, $ O_4 $ is some constant given in Lemma \ref{lem2}, and $\tr(\cdot)$ stands for the trace operator.
		
		We proceed to prove for the case of $ d \geq 2 $. In the same fashion of deriving \eqref{neweq.L14}, applying \eqref{neweq.L4}--\eqref{eqn:lemma1-extra-order} and Lemma \ref{lem2}, from \eqref{neweq.FL003} we can obtain that 
		\begin{align} \label{neweq.FL006}
			\mathbb{E} & \, D_n(s) (\bx)		 = \mu(\bx) + \frac{f(\bx)\tr(\mu''(\bx)) + 2 \, \mu'(\bx)^Tf'(\bx)}{2 \, d \, f(\bx)} \;  \mathbb{E} r_{(1)}^2 + O_4 \mathbb{E} r_{(1)}^4   \non
			& = \mu(\bx) + \frac{ f(\bx)\tr(\mu''(\bx)) + 2 \, \mu'(\bx)^Tf'(\bx)} {2 \, d \, f(\bx)} \times \left( \frac{\Gamma(2/d + 1)}{(f(\bx) V_d)^{2/d}} s^{-2/d} - C (d, f, \mu, \bx) s^{-4/d}  \right) \non
			& \quad+ O_4 \frac{\Gamma(4/d + 1)}{(f(\bx) V_d)^{4/d}} s^{-4/d} + o(s^{-4/d}) \non
			& = \mu(\bx) 
			+ \Gamma(2/d + 1) \frac{f(\bx)\tr(\mu''(\bx)) + 2 \, \mu'(\bx)^Tf'(\bx)}{2 \, d \, V_d^{2/d} \, f(\bx)^{2/d + 1}}    s^{-2/d}  + R (s)  ,
		\end{align}
		where $ R (s) = O(s^{-4/d})  $.
		
		Therefore, combining the above results, we obtain the desired higher-order asymptotic expansion for the bias term of the single-scale DNN estimator $B(s) = \mathbb{E} \, D_n (s) ({\bf x }) - \mu(\bx)$. This completes the proof of Theorem \ref{thm: bias}.

	}
	
	\subsection{Proof of Theorem \ref{thm: DNN-normality}} \label{SecA.2}
	
	We now proceed to prove the asymptotic normality of the single-scale DNN estimator $D_n (s) ({\bf x })$. Recall that in Theorem \ref{thm: bias}, the higher-order asymptotic expansion for the bias term $B(s) $ of $D_n (s) ({\bf x })$ requires the assumption that the subsampling scale $s \rightarrow \infty$ as sample size $n$ increases. As shown in the proof of Theorem \ref{thm: bias} in Section \ref{SecA.1}, the single-scale DNN estimator $D_n (s) ({\bf x })$ reduces to the 1NN estimator when we choose $s = n$, since in such a case, there is a single subsample with size $s = n$, i.e., the full sample. We immediately realize that although the choice of $s = n$ satisfies the need on the bias side, it does not make the variance shrink asymptotically. Intuitively, we would need to form the empirical average over a diverging number of such individual estimates in order to establish the desired asymptotic normality. This naturally calls for the assumption of $s = o(n)$, which entails that the total number of these individual estimates ${n \choose s}$ diverges as sample size $n$ increases. Thus we will work with the asymptotic regime of subsampling scale with $s \rightarrow \infty$ and $s = o(n)$.
	
	In view of the U-statistic representation of $D_n (s) ({\bf x })$ given in (\ref{neweq.FL001}), a natural idea of the proof for the asymptotic normality of the single-scale DNN estimator is to exploit the asymptotic theory of the U-statistic framework. However, the classical U-statistic asymptotic theory is not readily applicable due to the common assumption of \textit{fixed} subsampling scale $s$. In contrast, as discussed above, our asymptotic analysis needs the opposite assumption of \textit{diverging} subsampling scale $s$, i.e., $s \rightarrow \infty$. Such a discrepancy causes additional technical challenges when we derive the asymptotic normality. 
	
	
	Let us first exploit Hoeffding's canonical decomposition introduced in \cite{Hoeffding_A_1948}, which is an extension of the projection idea. For each $ 1 \leq i \leq s $, we define the centered conditional expectation
	\begin{align} \label{neweq.FL007}
		\widetilde{\Phi}_i ({\bf x}; {\bf z}_1, \ldots, {\bf z}_i ) & = \e [ \Phi ({\bf x}; {\bf z}_1, \ldots, {\bf z}_i, {\bf Z}_{i + 1}, \ldots, {\bf Z}_{s} ) \, | \, {\bf z}_1, \ldots, {\bf z}_i] \non
		&\quad- \e \Phi ({\bf x}; {\bf Z}_1, \ldots, {\bf Z}_s ),
	\end{align}
	where $\Phi(\bx; \cdot)$ is the kernel defined in (\ref{neweq.FL002}) for the U-statistic representation of the single-scale DNN estimator. Then in light of  (\ref{neweq.FL007}), for each $ 1 \leq i \leq s $ we can successively define the canonical term 
	\begin{equation} \label{neweq.FL008}
		g_i ( {\bf x}; {\bf z}_1, \ldots, {\bf z}_i )  = \widetilde{\Phi}_i ({\bf x}; {\bf z}_{1}, \ldots, {\bf z}_i)  -  \sum_{j = 1}^{i - 1} \sum_{ 1 \leq \alpha_1 < \ldots < \alpha_j \leq i } g_j ({\bf x}; {\bf z}_{\alpha_1}, \ldots, {\bf z}_{\alpha_j}),
	\end{equation}
	where $g_1 ( {\bf x}; {\bf z}_1)  = \widetilde{\Phi}_1 ({\bf x}; {\bf z}_{1})$ by definition. Combining (\ref{neweq.FL002}), (\ref{neweq.FL007}), and (\ref{neweq.FL008}), we see that the kernel $\Phi(\bx; \cdot)$ can be rewritten as a sum of the canonical terms
	\begin{equation} \label{decom-phi}
		\Phi ({\bf x}; {\bf Z}_1, \ldots, {\bf Z}_s ) -  \e \Phi ({\bf x}; {\bf Z}_1, \ldots, {\bf Z}_s ) = \sum_{j = 1}^s \sum_{ 1 \leq \alpha_1 < \ldots < \alpha_j \leq s} g_j ( {\bf x}; {\bf Z}_{\alpha_1 }, \ldots, {\bf Z}_{\alpha_j} ).  
	\end{equation}
	Moreover, it holds that
	\begin{equation} \label{decom-varphi}
		\Var( \Phi ( {\bf x} ; {\bf Z}_{1}, \ldots, {\bf Z}_s ))  = \sum_{j = 1}^s {s \choose j } \Var ( g_j ( {\bf x} ; {\bf Z}_{1}, \ldots, {\bf Z}_j ) ).
	\end{equation}
	The above Hoeffding's canonical decomposition in (\ref{decom-phi}) plays an important role in establishing the asymptotic normality. 
	
	In view of (\ref{neweq.FL001}), (\ref{neweq.FL007}), and (\ref{decom-phi}), we can deduce that 
	\begin{align} \label{neweq.FL009}
		& D_n(s)  - \mathbb{E} \, D_n(s)  = \binom{n}{s}^{-1} \sum_{1 \le i_1<i_2< \ldots < i_s \le n} \widetilde \Phi_s({\bf x} ;\bZ_{i_1}, \bZ_{i_2}, \ldots, \bZ_{i_s}) \non
		& = \binom{n}{s}^{-1} \Big\{ \binom{n-1}{s-1} \sum_{i_1=1}^n g_1({\bf x} ;\bZ_{i_1}) + \binom{n-2}{s-2} \sum_{1 \le i_1 < i_2 \le n} g_2({\bf x} ;\bZ_{i_1}, \bZ_{i_2})  + \ldots \non
		& \quad + \binom{n-s}{s-s} \sum_{1 \le i_1 < i_2 < \ldots < i_s \le n} g_s({\bf x} ;\bZ_{i_1}, \bZ_{i_2}, \ldots, \bZ_{i_s}) \Big \}.
	\end{align}
	From the above Hoeffding's canonical decomposition in (\ref{neweq.FL009}) for the single-scale DNN estimator, we see that the H{\'a}jek projection introduced in \cite{Hjek_Asymptotic_1968} of the centered DNN estimator $D_n(s) - \mathbb{E} D_n(s)$ is given by 
	\begin{equation} \label{neweq.FL010}
		\widehat D_n(s) = \binom{n}{s}^{-1} \binom{n-1}{s-1} \sum_{i=1}^{n} g_1({\bf x} ;\bZ_i),
	\end{equation}
	which is the first-order part of the decomposition in (\ref{neweq.FL009}). 
	
	A useful observation is that the H{\'a}jek projection given in (\ref{neweq.FL010}) involves the sum of some independent and identically distributed (i.i.d.) terms. Denote by $\sigma_n^2$ the variance of the H{\'a}jek projection. Then it follows from $g_1 ( {\bf x}; {\bf z}_1)  = \widetilde{\Phi}_1 ({\bf x}; {\bf z}_{1})$ and (\ref{neweq.FL007}) that 
	\begin{align} \label{neweq.FL011}
		\sigma_n^2 & = \Var  (\widehat D_n(s)) = \frac{s^2}{n}   \Var  (\widetilde{\Phi}_1(\bx; \bZ_1)) \non 
		&= \frac{s^2}{n}   \Var  (\Phi_1(\bx; \bZ_1)) = \frac{s^2}{n} \eta_1,
	\end{align}
	where the non-centered conditional expectation $\Phi_1(\bx; \bZ_1)$ is defined later in (\ref{neweq.FL012}) and $\eta_1$ is defined as the variance of $\Phi_1(\bx; \bZ_1)$. From (\ref{neweq.FL007}), we see that each term $g_1 ( {\bf x}; {\bf Z}_i)  = \widetilde{\Phi}_1 ({\bf x}; {\bf Z}_{i})$ of the i.i.d. sum in (\ref{neweq.FL010}) has zero mean. Thus by (\ref{neweq.FL011}), an application of the Lindeberg--L\'evy central limit theorem in \cite{Borovkov_Probability_2013} leads to 
	\begin{equation} \label{neweq.FL013}
		\frac{\widehat D_n(s)}{\sigma_n} \toD N(0,1),
	\end{equation}
	which establishes the asymptotic normality of the H{\'a}jek projection $\widehat D_n(s)$.
	
	Finally, we aim to show that similar asymptotic normality as above holds when the H{\'a}jek projection $\widehat D_n(s)$ in the numerator on the left-hand side of (\ref{neweq.FL013}) is replaced with the centered single-scale DNN estimator $D_n (s) ({\bf x }) - \mathbb{E} \, D_n (s) ({\bf x }) = D_n (s) ({\bf x })  - \mu(\bx) - B(s)$, where $B(s)$ is the bias term identified in Theorem \ref{thm: bias}. With the aid of Slutsky's lemma, we see that it suffices to show that 
	\begin{equation} \label{neweq.FL014}
		\frac{D_n(s) - \mathbb{E}D_n(s) - \widehat D_n(s)}{\sigma_n} =o_P(1). 
	\end{equation}
		Following Lemma 3.3 in \cite{wager2018estimation} and replacing ``tree in forest" with ``kernel in U-statistics" in the proof, we can easily see that	
	\begin{align} \label{remainder}
		\mathbb{E} & [D_n(s) - \mathbb{E}D_n(s) - \widehat D_n(s)]^2 		 \le \frac{s^2}{n^2} \Var ( \Phi ( {\bf x} ; {\bf Z}_{1}, \ldots, {\bf Z}_s )).
	\end{align}
	It remains to bound the variance term $\Var ( \Phi ( {\bf x} ; {\bf Z}_{1}, \ldots, {\bf Z}_s ))$ above.
	
	By Lemma \ref{new.lem.3} in Section \ref{SecB.7}, we have an important result that
	\begin{equation} \label{neweq.FL015}
		\Var ( \Phi ( {\bf x} ; {\bf Z}_{1}, \ldots, {\bf Z}_s )) = o(n \eta_1).
	\end{equation}
	Combining (\ref{neweq.FL011}), (\ref{remainder}), and (\ref{neweq.FL015}), it holds that
	\begin{align} \label{neweq.FL016}
		\mathbb{E} & \left[\frac{D_n(s) - \mathbb{E}D_n(s) - \widehat D_n(s))}{\sigma_n}\right]^2 = o\left\{\frac{1}{\sigma_n^2}\frac{s^2}{n^2} (n \eta_1)\right\}\non
		&= o\left\{\frac{n}{s^2 \eta_1}\frac{s^2}{n^2} (n \eta_1)\right\} = o(1).
	\end{align}
	Therefore, we are ready to see that (\ref{neweq.FL016}) entails the desired claim (\ref{neweq.FL014}). Finally, by (\ref{neweq.FL011}) and (\ref{neweq.FL017}) obtained in the proof of Lemma \ref{new.lem.3} in Section \ref{SecB.7}, we see that $\sigma_n$ is of order $(s/n)^{1/2}$, which concludes the proof of Theorem \ref{thm: DNN-normality}.

	\subsection{Proof of Theorem \ref{thm1}} \label{SecA.3}
	
	We further prove the asymptotic normality of the two-scale DNN estimator $ D_n (s_1, s_2) (\bf x) $ introduced in (\ref{TDNN}). It is worth mentioning that Theorem \ref{thm1} is not a simple consequence of Theorem \ref{thm: DNN-normality} since the marginal asymptotic normalities do not necessarily lead to the joint asymptotic normality. This means that we need to analyze the two single-scale DNN estimators involved in the definition of the two-scale DNN estimator in a joint fashion. To this end, we will exploit the ideas in the proof of Theorem \ref{thm: DNN-normality} in Section \ref{SecA.2}. To facilitate the technical analysis, some key technical tools are provided in Lemmas \ref{le-U-pre}--\ref{lemma-eta} in Sections \ref{Sec.B5}--\ref{SecB.10}, respectively. 
	
	Without loss of generality, let us assume that $s_1 < s_2$ for the two subsampling scales. In particular, we make the assumptions that $s_1, s_2 \rightarrow \infty$, $s_1, s_2 = o(n)$, and $c_1 \leq s_1/s_2 \leq c_2$ for some constants $0 < c_1 < c_2 < 1$. From Lemma \ref{le-U-pre} in Section \ref{Sec.B5}, we see that the two-scale DNN estimator $ D_n (s_1, s_2) (\bf x) $ is also a U-statistic of order $s_2$ with a new kernel $\Phi^{*} ( {\bf x}; {\bf Z}_1, {\bf Z}_2, \ldots,  {\bf Z}_{s_2} )$ introduced later in (\ref{phistar}). Thus Hoeffding's canonical decomposition for U-statistics can be applied to derive the asymptotic normality of the two-scale DNN estimator. For each $ 1 \leq i \leq s_2 $, let us define
	\begin{equation} \label{neweq.FL018}
		\Phi^*_i ({\bf x}; {\bf z}_1, \ldots, {\bf z}_i )= \e [ \Phi ^*( {\bf x}; {\bf z}_1, \ldots, {\bf z}_{i}, {\bf Z}_{i + 1}, \ldots, {\bf Z}_{s_2} )  \, | \, {\bf z}_1, \ldots, {\bf z}_{i}],
	\end{equation}
	\begin{align} \label{neweq.FL019}
		g_i^* (  {\bf x};{\bf z}_1, \ldots, {\bf z}_i ) & = \Phi_i^* ( {\bf x}; {\bf z}_1, \ldots, {\bf z}_i ) - \e \Phi_i^* ( {\bf x}; {\bf Z}_1, \ldots, {\bf Z}_i ) \non
		&\quad- \sum_{j = 1}^{i - 1} \sum_{1 \leq \alpha_1 < \ldots < \alpha_{j} \leq i} g_j^* (  {\bf x};{\bf z}_{\alpha_1}, \ldots, {\bf z}_{\alpha_j} ),
	\end{align}
	where $g_1^* ( {\bf x}; {\bf z}_1) = \Phi_1^* ( {\bf x}; {\bf z}_1) - \e \Phi_1^* ( {\bf x}; {\bf Z}_1)$ by definition. We further define
	\begin{equation} \label{neweq.FL020}
		\Var \Phi^* = \Var (\Phi^* ({\bf x}; {\bf Z}_1, {\bf Z}_2, \ldots, {\bf Z}_{s_2})) \ \text{ and } \ \eta_1^* = \Var (\Phi^*_1 ( {\bf x}; {\bf Z}_1 )). 
	\end{equation}
	
	In view of (\ref{TDNN}), (\ref{neweq.FL018}), and (\ref{neweq.FL019}), an application of similar U-statistic and Hoeffding's canonical decomposition arguments to those in the proof of Theorem \ref{thm: DNN-normality} in Section \ref{SecA.2} entails that
	\begin{equation} \label{neweq.FL021}
		( n^{-1} s_2^2 \eta_1^* )^{-1 /2}  \big( D_n (s_1, s_2) ({\bf x }) - \e [ D_n (s_1, s_2) (\bf x) ] \big)
	\end{equation}
	can be approximated by the first-order part of Hoeffding's canonical decomposition that converges to a normal distribution with the remainders asymptotically negligible, where $\eta_1^*$ is given in (\ref{neweq.FL020}). More specifically, denote by
	\begin{equation} \label{neweq.FL022}
		\widehat{D}_n (s_1, s_2) =  \frac { s_2  } { n  }  \sum_{i = 1}^n g_1^* ( {\bf x}; {\bf Z}_i ),
	\end{equation}
	where $g_1^* ( {\bf x}; {\bf Z}_i)$ is defined in (\ref{neweq.FL019}). It follows from (\ref{neweq.FL020}), (\ref{neweq.FL022}), and the classical central limit theorem for i.i.d. random variables that 
	\begin{equation} \label{neweq.FL023}
		\frac { \widehat{D}_n (s_1, s_2) } { \sqrt{ n^{-1} s_2^2  \eta_1^* } } \toD N (0, 1),
	\end{equation}
	since it holds that $\Var(g_1^* ( {\bf x}; {\bf Z}_1 )) = \Var(\Phi^*_1 ( {\bf x}; {\bf Z}_1 )) = \eta_1^*$.

	Similar to \eqref{remainder}, by (\ref{neweq.FL019}), (\ref{neweq.FL020}), and (\ref{neweq.FL022}) we can deduce that
	\begin{align} \label{varphi*}
		& \frac {  \e \big[  D_n (s_1, s_2) ({\bf x }) -  \e  D_n (s_1, s_2) ({\bf x })  - \widehat{D}_n (s_1, s_2) \big]^2 } { n^{-1} s_2 ^2 \eta_1^* }  \non
		&\quad\leq  \frac { n^{-2} s_2^2 \Var \Phi^*  } { n^{-1} s_2^2 \eta_1^*  } =  \frac { \Var \Phi^*  } { n \eta_1^* }.   
	\end{align}
	Moreover, it follows from the upper bound on $\Var \Phi^*$ obtained in Lemma \ref{lemma-phi} in Section \ref{SecB.9} and the asymptotic order of $\eta_1^*$ established in Lemma \ref{lemma-eta} in Section \ref{SecB.10} that 
	\begin{equation} \label{neweq.FL024}
		\Var \Phi^* / (n \eta_1^*) \to 0 
	\end{equation}
	since $ s_2 / n \to 0  $ by assumption. Therefore, combining (\ref{neweq.FL023})--(\ref{neweq.FL024}), an application of Slutsky's lemma yields the desired claim in (\ref{neweq.FL021}), that is, 
	\begin{equation} \label{neweq.FL027}
		\frac { D_n (s_1, s_2) ({\bf x}) - \e  D_n (s_1, s_2) ({\bf x}) } { \sigma_n } \toD  N (0, 1),
	\end{equation} 
	where we define $ \sigma_n^2 = n^{-1} s_2^2 \eta_1^*$. Finally, we see from Lemma \ref{lemma-eta} that $\sigma_n = (n^{-1} s_2^2 \eta_1^*)^{1/2}$ is of order $(s_2/n)^{1/2}$, and from the higher-order asymptotic expansion of the bias term in Theorem \ref{thm: bias} that {\color{black}
		\begin{equation*}
			\Lambda = \e  D_n (s_1, s_2) ({\bf x}) - \mu(\bx) = \left\{ 
			\begin{aligned}
				& O(s_1^{-4/d} + s_2^{-4/d}) , \quad & d \geq 2, \\
				& O (s_1^{ - 3} + s_2 ^{ - 3}), \quad  & d = 1.
			\end{aligned}
			\right.
		\end{equation*}
	}
	This together with (\ref{neweq.FL027}) completes the proof of Theorem \ref{thm1}.
	
	\subsection{Proof of Theorem \ref{thm-MSE}} \label{SecA.41}
	
	 The main idea of the proof is to apply the bias-variance decomposition for the mean-squared error. Recall that $ D_n (s_1, s_2) (\bx) = w_1^* D_n (s_1) (\bx) + w_2^* D_n (s_2) (\bx) $ and 
 $$ \e [ D_n (s_1, s_2) (\bx) ] =  w_1^* \e [ \mu (\bX_{(1)} (s_1)) ]  + w_2^* \e [ \mu (\bX_{(1)} (s_2)) ],  $$
 where $\bX_{(1)} (s_1) = \bX_{(1)} (\bX_1, \ldots, \bX_{s_1})  $ denotes the 1-nearest neighbor of $\bx$ among $\{\bX_1, \ldots, \bX_{s_1}\}$ and similarly, $\bX_{(1)} (s_2) = \bX_{(1)} (\bX_1, \ldots, \bX_{s_2}) $.
 Then we have the bias-variance decomposition
\begin{equation}
	\begin{split}
	    &  \e \big(  [ D_{n} (s_1, s_2) (\bx) - \mu (\bx)  ]^2 \big)  \\
	    &  = \e \Big\{ \big( D_n (s_1, s_2) (\bx) - w_1^* \e [ \mu (\bX_{(1)} (s_1)) ] - w_2^*  \e [ \mu (\bX_{(1)} (s_2)) ]  \big)^2 \Big\}\\
	    & \quad + \big[  \e ( D_n (s_1, s_2) (\bx) ) - \mu (\bx)  \big]^2  \\
	    &:= I_1 (\bx) + I_2(\bx).
	 \end{split}
\end{equation}
Let us first deal with the bias term $I_2 (\bx)$. Using the similar arguments to those in the proofs of Lemmas \ref{lemma1} and \ref{lem2}, we can deduce that 
\[  I_2 (\bx) \leq  \left\{
\begin{array}{ll}	 
	\frac { R_1^2 (\bx, d, f, \mu) } { (c - 1)^2 }    c^{-1} s_2^{- 6 }, & d = 1, \\
	\\
   	\frac { R_2^2 (\bx, d, f, \mu) } { (c - 1)^2 }   c^{-2} s_2^{ - 8/ d}, &  d \geq 2, \\ 
\end{array} 
\right. \]
where $R_1 (\bx, d, f, \mu)$ and $R_2 (\bx, d, f, \mu)$ are some constants depending on the bounds for the first four derivatives of $f(\cdot)$ and $\mu(\cdot)$ in a neighborhood of $\bx$.

We now analyze the variance term $ I_1 (\bx) $. It holds that 
\begin{equation}
	\begin{split}
		 I_1 (\bx) &\leq (w_1^*)^2 \e \bigg\{ {n \choose s_1}^{-1} \sum_{1 \leq i_1 <  i_2< \cdots < i_{s_1} \leq n}  \Big( Y_{(1)} ( \bZ_{i_1}, \ldots, \bZ_{i_{s_1}}) -  \e \mu ( \bX_{(1)} ( s_1 ) ) \Big)
		 \bigg\} ^2 \\
		 & +  (w_2^*)^2 \e \bigg\{ {n \choose s_2}^{-1} \sum_{1 \leq i_1 <  i_2< \cdots <  i_{s_2} \leq n}  \Big( Y_{(1)} ( \bZ_{i_1}, \ldots, \bZ_{i_{s_2}}) -  \e \mu ( \bX_{(1)} (s_2) ) \Big) \bigg\}^2 .
	\end{split}
\end{equation}
By the variance decomposition for the U-statistics shown in the proof of Theorem \ref{thm: DNN-normality} in Section \ref{SecA.2}, we can obtain that 
\begin{equation}
	 \begin{split}
	 	    I_1 (\bx) & \leq (w_1^*)^2 \Big( \frac { s_1^2 } { n^2  } \var ( Y_{(1)} ( \bZ_{1}, \ldots, \bZ_{s_1} ) ) +  \frac {s_1^2} { n } \var( \e [ Y_{(1)} ( \bZ_1, \ldots, \bZ_{s_1} )  | \bX_1 ]  ) \Big) \\
	 	    &  \quad + (w_2^*)^2 \Big( \frac { s_2^2 } { n^2  } \var ( Y_{(1)} ( \bZ_{1}, \ldots, \bZ_{s_2} ) ) +  \frac {s_2^2} { n } \var( \e [ Y_{(1)} ( \bZ_1, \ldots, \bZ_{s_2} )  | \bX_1 ]  ) \Big). 
	 \end{split}
\end{equation}

Observe that we have shown in the proof of Lemma \ref{new.lem.3} that 
\begin{equation}
	\begin{split}
	 \var( Y_{(1)} ( \bX_1, \ldots, \bX_{s_1} ) ) & = \var ( \mu ( X_{(1)} ( \bX_1, \ldots, \bX_{s_1}  )  ) + \epsilon ) \\
	 & \leq \mu^2 (\bx) + \sigma^2 + o(1).
	\end{split}
\end{equation}
Moreover, it follows from \eqref{u-bdd} that 
\begin{align*}
\var( \e [ Y_{(1)} ( \bX_1, \ldots, \bX_{s_1} )  | \bX_1 ]  ) & \leq s_1^{-1} \var ( Y_{(1)} ( \bX_{1}, \ldots, \bX_{s_1} ) ) \\
& \leq s_1^{-1} ( \mu^2 (\bx)  + \sigma^2 + o(1)).
\end{align*}
Similar results also hold for terms related to $s_2$. Thus, we have 
\begin{equation}
	  I_1 (\bx) \leq (\mu^2 (\bx)  + \sigma^2 + o(1)) \Big[ (w_1^*)^2 \cdot \frac {s_1} {n} +  (w_2^*)^2 \cdot \frac {s_2} {n}  \Big].
\end{equation}
 Finally, the desired results can be derived by combining the above bounds for the bias and variance. This concludes the proof of Theorem \ref{thm-MSE}.

	\subsection{Proof of Theorem \ref{thm-jack}} \label{SecA.5}
	
	We now aim to establish the consistency of the jackknife estimator $\hat{\sigma}_{J}^2$ introduced in (\ref{jack}) for the variance $\sigma_n^2$ of the two-scale DNN estimator $ D_n (s_1, s_2) (\bf x)$ as defined in (\ref{neweq.FL032}). We will build on the technique in \cite{A1969} that expands and reorganizes the jackknife estimator $\hat{\sigma}_J^2 $. However, a major theoretical challenge is that instead of an application of the classical asymptotic theory for the case of fixed order, a more delicate technical analysis of the remainders is essential to proving the consistency under our current assumption of diverging order $ s_2  \to \infty$.
	
	More specifically, we will show that the jackknife estimator $ \hat{\sigma}_J^2  $ can be written as a weighted sum of a sequence of U-statistics $ \{U_c\}_{0 \leq c \leq s_2} $ to be introduced in (\ref{U_c}) later, where $U_0$ and $U_1$ are the dominating terms and the remaining ones are asymptotically negligible under the assumption of $s_2 = o(n^{1/3})$. Since U-statistics are symmetric with respect to the input arguments, it follows from (\ref{TDNN-U}) and (\ref{neweq.FL033}) that 
	\begin{align*}
		\sum_{i = 1} ^n {n - 1 \choose s_2 } U_{n - 1}^{(i)} = (n - s_2 ) { n \choose s_2  } D_n (s_1, s_2) ({\bf x}),
	\end{align*}
	which entails that 
	\begin{equation} \label{neweq.FL034}
		n^{-1} \sum_{i= 1}^n U_{n - 1}^{(i)} = D_n (s_1, s_2) ({\bf x}). 
	\end{equation}
	Thus, in light of the definition of the jackknife estimator $\hat{\sigma}_{J}^2$ in (\ref{jack}) and (\ref{neweq.FL034}), we can deduce that 
	\begin{align} \label{neweq.FL035}
		n \hat{\sigma}_{J}^2 &  = (n - 1) \bigg\{  \sum_{i = 1}^n \big( U_{n - 1}^{ (i) } \big)^2 - n ( D_n ( s_1, s_2 ) ( {\bf x} ) )^2 \bigg\} \non
		& = (n - 1) \bigg\{ {n - 1 \choose s_2 }^{ - 2} \sum_{i = 1}^n \sum_{i} \Phi^* ( {\bf x}; {\bf Z}_{\alpha_1^i }, \ldots, {\bf Z}_{\alpha_{s_2}^i } ) \Phi^* ( {\bf x}; {\bf Z}_{\beta_1^i}, \ldots, {\bf Z}_{\beta_{s_2}^i} ) \non
		& \quad - n {n \choose s_2}^{-2} \sum \Phi^* ({\bf x}; {\bf Z}_{\alpha_1}, \ldots, {\bf Z}_{\alpha_{s_2}}) \Phi^* ( {\bf x}; {\bf Z}_{\beta_1}, \ldots, {\bf Z}_{\beta_{s_2}} )
		\bigg\},
	\end{align}
	where we use the shorthand notation $ \sum \limits_{i} $ for  
	\begin{align} \label{neweq.FL036}
		\sum_{\substack{1 \leq \alpha_1^i < \alpha_2^i< \ldots < \alpha_{s_2}^i \leq n \\ 1 \leq \beta_1^i < \beta_2^i  < \ldots < \beta_{s_2}^i \leq n \\ \alpha_1^i , \alpha_2^i , \ldots, \alpha_{s_2}^i \neq i; \, \beta_1^i, \beta_2^i , \ldots, \beta_{s_2}^i \neq i} }
	\end{align}
	and $ \sum $ for 
	\begin{align} \label{neweq.FL037}
		\sum_{\substack{1 \leq \alpha_1 < \alpha_2 < \ldots < \alpha_{s_2} \leq n \\ 1 \leq \beta_1 < \beta_2 < \ldots < \beta_{s_2} \leq n  }}
	\end{align}
	to simplify the technical presentation.
	
	For each $ 0 \leq c \leq s_2 $, by calculating the number of terms with $ c $ overlapping components in $ \Phi^* ( {\bf x}; {\bf Z}_{\alpha_1}, \ldots, {\bf Z}_{\alpha_{s_2}  } ) \Phi^* ( {\bf x}; {\bf Z}_{\beta_{1} }, \ldots, {\bf Z}_{\beta_{s_2}} ) $, we can obtain from (\ref{neweq.FL035})--(\ref{neweq.FL037}) that 
	\begin{align}
		n \hat{\sigma}_{J}^2 & = (n - 1) \bigg\{ {n - 1 \choose s_2}^{- 2 } \sum_{c = 0}^{s_2} (n - 2 s_2 + c)   \sum \Phi^* ( {\bf x}; {\bf Z}_{\alpha_1}, \ldots, {\bf Z}_{\alpha_c}, {\bf Z}_{\beta_1}, \ldots, {\bf Z}_{\beta_{s_2 - c}} )  \non
		& \quad \cdot\Phi^* ({\bf x}; {\bf Z}_{\alpha_1}, \ldots, {\bf Z}_{\alpha_c}, {\bf Z}_{\gamma_1}, \ldots, {\bf Z}_{\gamma_{s_2 - c}}) \non
		&\quad  - n {n \choose s_2}^{-2} \sum_{c = 0 }^{s_2}   \sum \Phi^* ( {\bf x}; {\bf Z}_{\alpha_1}, \ldots, {\bf Z}_{\alpha_c}, {\bf Z}_{\beta_1}, \ldots, {\bf Z}_{\beta_{s_2 - c}} ) \non
		& \quad \cdot\Phi^* ({\bf x}; {\bf Z}_{\alpha_1}, \ldots, {\bf Z}_{\alpha_c}, {\bf Z}_{\gamma_1}, \ldots, {\bf Z}_{\gamma_{s_2 - c}})  \bigg\} \non
		& = \frac {n - 1} {n} {n - 1 \choose s_2}^{ - 2 } \sum_{c = 0}^{s_2} (c n - s_2^2 ) {n \choose 2 s_2 - c} {2 s_2 - c \choose s_2} {s_2 \choose c} U_c, 
		\label{sigma_g}
	\end{align}
	where we introduce a sequence of U-statistics $ \{U_c\}_{0 \leq c \leq s_2} $ defined as 
	\begin{align} \label{U_c}
		& U_c = \bigg\{ {n \choose 2 s_2 - c} {2 s_2 - c \choose s_2} {s_2 \choose c}  \bigg\}^{-1}  \non
		&  \quad \cdot \sum   \Phi^* ( {\bf x}; {\bf Z}_{\alpha_1}, \ldots, {\bf Z}_{\alpha_c}, {\bf Z}_{\beta_1}, \ldots, {\bf Z}_{\beta_{s_2 - c}} ) \Phi^* ({\bf x}; {\bf Z}_{\alpha_1}, \ldots, {\bf Z}_{\alpha_c}, {\bf Z}_{\gamma_1}, \ldots, {\bf Z}_{\gamma_{s_2 - c}}).
	\end{align}
	Here, with slight abuse of notation, $\sum$ is short for denoting the summation over all possible combinations of distinct $ \alpha_1, \ldots, \alpha_c, \beta_1, \ldots, \beta_{s_2 - c}, \gamma_1, \ldots, \gamma_{s_2 - c}$ satisfying that $1 \leq  \alpha_1<  \ldots < \alpha_c \leq n$, $1 \leq \beta_1<  \ldots <  {\beta}_{s_2 - c}\leq n$, and $1 \leq  \gamma_1<  \ldots<  \gamma_{s_2 - c} \leq n$. 
	
	Observe that by symmetrization, $ U_c $ defined in (\ref{U_c}) is indeed a U-statistic that can be represented as 
	\begin{align} \label{neweq.FL038}
		U_c = {n \choose 2s_2 - c}^{-1} \sum_{C_{2 s_2 - c} } K^{(c)} ( {\bf x}; {\bf Z}_{\alpha_1}, \ldots, {\bf Z}_{\alpha_{c}}, {\bf Z}_{\beta_1}, \ldots, {\bf Z}_{\beta_{s_2 - c}}, {\bf Z}_{\gamma_1}, \ldots, {\bf Z}_{\gamma_{s_2 - c} } ),
	\end{align}
	where $\sum \limits_{C_{2 s_2 - c} } $ represents the summation taken over all combinations of $ 1 \leq \alpha_1 < \ldots < \alpha_c < \beta_1 < \ldots  < \beta_{s_2 - c} < \gamma_1 < \ldots < \gamma_{s_2 - c} \leq n$, and the symmetrized kernel function $ K^{(c)} $ is given by 
	\begin{align} \label{neweq.FL039}
		& K^{(c)} ( {\bf x}; {\bf Z}_{\alpha_1}, \ldots, {\bf Z}_{\alpha_{c}}, {\bf Z}_{\beta_1}, \ldots, {\bf Z}_{\beta_{s_2 - c}}, {\bf Z}_{\gamma_1}, \ldots, {\bf Z}_{\gamma_{s_2 - c} } ) \non
		& =   \bigg\{ { 2 s_2 - c \choose c } {2 s_2 - 2 c \choose s_2 - c} \bigg\}^{-1} \sum_{  \prod_{2s_2 -c}} \Phi^* ( {\bf x}; {\bf Z}_{i_1}, \ldots, {\bf Z}_{i_c}, {\bf Z}_{i_{c+1}}, \ldots, {\bf Z}_{i_{s_2}} ) \non
		& \quad \cdot \Phi^* ( {\bf x}; {\bf Z}_{i_1}, \ldots, {\bf Z}_{i_c}, {\bf Z}_{i_{s_2 + 1}}, \ldots, {\bf Z}_{i_{2 s_2 - c}} )
	\end{align}
	with $\sum \limits_{\prod_{2 s_2 - c} } $ standing for the summation over all the $ {2 s_2 - c \choose c } {2 s_2 - 2 c \choose s_2 - c}   $ possible permutations of $ ( \alpha_1, \ldots, \alpha_c, \beta_1, \ldots, \beta_{s_2 - c}, \gamma_1, \ldots, \gamma_{s_2 - c} ) $ that are not permuted within sets $ (\alpha_1, \ldots, \alpha_c) $, $ (\beta_1, \ldots, \beta_{s_2 - c}) $, and $ (\gamma_1, \ldots, \gamma_{s_2 - c}) $. 
	
	From (\ref{sigma_g})--(\ref{neweq.FL039}) above, we can further deduce that as long as $ s_2 = o(\sqrt n) $, it holds that 
	\begin{align*}
		n \hat{\sigma}_{J}^2 & = \sum_{c = 0}^{s_2} ( c n - s_2^2 ) \frac { ( n - s_2 - 1 ) (n - s_2 - 2) \cdots (n - 2 s_2 + c + 1 ) } { (n - 2) ( n - 3) \cdots ( n - s_2 ) c !} \\
		& \quad \cdot [ s_2 (s_2 - 1) \cdots (s_2 - c + 1) ]^2 U_c  \\
		& = - s_2^2 \Big[ 1 + O  \big( \frac {s_2^2 } { n } \big) \Big] U_0 + s_2^2 [ 1 + O \big( \frac { s_2^2  } { n }  \big) ] U_1 +  \sum_{c = 2}^{s_2} O \big( \frac  {s_2^2} { n } \big)^{c - 1 } \frac { s_2^2 } { c ! } U_c \\
		& = s_2^2 ( U_1 - U_0 ) + O \big( \frac  { s_2^4} { n } \big) ( U_0 + U_1  ) + \sum_{c = 2}^{s_2} O \big( \frac  {s_2^2} { n } \big)^{c - 1 } \frac { s_2^2 } { c ! } U_c,
	\end{align*}
	which leads to 
	\begin{align} \label{neweq.FL040}
		\frac { n } {s_2^2}\hat{\sigma}_{J}^2 = U_1 - U_0 + O \big( \frac {s_2^2 } { n } \big) ( U_0 + U_1  ) + \sum_{c = 2} ^{s_2}  O \big( \frac { s_2^2  } { n } \big)^{c - 1 } \frac { U_c } { c ! }.
	\end{align}
	By (\ref{U_c}), for the mean we have 
	\begin{align} \label{neweq.FL041}
		\e U_c & = \e \big[ \Phi^* ( {\bf x}; {\bf Z}_{\alpha_1}, \ldots, {\bf Z}_{\alpha_c}, {\bf Z}_{\beta_1}, \ldots, {\bf Z}_{\beta_{s_2 - c}} )  \Phi^* ( {\bf x}; {\bf Z}_{\alpha_1}, \ldots, {\bf Z}_{\alpha_c}, {\bf Z}_{\gamma_1}, \ldots, {\bf Z}_{\gamma_{s_2 - c}}) \big] \non
		& = \e \big( [ \Phi^*_c ( {\bf x}; {\bf Z}_1, \ldots, {\bf Z}_c ) ]^2 \big),
	\end{align}
	where $\Phi_c^*( {\bf x}; {\bf Z}_{1}, \ldots, {\bf Z}_{c} ) = \e [ \Phi^* ( {\bf x}; {\bf Z}_1, \ldots, {\bf Z}_{s_2} ) | {\bf Z}_1,   \ldots, {\bf Z}_c ] $.
	
	As for the variance, it follows from Lemmas \ref{le-Uc} and \ref{le-k} in Sections \ref{SecB.2} and \ref{SecB.3}, respectively, that for each $ 0 \leq c \leq s_2 $ and fixed $ {\bf x} $, we have 
	\begin{align} \label{neweq.FL042}
		\Var ( U_c) = O ( s_2 / n  ).
	\end{align}
	Moreover, in view of (\ref{neweq.FL041}) and Jensen's inequality, it holds that for each $2 \leq  c \leq s_2 $, 
	\begin{equation} \label{neweq.FL043}
		\e U_c  \leq \e [ (\Phi^*)^2 ]. 
	\end{equation}
	Consequently, it follows from (\ref{neweq.FL040})--(\ref{neweq.FL043}) that 
	\begin{align} \label{neweq.FL044}
		&  \e \Big(\Big[ \frac {n} {s_2^2}\hat{\sigma}_{J}^2  - \Var ( \Phi^*_1 ( {\bf x}; {\bf Z}_1 ) ) \Big]^2  \Big)\non
		& \leq C \Big\{  \Var ( U_1 ) + \Var (U_0)  + \frac {s_2^4} {n^2 } \big[ ( \e \Phi^* )^4   +  \big( \e  [ \Phi^*_1 ( {\bf x}; {\bf Z}_1 ) ] ^2 \big)^2 \big] \non
		& \quad + \sum_{j = 2}^{s_2} \sum_{i = 2}^{s_2} \big(\frac {s_2^2} { n } \big)^{ i + j - 2  } \big[ \Var ( U_i )   + ( \e  (\Phi^*)^2  )^2 \big]^{1/2}  \big[ \Var ( U_j  ) + ( \e ( \Phi^*) ^2 )^2   \big]^{1/2}   \Big\},
	\end{align}
	where $C$ is some positive constant. Recall the facts that $ \e [ \Phi^* ] = O (1) $ and $ \e [ (\Phi^* )^2 ] = O(1)$, which have been shown previously in the proof of Theorem \ref{thm1} in Section \ref{SecA.3}. Combining (\ref{neweq.FL044}) with these facts yields 
	\begin{align}
		\e \Big( \Big[ \frac {n} {s_2^2}\hat{\sigma}_{J}^2  - \Var ( \Phi^*_1 ( {\bf x}; {\bf Z}_1 ) ) \Big]\Big)^2 \leq C \Big( \frac {s_2} {n} + \frac { s_2^4 } {n^2} \Big).  \label{error-jack}
	\end{align}
	
	Furthermore, it has been shown in the proof of Theorem \ref{thm1} in Section \ref{SecA.3} that 
	\begin{equation} \label{neweq.FL045}
		\Var ( \Phi^*_1 ( {\bf x}; {\bf Z}_1 ) ) \geq C s_2^{-1}
	\end{equation}
	with $C$ some positive constant. Thus, when $ s_2 = o (n^{1/3}) $, we can obtain from (\ref{error-jack}) and (\ref{neweq.FL045}) that 
	\begin{align} \label{neweq.FL046}
		\frac {\hat{\sigma}_{J}^2} { \frac {s_2^2 } {n} \Var ( \Phi^*_1 ({\bf x}; {\bf Z}_1) )} \stackrel{p}{\longrightarrow} 1. 
	\end{align}
	In addition, it follows from \eqref{varphi*} and the decomposition for the variance of the U-statistic that 
	as long as $s_2  = o(n)$, we have 
	\begin{equation} \label{neweq.FL047}
		\frac { \sigma^2  } { \frac {s_2^2 } {n} \Var( \Phi^*_1 ({\bf x}; {\bf Z}_1))} \to 1. 
	\end{equation}
	Therefore, combining (\ref{neweq.FL046}) and (\ref{neweq.FL047}) results in $ \hat{\sigma}_J^2 / \sigma_n^2\stackrel{p}{\longrightarrow} 1$, which establishes the desired consistency of the jackknife estimator $\hat{\sigma}_{J}^2$. This completes the proof of Theorem \ref{thm-jack}.

	\subsection{Proof of Theorem \ref{thm-boot}} \label{SecA.6}
	
	We now proceed with establishing the consistency of the bootstrap estimator $\hat{\sigma}_{B, n}^2$ introduced in (\ref{B-var}) for the variance $\sigma_n^2$ of the two-scale DNN estimator $ D_n (s_1, s_2) (\bf x)$ as defined in (\ref{neweq.FL032}). Let us define the bootstrap version of the quantity $\sigma_n^2$ conditional on the given sample $\{{\bf Z}_1, \ldots, {\bf Z}_n\}$ as 
	\begin{equation} \label{neweq.FL048}
		\hat{\sigma}_n^2  = \Var ( D_n^{*} (s_1, s_2) ({\bf x}) | {\bf Z}_1, \ldots, {\bf Z}_n),
	\end{equation}
	where $D_n^{*} (s_1, s_2)$ defined in (\ref{boot-Dn}) denotes the two-scale DNN estimator constructed as in (\ref{TDNN-U}) using the bootstrap sample $ \{{\bf Z}_1^*, \ldots, {\bf Z}_n^*\}$. In fact, the quantity introduced in (\ref{neweq.FL048}) above provides a crucial bridge. The main ingredients of the proof consist of two parts. First, we will show that the bootstrap estimator $ \hat{\sigma}_{B, n}^2$ is asymptotically close to $ \hat{\sigma}_{n}^2 $ given in (\ref{neweq.FL048}) as the number of bootstrap samples $ B \to \infty $. Second, we will prove that the bootstrap version $ \hat{\sigma}_n^2 $ is further asymptotically close to the population quantity $ \sigma^2 $ under the assumption of $ s_2  = o(n^{1/3}) $. It is worth mentioning that the technical analysis for the second part relies on the consistency of the jackknife estimator $\hat{\sigma}_{J}^2$ established in Theorem \ref{thm-jack}.
	
	For each $ 1 \leq b \leq B $, denote by $D_n^{(b)} (s_1, s_2) ({\bf x})$ the two-scale DNN estimator $D_n^{*} (s_1, s_2)$ constructed using the $b$th bootstrap sample. It is easy to see from (\ref{neweq.FL048}) that for each $ 1 \leq b \leq B $, 
	\begin{equation} \label{neweq.FL049}
		\Var ( D_n^{(b)} (s_1, s_2) ({\bf x})  | {\bf Z}_1, \ldots, {\bf Z}_n)  = \hat{\sigma}_n^2.
	\end{equation}
	Since the sample variance defined in \eqref{B-var} is an unbiased estimator for the population variance, by (\ref{neweq.FL049}) it holds that 
	\begin{equation} \label{neweq.FL050}
		\e [ \hat{\sigma}_{B, n}^2 | {\bf Z}_1, {\bf Z}_2, \ldots, {\bf Z}_n] = \hat{\sigma}_n^2 . 
	\end{equation}
	Thus, in view of (\ref{B-var}) and (\ref{neweq.FL050}), we can obtain
	\begin{equation} \label{neweq.FL051}
		\e [ ( \hat{\sigma}_{B, n}^2 - \sigma^2 )^2  ] = \e [  ( \hat{\sigma}_{B, n}^2 - \hat{\sigma}_n^2 )^2 ] + \e [ ( \hat{\sigma}_n^2 - \sigma^2 )^2  ].
	\end{equation}
	Without loss of generality, let us assume that $ \e [ D_n (s_1, s_2) ({\bf x}) ] = 0 $ to ease our technical presentation; otherwise we can subtract the mean first. 
	
	We begin with considering the first term $ \e [  ( \hat{\sigma}_{B, n}^2 - \hat{\sigma}_n^2 )^2 ] $ on the right-hand side of (\ref{neweq.FL051}). Since $ \{ D_n^{(b)} (s_1, s_2) ({\bf x}) \}_{1 \leq b \leq B} $ are i.i.d. random variables conditional on the given sample $ \{{\bf Z}_1, \ldots , {\bf Z}_n \}$, we can deduce that 
	\begin{align} \label{neweq.FL052}
		& \e [ ( \hat{\sigma}_{B, n}^2 - \hat{\sigma}_n^2 )^2 | {\bf Z}_1, \ldots, {\bf Z}_n ]\non
		& = \e \Big[ \frac {  1 } { (B  -  1)^2  }  \Big( \sum_{b = 1}^B  \big( [D_n^{(b)} ( s_1, s_2 ) ({\bf x}) ]^2 - \hat{\sigma}_n^2  \big) - ( B \bar{D}_{B, n}^2  - \hat{\sigma}_n^2 ) \Big)^2    \Big|  {\bf Z}_1, \ldots, {\bf Z}_n  \Big] \non
		& \leq \frac { 2  } { (B  -  1 )^2 } \bigg\{ \e \Big[ \Big( \sum_{b = 1}^B \big( [ D_n^{(b)} (s_1, s_2) ({\bf x}) ]^2 - \hat{\sigma}_n^2 \big)  \Big)^2  \Big|  {\bf Z}_1, \ldots, {\bf Z}_n   \Big] \non
		& \quad + \e \big[ ( B \bar{D}_{B, n}^2 - \hat{\sigma}_n^2 )^2 |  {\bf Z}_1, \ldots, {\bf Z}_n  \big]
		\bigg\} \non
		& \leq  \frac { 2  } { (B  -  1 )^2 } \Big\{ B \e \big[   \big( [ D_n^{(1)} (s_1, s_2) ({\bf x}) ]^2 - \hat{\sigma}_n^2 \big)^2   \big|  {\bf Z}_1, \ldots, {\bf Z}_n   \big] \non
		& \quad +  \frac {2 B} {B^2} \e \big[   \big( [ D_n^{(1)} (s_1, s_2) ({\bf x}) ]^2 - \hat{\sigma}_n^2 \big)^2    \big|  {\bf Z}_1, \ldots, {\bf Z}_n  \big] \non
		& \quad +  \frac {4} {B^2} \sum_{1 \leq i \neq j \leq B}  \e \big[  \big(D_n^{(i)} (s_1, s_2) ({\bf x})  \big)^2   \big(D_n^{(j)} (s_1, s_2) ({\bf x})  \big)^2   \big|  {\bf Z}_1, \ldots, {\bf Z}_n   \big]  
		\Big\} \non
		& \leq \frac { C } {B} \e \big[   \big( [ D_n^{(1)} (s_1, s_2) ({\bf x}) ]^2 - \hat{\sigma}_n^2 \big)^2    \big|  {\bf Z}_1, \ldots, {\bf Z}_n  \big]  \non
		& \quad + \frac {C } {B^2 }  \Big( \e \big[   [D_n^{(1)} (s_1, s_2) ({\bf x})]^2  \big|    {\bf Z}_1, \ldots, {\bf Z}_n  \big] \Big)^2  \non
		& \leq \frac {C} {B} \e \big[ [  D_n^{(1)} (s_1, s_2) ({\bf x}) ]^4   \big|  {\bf Z}_1, \ldots, {\bf Z}_n  \big],
	\end{align}
	where the last inequality follows from the conditional Jensen's inequality. 
	
	Let $ k = [n / s_2] $ be the integer part of the number $ n / s_2 $. We define 
	\begin{align} \label{neweq.FL053}
		h({\bf Z}_1, \ldots, {\bf Z}_n ) & = k^{-1} \big( \Phi^* ({\bf x}; {\bf Z}_1, \ldots, {\bf Z}_{s_2}) +  \Phi^* ( {\bf x}; {\bf Z}_{s_2 + 1}, \ldots, {\bf Z}_{2 s_2} ) \non
		& \quad  + \ldots + \Phi^* ({\bf x}; {\bf Z}_{k s_2 - s_2 + 1}, \ldots, {\bf Z}_{k s_2 } ) \big).
	\end{align}
	Note that it has been shown in (2.1.15) in \cite{Korolyuk_Theory_1994} that 
	\begin{equation} \label{neweq.FL054}
		\e \big[ [  D_n^{(1)} (s_1, s_2) ({\bf x}) ]^4   \big|  {\bf Z}_1, \ldots, {\bf Z}_n  \big] \leq \e [ h^4  ( {\bf Z}_{1}^*, \ldots, {\bf Z}_{n}^* )  | {\bf Z}_1, \ldots, {\bf Z}_n]
	\end{equation}
	with the functional $h(\cdot)$ given in (\ref{neweq.FL053}). Moreover, with an application of Rosenthal's inequality for independent random variables, we can obtain that
	\begin{align} \label{neweq.FL055}
		&  \e [  h^4  ( {\bf Z}_{1}^*, \ldots, {\bf Z}_{n}^* )  | {\bf Z}_1, \ldots, {\bf Z}_n ] \non
		& \leq C k^{-4} k^2 \e \big( [  \Phi^* ({\bf x}; {\bf Z}_1 ^*, \ldots, {\bf Z}_{s_2}^*)  ]^4  \big| {\bf Z}_1, \ldots, { \bf Z }_n \big),
	\end{align}
	where $C$ is some positive constant. Then in light of (\ref{neweq.FL055}), it remains to bound the quantity $\e ( [  \Phi^* ({\bf x}; {\bf Z}_1 ^*, \ldots, {\bf Z}_{s_2}^*)  ]^4)$, which has been dealt with in Lemma \ref{le-boot} in Section \ref{SecB.4}. Thus, it follows from (\ref{neweq.FL052}), (\ref{neweq.FL054})--(\ref{neweq.FL055}), and Lemma \ref{le-boot} that 
	\begin{align} \label{B-part}
		\e [  ( \hat{\sigma}_{B, n}^2 - \hat{\sigma}_n^2 )^2 ] & \leq \frac { C }{B} \frac  {s_2^2 }{n^2 }  n^{-s_2}  \sum_{i_1 = 1}^n \ldots \sum_{i_{s_2} = 1}^n \e \big( [ \Phi^* ({\bf x}; {\bf Z}_{i_1}, \ldots, {\bf Z}_{i_{s_2}}) ]^4 \big) \non
		& \leq  \frac { C M s_2^2  } {B n^2 },
	\end{align}
	where $M$ is some positive constant given in Lemma \ref{le-boot}.
	
	We next proceed with analyzing the second term $ \e [ ( \hat{\sigma}_n^2 - \sigma^2 ) ^2 ] $ on the right-hand side of (\ref{neweq.FL051}). Recall the definition of the bootstrap version $ \hat{\sigma}_n^2 $ for the population quantity $ \sigma^2 $ introduced in (\ref{neweq.FL048}). Let us define
	\begin{equation} \label{neweq.FL056}
		m_n = \e [  \Phi^* ({\bf x}; {\bf Z}_1^*, \ldots, {\bf Z}_{s_2}^* ) | {\bf Z}_1, \ldots, {\bf Z}_n ] 
	\end{equation}
	and 
	\begin{equation} \label{neweq.FL057}
		h_1 ( {\bf z} ) = \e [ \Phi^* ({\bf x}; {\bf Z}_1^*, \ldots, {\bf Z}_{s_2}^* )  - m_n | {\bf Z}_1^* = {\bf z}]. 
	\end{equation}
	Then applying similar arguments as for \eqref{remainder} in the proof of Theorem \ref{thm: DNN-normality} in Section \ref{SecA.2}, we can deduce that 
	\begin{equation} \label{neweq.FL058}
		\hat{\sigma}_n^2  = \frac {s_2^2 } {n} \e [ h_1^2  ({\bf Z}_1^* ) | {\bf Z}_1, \ldots, {\bf Z}_n ] +  \Delta_1, 
	\end{equation}
	where $ 0 \leq  \Delta_1  \leq  \frac { s_2^2 } { n^2  } \Var ( \Phi^* ({\bf x};  {\bf Z}_1^* , \ldots, {\bf Z}_{s_2}^* )  | {\bf Z}_1, \ldots,  {\bf Z}_n )   $ and function $h_1(\cdot)$ is given in (\ref{neweq.FL057}) and (\ref{neweq.FL056}). Similarly, it holds that 
	\begin{equation} \label{neweq.FL059}
		\sigma_n^2 = \frac { s_2^2 } {n} \e [ g_1^2  ({\bf Z}_1 ) ] + \Delta_2 ,
	\end{equation}
	where $ g_1 ({\bf Z}_1) = \e [  \Phi^* ({\bf x}; {\bf Z}_1, \ldots, {\bf Z}_n) | {\bf Z}_1 ] $ and $ 0 \leq \Delta_2  \leq \frac {s_2^2 } {n^2}  \Var (\Phi^* ( {\bf x}; {\bf Z}_1, \ldots, {\bf Z}_{s_2} ) )$. Hence, by (\ref{neweq.FL058}) and (\ref{neweq.FL059}) we can obtain that 
	\begin{align} \label{part2}
		& \e [ ( \hat{\sigma}_n^2 - \sigma_n^2 )^2 ] \non
		& \leq C \e \Big( \frac {s_2^4} {n^2} \big[ \e [ h_1^2 ({\bf Z}_1^*) | {\bf Z}_1, \ldots, {\bf Z}_n ] - \e [ g_1^2 ({\bf Z}_1) ] \big]^2 + \Delta_1^2 + \Delta_2^2 \Big),
	\end{align}
	where $C$ is some positive constant. 
	
	Observe that
	\begin{equation} \label{neweq.FL060}
		\Delta_2^2 = O (\frac {s_2^4 } {n^4})
	\end{equation}
	and 
	\begin{align} \label{neweq.FL061}
		\e (\Delta_1^2) & \leq \frac { s_2^4 } { n^4 } \e \big[ \e \big( [ \Phi^* ({\bf x}; {\bf Z}_1^*, \ldots, {\bf Z}_{s_2}^* ) ]^2 | {\bf Z}_1, \ldots, {\bf Z}_n \big) \big]^2 \non
		& \leq \frac { s_2^4  } {n^4} \e \big( [ \Phi^* ({\bf x}; {\bf Z}_1^*, \ldots, {\bf Z}_{s_2}^*) ]^4 \big) \non
		& \leq \frac {M s_2^4 } { n^4  },    
	\end{align}
	where the last inequality follows from Lemma \ref{le-boot} with $M$ some positive constant. In addition, it holds that 
	\begin{equation} \label{neweq.FL062}
		\e  [ h_1^2 ({\bf Z}_1^* ) | {\bf Z}_1, \ldots, {\bf Z}_n ]  = \frac {1} {n} \sum_{i= 1}^n h_1^2 ({\bf Z}_i )
	\end{equation}
	and 
	\begin{align} \label{neweq.FL063}
		h_1 ({\bf Z}_i ) &= n^{ - s_2 + 1 } \sum_{i_{2} = 1}^n \cdots \sum_{i_{s_2} = 1}^n \Phi^* ({\bf x }; {\bf Z}_i, {\bf Z}_{i_2}, \ldots, {\bf Z}_{i_{s_2}} )  \non
		& \quad - n^{- s_2} \sum_{i_1 = 1}^n \cdots \sum_{i_{s_2} = 1}^n \Phi^* ({\bf x}; {\bf Z}_{i_1}, \ldots, {\bf Z}_{i_{s_2}}).      
	\end{align}
	
	Let us further define  
	\begin{align} \label{S_i}
		S_i = {n - 1  \choose s_2 - 1 }^{-1} \sum_{ \substack{1 \leq  j_1 < j_2 < \ldots < j_{s_2 - 1} \leq n \\ j_1, j_2, \ldots, j_{s_2 - 1} \neq i }}  \Phi^* ( {\bf x}; {\bf Z}_{i}, {\bf Z}_{j_1},   \ldots, {\bf Z}_{j_{s_2 - 1}} ) . 
	\end{align}
	From the equality $ {n- 1 \choose  s_2 } U_{n - 1}^{(i)}  + {n - 1 \choose s_2 - 1} S_i = {n \choose s_2} D_n (s_1, s_2) ({\bf x}) $ in view of (\ref{S_i}), it is easy to see that the jackknife estimator $\hat{\sigma}_{J}^2$ introduced in \eqref{jack} satisfies that 
	\begin{align} \label{neweq.FL064}
		\frac { n \hat{\sigma}_J^2 } {s_2^2 } =  \frac { n - 1 } { ( n -  s_2 )^2 } \sum_{i = 1}^n  \big( S_i -  D_n (s_1, s_2) ({\bf x}) \big)^2.
	\end{align}
	Then the main idea of the remaining proof is to show that under the assumption of $ s_2 = o(n^{1/3}) $, $ h_1 ({\bf Z}_i ) $ is asymptotically close to $ S_i - D_n (s_1, s_2) ({\bf x}) $ and thus $\e  [ h_1^2 ({\bf Z}_1^* ) | {\bf Z}_1, \ldots, {\bf Z}_n ]   $ is asymptotically close to $ \frac { n \hat{\sigma}_J^2 } {s_2^2 } $. Observe that 
	\[ n^{ - s_2 + 1 } {n- 1 \choose s_2 - 1} (s_2 - 1)! = 1 + O ( s_2^2 / n ) \] 
	and \[ n^{ - s_2 } {n \choose s_2 } s_2!= 1 + O (s_2^2 / n),\]
	which entail that 
	\[ \Big( n^{s_2 - 1}  -  {n - 1 \choose s_2 - 1} (s_2 - 1)! \Big) n^{ - s_2 + 1} = O (s_2^2 / n) \]
	and \[ \Big(n^{s_2} - {n \choose s_2} s_2! \Big) n^{- s_2 }  = O (s_2 ^2 / n). \] 
	Thus, it follows from (\ref{neweq.FL063}) and these facts  that 
	\begin{align} \label{neweq.FL065}
		h_1 ({\bf Z}_i )   &= (1 + O (s_2^2 / n) )  \big[ S_i  - D_n (s_1 , s_2 ) ({\bf x}) \big] + n^{ - s_2 + 1}    \sum_{\mathscr{D}_1} \Phi^* ( {\bf x}; {\bf Z}_i, {\bf Z}_{i_2}, \ldots, {\bf Z}_{i_{s_2}}) \non
		& \quad  - n^{ - s_2} \sum_{\mathscr{D}_2} \Phi^* ( {\bf x}; {\bf Z}_{i_1}, {\bf Z}_{i_2}, \ldots, {\bf Z}_{i_{s_2}}),
	\end{align}
	where $ \mathscr{D}_1 = \{(i_2, \ldots, i_{s_2} ) : \mbox{there is at least one pair that are equal or there is a component} $ $\mbox{that is equal to}~ i\} $ and $ \mathscr{D}_2 = \{ (i_1, \ldots, i_{s_2}): \mbox{there is at least one pair of components that are}$ $ \mbox{equal}\}$. 
	
	With an application of similar arguments as in the proof of Lemma \ref{le-boot} in Section \ref{SecB.4}, we can obtain that 
	\begin{equation} \label{neweq.FL066}
		\e \big( [\Phi^* ( {\bf x}; {\bf Z}_{i_1}, {\bf Z}_{i_2}, \ldots, {\bf Z}_{i_{s_2}} ) ]^4 \big) \leq M
	\end{equation}
	with $M$ some positive constant, regardless of how many components of $ (i_1, i_2, \ldots, i_{s_2}) $ are equal. As a consequence, by (\ref{neweq.FL066}) it holds that 
	\begin{align}
		& \e \Big[ \Big( \frac {1} {n} \sum_{i = 1}^n \Big(  n^{ - s_2 + 1}   \sum_{\mathscr{D}_1} \Phi^* ( {\bf x}; {\bf Z}_i , {\bf Z}_{i_2}, \ldots, {\bf Z}_{i_{s_2}}) \Big) ^2  \Big)^2  \Big]  \non
		& \leq \frac {1} {n} \sum_{i=1}^n \e \Big[ \Big(  n^{ - s_2 + 1} \sum_{\mathscr{D}_1} \Phi^* ( {\bf x}; {\bf Z}_{i}, {\bf Z}_{i_2}, \ldots, {\bf Z}_{i_{s_2}} ) \Big)^4 \Big]  \leq \frac { C M s_2^8 } { n^4 }
		\label{e1}
	\end{align}
	and similarly, 
	\begin{equation}
		\e \Big[ \Big( \frac {1} {n} \sum_{i = 1}^n \Big( n^{ - s_2} \sum_{\mathscr{D}_2} \Phi^* ({\bf x}; {\bf Z}_{i_1}, {\bf Z}_{i_2}, \ldots, {\bf Z}_{i_{s_2}} ) \Big) \Big)^2  \Big] \leq  \frac { C M s_2^8  } {n^4 },
		\label{e2}
	\end{equation}
	where $C$ represents some positive constant whose value may change from line to line. Hence, combining (\ref{neweq.FL062}), (\ref{neweq.FL065}), and (\ref{e1})--(\ref{e2}), we can deduce that as long as $ s_2 = o(n^{1/3}) $, it holds that 
	\begin{align} \label{neweq.FL067}
		& \e \Big( \big[ \e [ h_1^2 ({\bf Z}_{1}^* )  | {\bf Z}_1, \ldots, {\bf Z}_{n} ] - \e [ g_1^2 ({\bf Z}_1) ] \big]^2  \Big)  \non
		& \leq C \e \Big[ \Big( \frac {1} {n} \sum_{i = 1}^n ( 1 + O (s_2^2 / n) )^2 [ S_i - D_n (s_1, s_2) ({\bf x}) ]^2  - \Var (\Phi^*_1 ( {\bf x}; {\bf Z}_1 )) \Big)^2 \Big] \non
		& \quad + \frac { C M s_2^8 } {n^4} \non
		& \leq  C \e \Big[ \Big( \frac { (n - s_2)^2 } {n (n - 1)} (1 + O(s_2^2 / n) ) \frac { n } {s_2^2} \hat{\sigma}_J^2 - \Var (\Phi_1^* ({\bf x}; {\bf Z}_1))  \Big)^2 \Big] + \frac {C M s_2^8} {n^4} \non
		& \leq C  ( 1 + O(s_2^2 / n) ) \e \big( [ \frac { n } {s_2^2} \hat{\sigma}_J^2  - \Var (\Phi^*_1 ({\bf x}; {\bf Z}_1)) ]^2\big) + \frac {C s_2^4} {n^2 } ( \Var (\Phi^*_1 ({\bf x}; {\bf Z}_1)) )^2 \non
		& \quad + \frac { C M s_2^8 } {n^4}\non
		& \leq  \frac {C s_2} {n} + \frac {s_2^2 } {n^2 } + \frac {C M s_2^8} {n^4 } \leq \frac {C (M + 1) s_2} {n},
	\end{align}
	where the second to the last inequality comes from \eqref{error-jack} and \eqref{var-ub} in the proof of Lemma \ref{lemma-eta} in Section \ref{SecB.10}. 
	
	Substituting the above bounds in (\ref{neweq.FL060})--(\ref{neweq.FL061}) and (\ref{neweq.FL067}) into \eqref{part2} leads to
	\begin{equation} \label{neweq.FL068}
		\e [ ( \hat{\sigma}_n^2 - \sigma_n^2 )^2 ]  = O  \Big( \frac {s_2^5} {n^3} + \frac {s_2^4} {n^4} \Big).
	\end{equation}
	Thus, combining (\ref{B-part}) and (\ref{neweq.FL068}), we can obtain that 
	\begin{equation} \label{neweq.FL069}
		\e [ ( \hat{\sigma}_{B, n}^2 - \sigma_n^2 )^2 ] = O \Big(\frac {s_2^5} {n^3} + \frac {s_2^2} {B n^2} \Big). 
	\end{equation}
	Recall the fact that $ \sigma^2 = O ( \frac {s_2} {n} ) $ under the assumption of $ s_2 = o(n) $. Consequently, such fact along with (\ref{neweq.FL069}) entails that 
	\begin{equation} \label{neweq.FL070}
		\e \Big[ \Big( \frac { \hat{\sigma}_{B, n}^2 } {\sigma_n^2} - 1 \Big)^2 \Big]  = O ( \frac {s_2^3 } {n}  + \frac {1} {B} ).
	\end{equation}
	Therefore, combining (\ref{neweq.FL070}) and the assumptions of $ s_2 = o(n^{1/3}) $ and $ B \to \infty $ yields $ \hat{\sigma}_{B, n}^2 / \sigma_n^2\stackrel{p}{\longrightarrow} 1$, which establishes the desired consistency of the bootstrap estimator $\hat{\sigma}_{B, n}^2$. This concludes the proof of Theorem \ref{thm-boot}.
	
\ignore{	
	\red{
		\subsection{Proof of Theorem \ref{thm-boot-semiparam}}
		The assumptions (i-iv) are special cases of the conditions (C1-ND), (C2), (C3) and (C4) in  \cite{song2019approximating}. Hereby, Theorem \ref{thm-boot-semiparam} follows from Theorem 3.1 in \cite{song2019approximating}.
	}
}

\subsection{Proof of Theorem \ref{thm-boot-dist}}	\label{pf-thm-boot-dist}

The main idea of the proof is to show that both the TDNN estimator $D_n(s_1, s_2) (\bx)$ and its bootstrap version $ D_n^*(s_1, s_2) (\bx) $ are asymptotically normal and in addition, their asymptotic variances are close to each other. Then the conditional distribution of $D_n^*(s_1, s_2)(\bx)$ given $(\bZ_1, \ldots, \bZ_n)$   approaches the distribution of $D_n(s_1, s_2)(\bx)$ as the sample size $n$ increases. To this end, let us first recall that it has been shown in Theorem \ref{thm1} that $D_n(s_1, s_2) (\bx)$ is asymptotically normal. Since the normal distribution $\Phi(\cdot)$ is continuous, it follows that 
\begin{equation} \label{error-original-dist}
    \sup_{u \in \mathbb{R}}\big|\mathbb{P} ( \sigma_n^{-1}(D_n(s_1, s_2) (\bx) - \mu(\bx) - \Lambda) \leq u ) - \Phi(u) \big| = o(1),
\end{equation}
where $\sigma_n ^2 = \Var(D_n(s_1, s_2) (\bx))$.

We next deal with the bootstrapped statistic $D_n^*(s_1, s_2) (\bx)$. In light of Hoeffding's decomposition for the U-statistic, we have 
\begin{equation*}
   {D}_n^* (s_1, s_2) (\bx) - \theta^* = \frac{s_2} {n} \sum_{i = 1}^n \hat{g}_1(\bx;\bZ_i^*) + R_n^*,
\end{equation*}
where $\hat{g}_1^* (\bx; \bz) = \mathbb{E}^* \big[ \Phi^*(\bx; \bZ_1^*, \ldots, \bZ_{s_2}^*)  | \bZ_1^* = \bz \big] - \theta^*$ with the expectation $\mathbb{E}^*$ taken with respect to the bootstrap resampling distribution of $(\bZ_2^*, \ldots, \bZ_n^*)$ given $(\bZ_1, \ldots, \bZ_n)$, and $R_n^*$ is the higher-order remainder. Given $(\bZ_1, \ldots, 
\bZ_n)$, the Berry--Esseen theorem for the sum of i.i.d. random variables \citep{berry1941} leads to 
\begin{align*}
    & \sup_{u \in \mathbb{R}} \Big| \mathbb{P}^* \Big( [ n \Var^*(\hat{g}_1(\bx; \bZ_1^*)) ]^{-1/2} \sum_{i = 1}^n \hat{g}_1 (\bx; \bZ_i^*)  \leq u \Big) - \Phi(u)
    \Big| \\
    & \quad \leq \frac{ \mathbb{E}^* ( | \hat{g}_1(\bx; \bZ_1^*) |^3 ) } { \sqrt n \Var^* (\hat{g}_1(\bx; \bZ_1^*)) },
\end{align*}
where the variance $\Var^*$ is again taken with respect to the bootstrap resampling distribution given $(\bZ_1, \ldots, \bZ_n)$. 

An application of similar arguments as in the proof of Lemma \ref{lemma-eta} yields 
\[
\Var^*(\hat{g}_1 (\bx; \bZ_1^*) ) \sim O_p(s_2^{-1}). 
\]
It follows from Jensen's inequality that 
\[
\mathbb{E}^*(|\hat{g_1}(\bx; \bZ_1^*) |^3) \leq \mathbb{E}^*(|\Phi^*(\bx; \bZ_1^*, \ldots, \bZ_{s_2}^*)|^3).
\]
Similar to Lemma \ref{lemma-phi}, we can deduce that 
\[
\mathbb{E}( |\Phi^*(\bx; \bZ_1, \ldots, \bZ_{s_2}) |^3) \leq M
\]
for some positive constant $M$. Hence, it holds that $ \mathbb{E}^*(|\Phi^*(\bx; \bZ_1^*, \ldots, \bZ_{s_2}^*)|^3) = O_p(1) $ and $  \mathbb{E}^*(|\hat{g_1}(\bx; \bZ_1^*) |^3) = O_p(1)$. Consequently, the approximation error satisfies that 
\begin{align} \label{error-boot-dist}
    & \sup_{u \in \mathbb{R}} \Big| \mathbb{P}^* \Big( [ n \Var^*(\hat{g}_1(\bx; \bZ_1^*)) ]^{-1/2} \sum_{i = 1}^n \hat{g}_1 (\bx; \bZ_i^*)  \leq u \Big) - \Phi(u)
    \Big| \nonumber \\
    &\quad = O_p(s_2/\sqrt n) = o_p(1) 
\end{align} 
since $s_2 = o(n^{1/3})$. 

Let us define $\hat{\sigma}_n^2 = \Var[D_n^*(s_1, s_2)(\bx)|\bZ_1, \ldots, \bZ_n]$. Then the variance decomposition of the U-statistic implies that 
\begin{equation*}
    \hat{\sigma}_n^2 = \frac {s_2^2} {n} \Var^*( \hat{g}_1(\bx; \bZ_1^*)) + \Var^*(R_n^*).
\end{equation*}
Note that from the similar argument as in \eqref{remainder}, we see that the remainder $R_n^*$ above satisfies that 
\begin{equation*}
    \Var^* (R_n^*) \leq \frac {s_2^2} {n^2} \Var^* ( \Phi^*(\bx; \bZ_1^*, \ldots, \bZ_{s_2}^*) ).
\end{equation*}
Since it has been shown in Section \ref{SecB.9}  that the second moment $\e \{ [\Phi^*(\bx; \bZ_1, \ldots, \bZ_{s_2})]^2 \} \leq M $ for some positive constant $M$, we have 
\[
\Var^* ( \Phi^*(\bx; \bZ_1^*, \ldots, \bZ_{s_2}^*) ) = O_p(1), 
\]
and thus $ \Var^*(R_n^*) = O_p(s_2^2 /n^2) $. Furthermore, it follows from $\Var^*(\hat{g}_1(\bx; \bZ_1^*)) \sim O_p(s_2^{-1})$ that 
\[
\frac{s_2^2} {n} \Var^*(\hat{g}_1 (\bx; \bZ_1^*)) / \hat{\sigma}_n^2 \stackrel{p}{\longrightarrow} 1.
\]
Hence, \eqref{error-boot-dist} entails that 
\begin{equation}
    \sup_{u \in \mathbb{R}} \Big| \mathbb{P}^* \Big( \hat{\sigma}_n^{-1} (D_n^*(s_1, s_2)(\bx) - \theta^*)  \leq u \Big) - \Phi(u)
    \Big| = o_p(1).
\end{equation}

Moreover, we have shown in Section \ref{SecA.3} that $ \sigma_n $ is of order $(s_2 / n)^{1/2}$ and in Section \ref{SecA.6} that $\hat{\sigma}_n / \sigma_n \stackrel{p}{\longrightarrow} 1 $. Therefore, combining \eqref{error-original-dist} and \eqref{error-boot-dist} results in 
\begin{align*}
   & \sup_{u \in \mathbb{R}} \Big| \mathbb{P}^*\Big(\sigma_n^{-1} (D_n^*(s_1, s_2)(\bx) - \theta^*) \leq u \Big) - \mathbb{P} \Big( \sigma_n ^{-1} (D_n(s_1, s_2)(\bx) - \mu(\bx) - \Lambda) \Big) \Big| \\
   & \quad = o_p(1).
\end{align*}
Since $\sigma_n$ is of order $(s_2/n)^{1/2}$ and unknown in practice, we can rewrite the above approximation error as 
\begin{align*}
     & \sup_{u \in \mathbb{R}} \Big| \mathbb{P}^*\Big( (s_2/n)^{-1/2} (D_n^*(s_1, s_2)(\bx) - \theta^*) \leq u \Big) - \mathbb{P} ( (s_2/n) ^{-1/2} (D_n(s_1, s_2)(\bx) - \mu(\bx) - \Lambda) ) \Big| \\
     & \quad = o_p(1).
\end{align*}
This completes the proof of Theorem \ref{thm-boot-dist}.

\subsection{Proof of Theorem \ref{thm2}} \label{SecA.4}
	We now aim to prove the asymptotic normality of the HTE estimator 
	\[
	\widehat{\tau}(\bx) = D_{n_1}^{(1)} ( s_1^{(1)}, s_2^{(1)} ) ( {\bf x} ) \\- D_{n_0}^{(0)} ( s_1^{(0)}, s_2^{(0)} ) ( {\bf x} )
	\]
	introduced in (\ref{neweq.FL031}), where $D_{n_1}^{(1)} ( s_1^{(1)}, s_2^{(1)}) ( {\bf x} )$ and $D_{n_0}^{(0)} ( s_1^{(0)}, s_2^{(0)}) ( {\bf x} )$ denote the two-scale DNN estimators constructed using the treatment sample of size $n_1$ and the control sample of size $n_0$, respectively.  Denote by $n = n_0 + n_1$ the total sample size.  By the assumption $P(T=1|\bX, Y_{T=0}, Y_{T=1})=1/2$, it is easy to see that $n_0/n_1 \stackrel{p}{\longrightarrow} 1$  as $n\rightarrow \infty$.  For each of the treatment and control groups in the randomized experiment, by the assumptions a separate application of Theorem \ref{thm1} shows that there exist some positive numbers $\sigma_{n_1}$ of order  $(s_2^{(1)}/n_1)^{1/2}$ and $\sigma_{n_0}$ of order  $(s_2^{(0)}/n_0)^{1/2}$ such that 
	\begin{equation} \label{neweq.FL025}
		\frac { D_{n_1}^{(1)} ( s_1^{(1)}, s_2^{(1)} ) ( {\bf x} ) - \e [ D_{n_1}^{(1)} ( s_1^{(1)}, s_2^{(1)} ) ( {\bf x} ) ] }  { \sigma_{n_1} } \toD N (0, 1) 
	\end{equation}
	and 
	\begin{equation} \label{neweq.FL026}
		\frac { D_{n_0}^{(0)} ( s_1^{(0)}, s_2^{(0)} ) ( {\bf x} ) - \e [ D_{n_0}^{(0)} ( s_1^{(0)}, s_2^{(0)} ) ( {\bf x} ) ] }  { \sigma_{n_0} } \toD N (0, 1).
	\end{equation}
	
	In view of the randomized experiment assumption, the treatment sample and control sample are independent of each other, which entails that the two separate two-scale DNN estimators $D_{n_1}^{(1)} ( s_1^{(1)}, s_2^{(1)}) ( {\bf x} )$ and $D_{n_0}^{(0)} ( s_1^{(0)}, s_2^{(0)}) ( {\bf x} )$ are independent. Thus it follows from (\ref{neweq.FL025}) and (\ref{neweq.FL026}) that 
	\begin{align} \label{neweq.FL028}
		& \frac { D_{n_1}^{(1)} ( s_1^{(1)}, s_2^{(1)} ) ( {\bf x} ) - D_{n_0}^{(0)} ( s_1^{(0)}, s_2^{(0)} ) ( {\bf x} ) - \e [ D_{n_1}^{(1)} ( s_1^{(1)}, s_2^{(1)} ) ( {\bf x} ) - D_{n_0}^{(0)} ( s_1^{(0)}, s_2^{(0)} ) ( {\bf x} )] }  { \sigma_{n} } \non
		&\quad \toD N (0, 1),
	\end{align}
	where we define $\sigma_n = (\sigma_{n_1}^2 + \sigma_{n_0}^2)^{1/2}$. Moreover, from the higher-order asymptotic expansion of the bias term in Theorem \ref{thm: bias} applied to the potential treatment and control responses, respectively, and the definition of the heterogeneous treatment effect (HTE) $ \tau ({\bf x}) $ introduced in (\ref{neweq.FL112}), we see that 
	\begin{equation} \label{neweq.FL030}
		\e [ D_{n_1}^{(1)} ( s_1^{(1)}, s_2^{(1)} ) ( {\bf x} ) ]  -  \e [ D_{n_0}^{(0)} ( s_1^{(0)}, s_2^{(0)} ) ( {\bf x} ) ] = \tau ({\bf x})+ \Lambda,
	\end{equation}
	where {\color{black} $\Lambda = O\{(s_1^{(1)})^{-4/d}+(s_2^{(1)})^{-4/d}+(s_1^{(0)})^{-4/d}+(s_2^{(0)})^{-4/d}\}$ for $ d \geq 2 $ and $ \Lambda = O\{(s_1^{(1)})^{-3}+(s_2^{(1)})^{-3}+(s_1^{(0)})^{-3}+(s_2^{(0)})^{-3}\} $ for $ d = 1 $}. Therefore, combining (\ref{neweq.FL028}) and (\ref{neweq.FL030}) yields the desired asymptotic normality of the HTE estimator $\widehat{\tau}(\bx)$ based on the two-scale DNN estimators. This concludes the proof of Theorem \ref{thm2}.

\renewcommand{\thesubsection}{E.\arabic{subsection}}
	\section{Some key lemmas and their proofs} \label{SecB}

\ignore{	
	\subsection{Proof of Lemma \ref{le-L}} \label{SecB.1}
	
	Observe that the set of possible values $ Y_{(1)}  ( {\bf Z}_{i_1}, {\bf Z}_{i_2}, \ldots, {\bf Z}_{i_s} ) $ can take is $ \{Y_{(1)}, Y_{(2)}, \ldots, Y_{(n - s +1)} \}$. If $ Y_{(1)}  ( {\bf Z}_{i_1}, {\bf Z}_{i_2}, \ldots, {\bf Z}_{i_s} ) = Y_{(1)} $, then the observation corresponding to $ Y_{(1)} $ must be selected and there are $ {n - 1 \choose s - 1} $ options for the remaining $ s - 1 $ places in the subsample. In general, if $ Y_{(1)}  ( {\bf Z}_{i_1}, {\bf Z}_{i_2}, \ldots, {\bf Z}_{i_s} ) = Y_{(j)} $ for some $ 1 \leq j \leq n - s + 1 $, then the observation corresponding to $ Y_{(j)} $ must be selected and the observations corresponding to $ Y_{(1)}, Y_{(2)}, \ldots, Y_{(j - 1)}$ will not be selected, which entails that there are $ {n - j  \choose s - 1} $ options for the remaining $ s - 1 $ places in the subsample. Applying these arguments, we can obtain the representation in \eqref{eqn:dnn-lstat} for the single-scale DNN estimator $D_n (s) ({\bf x })$.
	
	It remains to show that for any given ${\bf x}$, $ D_n (s) ({\bf x}) $ is in fact an L-statistic. Denote by $ Y_{1:n}, Y_{2:n}, \ldots$, $ Y_{n:n}$ the order statistics for $n$ scalars $ Y_1, Y_2, \ldots, Y_n  $. Note that for each given ${\bf x}$, $ Y_{(1)}, Y_{(2)}, \ldots, Y_{(n - s - 1)}$ are fixed and there exists some set $ S ({\bf x}) := \{i_j: 1 \leq j \leq n - s + 1 \} $ that may depend upon ${\bf x}$ such that 
	\begin{align} \label{neweq.FL071}
		(Y_{(1)}, Y_{(2)}, \ldots, Y_{(n - s + 1 )}) = ( Y_{i_1 : n}, Y_{i_2 : n }, \ldots, Y_{i_{n - s + 1} : n} ).
	\end{align}
	Conversely, for each $ k \in S ({\bf x }) $, there exists some $ 1 \leq m_k \leq n - s + 1 $ that may depend on ${\bf x}$ such that $ Y_{k : n}  = Y_{(m_k)} $. Consequently, combining such fact, (\ref{neweq.FL071}), and the representation in \eqref{eqn:dnn-lstat} established above, we can obtain that 
	\begin{align*}
		D_n (s) ({\bf x}) = {n \choose s}^{ - 1 } \sum_{k \in S(\bf {x})} { n - m_k \choose s - 1 } Y_{k : n},
	\end{align*}
	which shows that for any given ${\bf x}$, $ D_n (s) ({\bf x} ) $ is an L-statistic. This completes the proof of Lemma \ref{le-L}.
}	
	
	\subsection{Lemma \ref{le-Uc} and its proof} \label{SecB.2}
	
	\begin{lemma}  \label{le-Uc}
		Under the conditions of Theorem \ref{thm-jack}, we have that for each $ 0 \leq c \leq s_2 $ and fixed ${\bf x}$,   
		\begin{align}
			\Var ( U_c ) \leq \frac { 2 s_2  - c   } { n }  \Var ( K^{( c )} ), \label{bound-Uc}
		\end{align}
		where $U_c$ is the U-statistic defined in (\ref{U_c}) and $ K^{(c)} $ is the symmetrized kernel function given in (\ref{neweq.FL039}).
	\end{lemma}
	
	\noindent \textit{Proof}. For notational simplicity, we will drop the dependence of all the functionals on the fixed vector ${\bf x}$ whenever there is no confusion. For each $ 1 \leq  j \leq 2s_2 - c $, let us define 
	\begin{align*}
		K^{(c)}_j ( {\bf Z}_1, \ldots, {\bf Z}_j )   & = \e [ K^{(c)}  | {\bf Z}_1, \ldots, {\bf Z}_j ],  \\
		g^{(c)}_j ({\bf Z}_1, \ldots, {\bf Z}_j )   & = K^{(c) }_j - \e [ K^{(c)} ] - \sum_{i = 1}^{j - 1} \sum_{1 \leq  \alpha_1< \ldots < \alpha_i \leq j} g_{i}^{(c) }  ( {\bf Z}_{\alpha_1}, \ldots, {\bf Z}_{\alpha_i} ), 
	\end{align*}
	and $V_j  = \Var ( g^{(c)}_j ( {\bf Z}_1, \ldots, {\bf Z}_j )  ) $. Then it follows from Hoeffding's decomposition that 
	\begin{align} \label{neweq.FL072}
		U_c = \e [ K^{ (c) } ] + {n \choose 2 s_2 - c}^{-1}   \sum_{i = 1}^{2 s_2 - c }  {n - i \choose 2 s_2 - c  - i} \sum_{1 \leq \alpha_1 < \ldots < \alpha_i \leq n  } g^{(c)}_i ( {\bf Z}_{\alpha_1}, \ldots, {\bf Z}_{\alpha_i} ).
	\end{align}
	
	Observe that $ \Var ( K^{(c)} )  = \sum_{i = 1}^{2 s_2 - c} {2 s_2 - c \choose i } V_i $. Thus, in view of (\ref{neweq.FL072}), we can deduce that 
	\begin{align*}
		\Var ( U_c  ) & = \sum_{i = 1}^{2 s_2- c}   {n \choose 2 s_2 - c}^{-2}   {n - i \choose 2 s_2 - c  - i} ^2  {n \choose i } V_j \\
		& =   \sum_{i = 1}^{2 s_2 - c} \frac { ( 2 s_2 - c ) ! (n - i) ! } { n ! (2 s_2 - c - i) ! } {2 s_2 - c \choose i} V_i \\
		& \leq    \frac {  2 s_2 - c    } {  n } \sum_{i = 1}^{2 s_2 - c} {2 s_2 - c \choose i} V_i \\
		&= \frac {2 s_2 - c} { n } \Var ( K^{(c)} ),
	\end{align*}
	which establishes the desired upper bound in \eqref{bound-Uc}. This completes the proof of Lemma \ref{le-Uc}.

	\subsection{Lemma \ref{le-k} and its proof} \label{SecB.3}
	
	\begin{lemma} \label{le-k}
		Under the conditions of Theorem \ref{thm-jack}, it holds that for each $ 0 \leq c \leq s_2 $ and fixed ${\bf x}$, 
		\begin{align} \label{neweq.FL073}
			\Var ( K^{(c)}  ) \leq  C [ (w_1^* )^4 + (w_2^*)^4 ] \big(\mu^4 ({\bf x}) + 6 \mu^2 ({\bf x}) \sigma_{\epsilon} + 4 \mu ({\bf x}) + \e [ \epsilon_1^4 ] \big),
		\end{align}	    
		where $ K^{(c)} $ is the symmetrized kernel function given in (\ref{neweq.FL039}) and $C$ is some positive constant.
	\end{lemma}
	
	\noindent \textit{Proof}. By the Cauchy--Schwarz inequality, we can deduce that
	\begin{align} \label{neweq.FL074}
		& \Var ( K^{(c)} )  \leq
		\e [ ( K^{(c)} )^2 ] \non
		&  = \bigg[  {2 s_2 - c \choose c} { 2 s_2 - 2 c \choose s_2 - c } \bigg]^{ -2 }    \sum_{ \Pi_{2 s_2 - c} }  \sum_{ \Pi_{2 s_2 - c} }    \non
		& \quad \e \big\{ \Phi^* ( {\bf x}; {\bf Z}_{i_1}, \ldots, {\bf Z}_{i_c}, {\bf Z}_{i_{c + 1}}, \ldots, {\bf Z}_{i_{s_2}} )   \Phi^* ( {\bf x}; {\bf Z}_{i_1}, \ldots, {\bf Z}_{i_c}, {\bf Z}_{i_{s_2 + 1}}, \ldots, {\bf Z}_{i_{2 s_2 - c}} ) \non
		& \quad \times  \Phi^* ( {\bf x}; {\bf Z}_{j_1}, \ldots, {\bf Z}_{j_c}, {\bf Z}_{j_{c + 1}}, \ldots, {\bf Z}_{j_{s_2}} )   \Phi^* ( {\bf x}; {\bf Z}_{j_1}, \ldots, {\bf Z}_{j_c}, {\bf Z}_{j_{s_2 + 1}}, \ldots, {\bf Z}_{j_{2 s_2 - c}} ) \big\} \non
		&\leq \e \big\{  \big[  \Phi^* ( {\bf x}; {\bf Z}_{1}, {\bf Z}_{2}, \ldots, {\bf Z}_{s_2} )    \big]^4  \big\},
	\end{align}
	where $\sum \limits_{\prod_{2 s_2 - c} } $ denotes the summation introduced in (\ref{neweq.FL039}). In light of the definition of $\Phi^*$ in \eqref{phistar}, we have 
	\begin{align} \label{neweq.FL075}
		\e & \big\{  \big[  \Phi^* ( {\bf x}; {\bf Z}_{1}, {\bf Z}_{2}, \ldots, {\bf Z}_{s_2} )    \big]^4  \big\} 
		\leq 8 ( w_1^* )^4 \e [ \Phi^4  ( {\bf x}; {\bf Z}_{1}, \ldots, {\bf Z}_{{s_1}} ) ] \non
		&\quad+ 8 ( w_2^* )^4 \e [ \Phi^4  ( {\bf x}; {\bf Z}_{1}, \ldots, {\bf Z}_{s_2} ) ] .
	\end{align}
	
	Let us make some useful observations. Note that 
	\begin{align*}
		\e [ \Phi ^4 ( {\bf x}; {\bf Z}_{1}, \ldots, {\bf Z}_{s_1} ) ] 
		& = \e \Big[  \Big( \sum_{i = 1}^n y_i \zeta_{i, s_1} \Big)^4 \Big]  \\
		& = \sum_{i = 1 } \e [ y_i^4 \zeta_{i, s_1} ] = s_1 \e [ y_1^4 \zeta_{1, s_1} ] 
	\end{align*}
	and 
	\begin{align*}
		\e [ y_1^4 \zeta_{1, s_1} ] & = \e \big(  [ \mu ({\bf X}_1 ) + \epsilon_1 ]^4 \zeta_{1, s_1} \big) \\
		& = \e[ \mu^4 ({\bf X}_1)  \zeta_{1, s_1}] + 6 \e [ \mu^2 ({\bf X}_1) \zeta_{1, s_1} ] \sigma_{\epsilon}^2 + 4 \e [ \mu ({\bf X}_1) \zeta_{1, s_1} ] + \e [ \epsilon_1^4 ],
	\end{align*}
	where $ \zeta_{i, s} $ represents the indicator function for the event that  $ \bX_i $ is the 1NN of $ \bx $ among  $ \bX_1, \cdots, \bX_{s} $. Moreover, it follows from Lemma \ref{lem:delta-simple} in Section \ref{SecC.3} that as $s_1 \to \infty $,  
	\[ s_1  \e [   \mu^k ( {\bf X}_1  ) \zeta_{1, s_1} ] \to \mu^k ( {\bf x} ) \] 
	for $k = 1, 2, 4$. Hence, it holds that 
	\begin{align*}
		& \e [ \Phi ^4 ( {\bf x}; {\bf Z}_{1}, \ldots, {\bf Z}_{s_1} ) ]  =  s_1 \e [ y_1^4 \zeta_{1, s_1} ] \\
		& \to \mu^4 ({\bf x}) + 6 \mu^2 ({\bf x}) \sigma_{\epsilon} + 4 \mu ({\bf x}) + \e [ \epsilon_1^4 ]
	\end{align*}
	as $s_1 \to \infty $. 
	
	Using similar arguments, we can show that as $s_2 \to \infty$,
	\begin{align*}
		\e [ \Phi^4 ({\bf x}; {\bf Z}_{1}, \ldots, {\bf Z}_{s_2}) ] \to \mu^4 ({\bf x}) + 6 \mu^2 ({\bf x}) \sigma_{\epsilon} + 4 \mu ({\bf x}) + \e [ \epsilon_1^4 ].
	\end{align*}
	Therefore, combining the asymptotic limits obtained above, (\ref{neweq.FL074}), and (\ref{neweq.FL075}) results in  
	\begin{align*}
		\Var ( K^{(c)} )  \leq C [ (w_1^* )^4 + (w_2^*)^4 ] \big(\mu^4 ({\bf x}) + 6 \mu^2 ({\bf x}) \sigma_{\epsilon} + 4 \mu ({\bf x}) + \e [ \epsilon_1^4 ] \big),
	\end{align*}
	where $C$ is some positive constant. This concludes the proof of Lemma \ref{le-k}.

	\subsection{Lemma \ref{le-boot} and its proof} \label{SecB.4}
	
	\begin{lemma} \label{le-boot}
		Under the conditions of Theorem \ref{thm-boot}, there exists some constant $M > 0 $ depending upon $w_1^*$, $w_2^*$,  ${\bf x}  $, and the distribution of $ \epsilon $ such that
		\begin{equation}
			\e \big( [ \Phi^* ({\bf x}; {\bf Z}_1^* , \ldots, {\bf Z}_{s_2}^* ) ]^4 \big) \leq M.
		\end{equation}
	\end{lemma}
	
	\noindent \textit{Proof}. Since the observations in the bootstrap sample $ \{{\bf Z}_1^*, \ldots, {\bf Z}_n^* \} $ are selected independently and uniformly from the original sample $ \{{\bf Z}_1, \ldots, {\bf Z}_n \}$, we have
	\begin{equation*}
		\begin{split}
			\e \big( [ \Phi^* ({\bf x}; {\bf Z}_1^*, \ldots, {\bf Z}_{s_2}^* ) ]^4 \big)& = \e \Big(  \e \big( [ \Phi^* ({\bf x}; {\bf Z}_1^*, \ldots, {\bf Z}_{s_2}^* ) ]^4 \big| {\bf Z}_1, \ldots, {\bf Z}_n \big) \Big) \\
			& = n^{- s_2 } \sum_{i_1 = 1}^n \cdots \sum_{i_{s_2} = 1}^n \e \big( [ \Phi^* ( {\bf x}; {\bf Z}_{i_1}, \ldots, {\bf Z}_{i_{s_2}} ) ]^4 \big).
		\end{split}
	\end{equation*}
	Observe that for distinct $ i_1, \ldots, i_{s_2} $, we have shown in the proof of Lemma \ref{le-k} in Section \ref{SecB.3} that as $ s_2 \to \infty $, 
	$$ 
	\e \big( [ \Phi^* ({\bf x}; {\bf Z}_1, \ldots, {\bf Z}_{s_2}) ]  ^4 \big) \rightarrow A 
	$$ 
	for some positive constant $A$ that depends upon $w_1^*$, $w_2^*$, ${\bf x} $, and the distribution of $  \epsilon $. 
	
	Furthermore, note that if $ i_1 = i_2 = \ldots = i_c $ and the remaining arguments are distinct, then it holds that 
	\[ \Phi ({\bf x}; {\bf Z}_{i_1}, \ldots, {\bf Z}_{i_{s_2}} ) = \Phi ({\bf x};  {\bf Z}_{i_1}, {\bf Z}_{i_{c+ 1}}, \ldots, {\bf Z}_{i_{s_2}} ). \]
	Therefore, there exists some positive constant $ M $ depending upon $ w_1^*$, $w_2^*$, ${\bf x} $, and the distribution of $ \epsilon $ such that
	$$ \e \big( [ \Phi^* ( {\bf x}; {\bf Z}_{i_1}, \ldots, {\bf Z}_{i_{s_2}} ) ]^4 \big) \leq M $$
	for any $ 1 \leq i_1 \leq n , \ldots, 1 \leq i_{s_2} \leq n$. This completes the proof of Lemma \ref{le-boot}.

	\subsection{Lemma \ref{lemma1} and its proof} \label{SecB.5}
	
	In Lemma \ref{lemma1} below, we will provide the asymptotic expansion of $\mathbb{E} \; \|\bX_{(1)} - \bx\|^k	$ with $k \geq 1$ and its higher-order asymptotic expansion for the case of $k = 2$ as the sample size $n \rightarrow \infty$.  
	
	\begin{lemma} \label{lemma1} 
		Assume that Conditions \ref{cond:tail}--\ref{cond3} hold and $\bx \in \supp(\bX) \subset \mathbb{R}^d$ is fixed. Then the $1$-nearest neighbor (1NN) $\bX_{(1)}$ of $\bx$ in the i.i.d. sample $\{\bX_1, \cdots, \bX_n\}$ satisfies that for any 
		$ k \geq 1 $,  
		\begin{equation} \label{eqn:lemma1-main}
			\mathbb{E} \; \|\bX_{(1)} - \bx\|^k	= \frac{\Gamma(k/d + 1)}{(f(\bx) V_d)^{k/d}} n^{-k/d} + o(n^{-k/d})
		\end{equation}
		as $n \rightarrow \infty$, where	$\Gamma(\cdot)$ is the gamma function and $ V_d = \frac{\pi^{d/2}}{\Gamma(1 + d/2)} $. In particular, when $ k = 2 $, there are three cases. If $ d = 1 $, we have 
			\begin{equation} \label{neweq.L3}
				\mathbb{E} \; \|\bX_{(1)} - \bx\|^2	=   \frac{\Gamma(2/d + 1)}{(f(\bx) V_d)^{2/d}} n^{-2/d} -  \left(  \frac { \Gamma (2/d + 2) } {d  (f ({\bf x}) V_d)^{2/d} }  \right) n^{ - ( 1 + 2/ d )  }  + o ( n^{ - ( 1 + 2/ d )  }  ).
			\end{equation}
			If $d = 2$, we have
			\begin{align} \label{neweq.L4}
				\mathbb{E} \; \|\bX_{(1)} - \bx\|^2	& =   \frac{\Gamma(2/d + 1)}{(f(\bx) V_d)^{2/d}} n^{-2/d}  - \left( \frac{\tr(f''(\bx)) \Gamma(4/d + 1)}{f(\bx) (f(\bx) V_{d})^{4/d} d (d+2)} +   \frac { \Gamma (2/d + 2) } {d  (f ({\bf x}) V_d)^{2/d} }      \right) n^{- 4/ d} \non
				& \quad + o(n^{-4/d}),
			\end{align}
			where $f''(\cdot)$ stands for the Hessian matrix of the density function $f(\cdot)$. 
			If $ d \geq 3 $, we have 
			\begin{align} \label{eqn:lemma1-extra-order}
				\mathbb{E} \; \|\bX_{(1)} - \bx\|^2	& = \frac{\Gamma(2/d + 1)}{(f(\bx) V_d)^{2/d}} n^{-2/d} - \left( \frac{\tr(f''(\bx)) \Gamma(4/d + 1)}{f(\bx) (f(\bx) V_{d})^{4/d} d (d+2)}\right) n^{-4/d} \non
				&\quad + o(n^{-4/d}).
			\end{align}
	\end{lemma}
	
	\noindent \textit{Proof}. Denote by $\varphi$ the probability measure on $\mathbb{R}^d$ given by random vector $\bX$. We begin with obtaining an approximation of $ \varphi(B(\bx, r))$, where $B(\bx, r)$ represents a ball in the Euclidean space $\mathbb{R}^d$ with center $\bx$ and radius $r > 0$. {\color{black} Recall that by Condition \ref{cond2}, the density function $f(\cdot)$ of measure $ \varphi $ with respect to the Lebesgue measure $\lambda$ is four times continuously differentiable with bounded corresponding derivatives in a neighborhood of $\bx$}. Then using the Taylor expansion, we see that for any $ \bxi \in S^{d-1} $ and $ 0 < \rho < r $,
	\begin{equation} \label{neweq.FL076}
		f(\bx + \rho \bxi) = f(\bx) + f'(\bx)^{T} \bxi \rho + \frac{1}{2} \bxi ^{T} f''(\bx) \bxi \rho ^2 + o(\rho^2),
	\end{equation}
	where $S^{d-1} $ denotes the unit sphere in $\mathbb{R}^d$, and $f'(\cdot)$ and $f''(\cdot)$ stand for the gradient vector and the Hessian matrix, respectively, of the density function $f(\cdot)$. With the aid of the representation in (\ref{neweq.FL076}), an application of the spherical integration leads to 
	\begin{align} \label{neweq.FL077}
		\varphi(B(\bx, r)) & = \int_0^r \int _{S^{d-1}} f(\bx + \rho \bxi) \rho^{d-1} \nu(\mathrm{d} \bxi) \,\mathrm{d} \rho \non
		& = \int_0^r \int _{S^{d-1}} \left( f(\bx) + f'(\bx)^{T} \bxi \rho + \frac{1}{2} \bxi ^{T} f''(\bx) \bxi \rho ^2 + o(\rho^2)  \right) \rho^{d-1} \nu(\mathrm{d} \bxi) \,\mathrm{d} \rho \non
		& = \int_0^r  \Big[ f(\bx) d  V_d  \rho^{d-1}  + \frac{\tr(f''(\bx) ) V_d}{2} \rho^{d+1} + o(\rho^{d+1}) \Big] \, \mathrm{d} \rho \non
		& =  f(\bx) V_d  r^d  + \frac{\tr(f''(\bx) ) V_d}{2(d+2)} r^{d+2} + o(r^{d+2}),
	\end{align}
	where $\nu$ denotes a measure constructed on the unit sphere $\mathbb{S}^{d-1}$ 
	as characterized in Lemma \ref{lemma:spherical-integration} in Section \ref{sec:Supplemantary-SphericalIntegrationFormulas} and $\mathrm{d} \cdot$ stands for the differential of a given variable hereafter.
	
	We now turn our attention to the target quantity $ \mathbb{E} \, \|\bX_{(1)} - \bx\|^k $ for any $ k \geq 1 $. It holds that 
	\begin{align} \label{neweq.FL078}
		\mathbb{E} \, \|\bX_{(1)} - \bx\|^k
		& =  \int_0^\infty \mathbb{P} \, (\|\bX_{(1)} - \bx\|^k > t) \; \mathrm{d} t \non
		& =  \int_0^\infty \mathbb{P} \, (\|\bX_{(1)} - \bx\| > t^{1/k}) \; \mathrm{d} t \non
		& =  \int_0^\infty [1 - \varphi (B(\bx, t^{1/k}))]^n \; \mathrm{d} t\non
		& =  n^{-k/d} \, \int_0^\infty \left[1 - \varphi \left(B \left(\bx, \frac{t^{1/k}}{n^{1/d}} \right)\right)\right]^n \; \mathrm{d} t.
	\end{align}
	To evaluate the integration in (\ref{neweq.FL078}), we need to analyze the term $ \left[1 - \varphi \left(B \left(\bx, \frac{t^{1/k}}{n^{1/d}} \right)\right)\right]^n $. 
	It follows from the asymptotic expansion of $ \varphi(B(\bx, r)) $ in (\ref{neweq.FL077}) that 
	\begin{align} \label{eqn:x-nbhood-expansion}
		& \left[1 - \varphi \left(B \left(\bx, \frac{t^{1/k}}{n^{1/d}} \right)\right)\right]^n
		\non
		& = \Big[1 - \frac{f(\bx) V_d  t^{d/k}}{n}   -  \frac {\frac{\tr(f''(\bx) ) V_d}{2(d+2)} t^{(d+2) / k}}{n^{1+ 2/d}} \quad + o(n^{-(1+ 2/d)}) \Big]^n .
	\end{align}
	
	From (\ref{eqn:x-nbhood-expansion}), we see that for each fixed $ t > 0$, 
	
	\[ \lim_{n \rightarrow \infty }\left[1 - \varphi \left(B \left(\bx, \frac{t^{1/k}}{n^{1/d}} \right)\right)\right]^n  = \exp ( - f(\bx) V_d  t^{d/k} ). \]
	Moreover, by Condition \ref{cond:tail}, we have 
	\begin{align*} 
		\left[1 - \varphi \left(B \left(\bx, \frac{t^{1/k}}{n^{1/d}} \right)\right)\right]^n
		& \leq \left[\exp  \left( - \alpha \frac{t^{1/k}}{n^{1/d}} \right) \right]^n \\
		&\leq \exp \left( - \alpha t^{1/k}\right).
	\end{align*}
	Thus, an application of the dominated convergence theorem yields  
	\begin{align} \label{neweq.FL079}
		\lim_{n \rightarrow \infty} \int_0^\infty \left[1 - \varphi \left(B \left(\bx, \frac{t^{1/k}}{n^{1/d}} \right)\right)\right]^n \; \mathrm{d} t
		& =   \int_0^\infty  \lim_{n \rightarrow \infty}\left[1 - \varphi \left(B \left(\bx, \frac{t^{1/k}}{n^{1/d}} \right)\right)\right]^n \; \mathrm{d} t \non
		& = \int_0^\infty \exp ( - f(\bx) V_d  t^{d/k} ) \; \mathrm{d} t \non
		& = \frac{\Gamma(k/d  + 1)}{(f(\bx) V_d)^{k/d}},
	\end{align}    
	which establishes the desired asymptotic expansion in (\ref{eqn:lemma1-main}) for any $k \geq 1$.

	We further investigate higher-order asymptotic expansion for the case of $ k = 2 $. The leading term of the asymptotic expansion for $ \mathbb{E} \, \|\bX_{(1)} - \bx\|^2 $ has been identified in (\ref{neweq.FL079}) with the choice of $k = 2$. But we now aim to conduct a higher-order asymptotic expansion. To do so, we will resort to the higher-order asymptotic expansion given in (\ref{eqn:x-nbhood-expansion}). {\color{black} In view of (\ref{eqn:x-nbhood-expansion}), we can deduce from the Taylor expansion for function $ \log (1 - x) $ around $0$ that 
		\begin{align} \label{neweq.FL080}
			& \left[1 - \varphi \left(B \left(\bx, \frac{t^{1/2}}{n^{1/d}} \right)\right)\right]^n -  \exp \left\lbrace    - f(\bx) V_d  t^{d/2}   \right\rbrace \non
			& = \exp \left\lbrace n \log  \left[1 - \frac{f(\bx) V_d  t^{d/2}}{n}   -  \frac {\frac{\tr(f''(\bx) ) V_d}{2(d+2)} t^{(d+2) / 2}}{n^{1+ 2/d}}  + o(n^{-(1+ 2/d)}) \right] \right\rbrace  \non
			&\quad-  \exp \left\lbrace    - f(\bx) V_d  t^{d/2}   \right\rbrace \non
			& = \exp \left\lbrace    - f(\bx) V_d  t^{d/2}   -  \frac {\frac{\tr(f''(\bx) ) V_d}{2(d+2)} t^{(d+2) / 2}}{n^{2/d}}  - \frac {f^2 (\mathbf{x}) V_d^2 t^{d} } { 2 n } + o(n^{-(2/d)})  \right\rbrace \non
			&\quad-  \exp \left\lbrace    - f(\bx) V_d  t^{d/2}   \right\rbrace  
		\end{align}
		as $ n \to \infty$. To determine the order of the above remainders, there are three separate cases, that is, $ d = 1 $, $ d = 2 $, and $ d \geq 3 $. 
		
		First, for the case of $d = 1$, it follows from \eqref{neweq.FL080} that 
		\begin{align}   \label{neweq.L2}
			&  \left[1 - \varphi \left(B \left(\bx, \frac{t^{1/2}}{n^{1/d}} \right)\right)\right]^n -  \exp \left\lbrace    - f(\bx) V_d  t^{d/2}   \right\rbrace \non  
			& =  \exp \left\lbrace    - f(\bx) V_d  t^{d/2}   - \frac {f^2 (\mathbf{x}) V_d^2 t^{d} } { 2 n } + o(n^{-1})  \right\rbrace -  \exp \left\lbrace    - f(\bx) V_d  t^{d/2}   \right\rbrace  \non
			& = \exp  \left\lbrace    - f(\bx) V_d  t^{d/2}   \right\rbrace \left( 
			\exp \left\lbrace   - \frac {f^2 (\mathbf{x}) V_d^2 t^{d} } { 2 n }  + o(n^{-1})   \right\rbrace -  1 \right)  \non 
			& =  \exp  \left\lbrace    - f(\bx) V_d  t^{d/2}   \right\rbrace 	\left(   - \frac {f^2 (\mathbf{x}) V_d^2 t^{d} } { 2 n }  + o(n^{-1}) \right)
		\end{align}
		as $n \to \infty$. Furthermore, it holds that 
		\begin{align}  \label{neweq.L1}
			\int_0^{\infty} \exp \left\lbrace    - f(\bx) V_d  t^{d/2}   \right\rbrace  
			\left(  - \frac {f^2 (\mathbf{x}) V_d^2 t^{d}}  {2 }   \right)  \; \mathrm{d} t &  = -   \frac { \Gamma ( 2 / d + 2) } { d ( f ({\bf x})  V_d )^{2 / d } } ,
		\end{align}
		where we have used the fact that for any $a>0$ and $b>0$, 
		\begin{equation} \label{equality}
			\int_0^\infty x^{a- 1} \exp ( - b x^p ) \, \mathrm{d} x = \frac{1}{p} b^{-a/p} \Gamma(\frac{a}{p}). 
		\end{equation}
		Therefore, combining (\ref{neweq.FL078}), \eqref{neweq.FL079}, \eqref{neweq.L2}, and (\ref{neweq.L1}) results in the desired higher-order asymptotic expansion in (\ref{neweq.L3}) for the case of $ k = 2 $ and $ d = 1 $.

		When $d = 2$, noting that $ 2 / d = 1 $, it follows from \eqref{neweq.FL080} that 
		\begin{align}   \label{neweq.L5}
			&  \left[1 - \varphi \left(B \left(\bx, \frac{t^{1/2}}{n^{1/d}} \right)\right)\right]^n -  \exp \left\lbrace    - f(\bx) V_d  t^{d/2}   \right\rbrace \non  
			& =  \exp \left\lbrace     - f(\bx) V_d  t^{d/2}   -  \frac {\frac{\tr(f''(\bx) ) V_d}{2(d+2)} t^{(d+2) / 2}}{n^{2/d}}  - \frac {f^2 (\mathbf{x}) V_d^2 t^{d} } { 2 n^{ 2  / d} } + o(n^{-(2/d)})  \right\rbrace  \non
			& \quad -  \exp \left\lbrace    - f(\bx) V_d  t^{d/2}   \right\rbrace  \non
			& = \exp  \left\lbrace    - f(\bx) V_d  t^{d/2}   \right\rbrace \left( 
			\exp \left\lbrace    -  \frac {\frac{\tr(f''(\bx) ) V_d}{2(d+2)} t^{(d+2) / 2}}{n^{2/d}}  - \frac {f^2 (\mathbf{x}) V_d^2 t^{d} } { 2 n^{ 2  / d} } + o(n^{-(2/d)})  \right\rbrace -  1 \right)  \non 
			& =  \exp  \left\lbrace    - f(\bx) V_d  t^{d/2}   \right\rbrace 	\left(    -  \frac {\frac{\tr(f''(\bx) ) V_d}{2(d+2)} t^{(d+2) / 2}}{n^{2/d}}  - \frac {f^2 (\mathbf{x}) V_d^2 t^{d} } { 2 n^{ 2  / d} } + o(n^{-(2/d)}) \right)
		\end{align}
		as $n \to \infty$. 
		Applying equality \eqref{equality} again yields
		\begin{align} \label{neweq.FL081}
			& \int_0^{\infty} \exp \left\lbrace    - f(\bx) V_d  t^{d/2}   \right\rbrace  
			\left(    -   \frac{\tr(f''(\bx) ) V_d}{2(d+2) n^{ 2 / d }} t^{(d+2) / 2}   \right)  \; \mathrm{d} t \non
			& =    - \left(  \frac{\tr(f''(\bx) ) \Gamma (4/d + 1) }   { d (d+2) f(\bx) (f(\bx)  V_d )^{4/d } }  \right) n^{ - 2 / d}.
		\end{align}
		%
		Hence, combining (\ref{neweq.FL078}), (\ref{neweq.FL079}), \eqref{neweq.L1}, \eqref{neweq.L5}, and (\ref{neweq.FL081}) leads to the desired higher-order asymptotic expansion in (\ref{neweq.L4}) for the case of $ k = 2 $ and $ d = 2 $. 
		
		Finally, it remains to investigate the case of $d \geq 3$. In view of $ n^{ - 1} = o ( n ^{ - 2/ d}) $ for $d \geq 3$, we can obtain from \eqref{neweq.FL080} that 
		\begin{align}  \label{neweq.L6}
			&  \left[1 - \varphi \left(B \left(\bx, \frac{t^{1/2}}{n^{1/d}} \right)\right)\right]^n -  \exp \left\lbrace    - f(\bx) V_d  t^{d/2}   \right\rbrace \non  
			& =  \exp \left\lbrace     - f(\bx) V_d  t^{d/2}   -  \frac {\frac{\tr(f''(\bx) ) V_d}{2(d+2)} t^{(d+2) / 2}}{n^{2/d}}  + o(n^{-(2/d)})  \right\rbrace  -  \exp \left\lbrace    - f(\bx) V_d  t^{d/2}   \right\rbrace  \non
			& = \exp  \left\lbrace    - f(\bx) V_d  t^{d/2}   \right\rbrace \left( 
			\exp \left\lbrace    -  \frac {\frac{\tr(f''(\bx) ) V_d}{2(d+2)} t^{(d+2) / 2}}{n^{2/d}}  + o(n^{-(2/d)})  \right\rbrace -  1 \right)  \non 
			& =  \exp  \left\lbrace    - f(\bx) V_d  t^{d/2}   \right\rbrace 	\left(    -  \frac {\frac{\tr(f''(\bx) ) V_d}{2(d+2)} t^{(d+2) / 2}}{n^{2/d}}   + o(n^{-(2/d)}) \right). 
		\end{align}
		Consequently, combining (\ref{neweq.FL078}), (\ref{neweq.FL079}), (\ref{neweq.FL081}), and \eqref{neweq.L6} yields the desired higher-order asymptotic expansion in (\ref{eqn:lemma1-extra-order}) for the case of $ k = 2 $ and $ d \geq 3 $. This concludes the proof of Lemma \ref{lemma1}.
	}

	\subsection{Lemma \ref{lem2} and its proof} \label{SecB.6}
	As in \cite{biau2015lectures}, we define the projection of the mean function $\mu(\bX) = \mathbb{E}(Y | \bX)$ onto the positive half line $\mathbb{R}_+ = [0, \infty)$ given by $\|\bX - \bx\|$ as 
	\begin{equation} \label{neweq.FL004}
		m(r) = \lim_{\delta \rightarrow 0+} \mathbb{E} \, [\mu(\bX) \; | \; r \le \|\bX - \bx\| \le r + \delta] = \mathbb{E} \; [Y \; | \; \|\bX - \bx\| = r]
	\end{equation}
	for any $r \geq 0$. Clearly, the definition in (\ref{neweq.FL004}) entails that 
	\begin{equation} \label{neweq.FL082}
		m(0) = \mathbb{E} \, [Y \, | \, \bX = \bx] = \mu(\bx).
	\end{equation} 
	We will show in Lemma \ref{lem2} below that the projection $m(\cdot)$ admits an explicit higher-order asymptotic expansion as the distance $r \rightarrow 0$.

	{\color{black}
		\begin{lemma} \label{lem2}
			For each fixed $\bx \in \supp(\bX) \subset \mathbb{R}^d$, we have 
			%
			\begin{equation} \label{neweq.FL005}
				m(r) = m(0) + \frac{f(\bx) \, \tr (\mu''(\bx)) + 2 \, \mu'(\bx)^Tf'(\bx)}{2 \, d \, f(\bx)} \, r^2 + O_4 r^4 
			\end{equation}
			as $r \rightarrow 0$, where $O_4$ is some bounded quantity depending only on $d$ and the fourth-order partial derivatives of the underlying density function $ f (\cdot ) $ and regression function $ \mu (\cdot) $. Here $g'(\cdot)$ and $g''(\cdot)$ stand for the gradient vector and the Hessian matrix, respectively, of a given function $g(\cdot)$.
		\end{lemma}
	}
	
	\noindent \textit{Proof}. We will exploit the spherical coordinate integration in our proof. Let us first introduce some necessary notation. Denote by $B(\bzero,r)$ the ball centered at $\bzero$ and with radius $r$ in the Euclidean space $\mathbb{R}^d$, $\mathbb{S}^{d-1}$ the unit sphere in $\mathbb{R}^d$, $\nu$ a measure constructed on the unit sphere $\mathbb{S}^{d-1}$ as in (\ref{neweq.FL077}), and $\bxi = (\xi_i) \in \mathbb{S}^{d-1}$ an arbitrary point on the unit sphere. Let $V_d$ be the volume of the unit ball in $\mathbb{R}^d$ as given in (\ref{eqn:lemma1-main}). The integration with the spherical coordinates is equivalent to the standard integration through the identity
	\begin{equation} \label{neweq.FL083}
		\int_{\tiny B(\bzero,r)} \, f(\bx) \, \mathrm{d} \bx = \int_0^r u^{d-1} \int_{\mathbb{S}^{d-1}} \, f(u \, \bxi) \, \nu (\mathrm{d}  \bxi) \, \mathrm{d} u.
	\end{equation}
	From Lemma \ref{lemma:spherical-integration} in Section \ref{sec:Supplemantary-SphericalIntegrationFormulas}, we have the following integration formulas with the spherical coordinates
	\begin{eqnarray} \label{neweq.FL085}
		\int_{\mathbb{S}^{d-1}} \; \nu (\mathrm{d} \bxi) & = & d \, V_d,	\\
		\int_{\mathbb{S}^{d-1}} \bxi \; \nu(\mathrm{d} \bxi) & = & \bzero,	\\
		\int_{\mathbb{S}^{d-1}} \bxi^T A \, \bxi \; \nu (\mathrm{d} \bxi) & = & \tr (A) \, V_d,	\\
		\int_{\mathbb{S}^{d-1}}   \xi_{i} \xi_{j} \xi_{k} \nu (\mathrm{d} \bxi) & = & 0  \quad \mbox{for any} ~ 1 \leq i, j, k \leq d,    \label{neweq.L7} 
	\end{eqnarray}
	where $A$ is any $d \times d$ symmetric matrix. We will make use of the identities in (\ref{neweq.FL085})--(\ref{neweq.L7}) in our technical analysis.
	%

	Let us decompose $m(r)$ into two terms that we will analyze separately 
	\begin{align} \label{neweq.FL084}
		m(r) 
		& = \lim_{\delta \rightarrow 0+} \mathbb{E} \, [\mu(\bX) \; | \; r \le \|\bX - \bx\| \le r + \delta] \non
		& = \lim_{\delta \rightarrow 0+} \frac{\mathbb{E} \, [\mu(\bX) \mathbbm{1}(r \le \|\bX - \bx\| \le r + \delta)]}{\mathbb{P} \, (r \le \| \bX - \bx\| \le r + \delta)},
	\end{align}
	where $\mathbbm{1}(\cdot)$ stands for the indicator function. In view of (\ref{neweq.FL083}), we can obtain the spherical coordinate representations for the denominator and numerator in (\ref{neweq.FL084})
	\begin{equation} \label{neweq.FL087}
		\mathbb{P} \, (r \le \|\bX - \bx\| \le r + \delta)  = \int_r^{r+\delta} u^{d-1} \int_{\mathbb{S}^{d-1}} \, f(\bx + u \, \bxi) \, \nu (\mathrm{d} \bxi) \, \mathrm{d} u
	\end{equation}
	and 
	\begin{align} \label{neweq.FL088}
		& \mathbb{E} \, [\mu(\bX) \mathbbm{1}(r \le \|\bX - \bx\| \le r + \delta)] \non 
		& = \int_r^{r+\delta} u^{d-1} \int_{\mathbb{S}^{d-1}} \, \mu(\bx + u \, \bxi) f(\bx + u \, \bxi) \, \nu (\mathrm{d} \bxi) \, \mathrm{d} u.
	\end{align}
	Note that in light of (\ref{neweq.FL084})--(\ref{neweq.FL088}), an application of L'H\^opital's rule leads to 
	\begin{align} \label{eqn:mr-long-division}
		m(r) 
		& = \lim_{\delta \rightarrow 0+} \frac{\mathbb{E} \, [\mu(\bX) \mathbbm{1}(r \le \|\bX - \bx\| \le r + \delta)]}{\mathbb{P} \, (r \le \| \bX - \bx\| \le r + \delta)} \non
		& = \frac{\int_{\mathbb{S}^{d-1}} \mu(\bx + r \, \bxi)f(\bx + r \, \bxi) \; \nu (\mathrm{d} \bxi)}{\int_{\mathbb{S}^{d-1}} f(\bx + r \, \bxi) \; \nu(\mathrm{d} \bxi)}.
	\end{align}

	{\color{black}
		First let us expand the denominator. Using the spherical coordinate integration, we can deduce that 
		\begin{align} \label{neweq.FL089}
			& \int_{\mathbb{S}^{d-1}} f(\bx + r  \bxi) \; \nu(\mathrm{d} \bxi) \non
			& = \int_{\mathbb{S}^{d-1}} \Big(f(\bx) + f'(\bx)^T\, \bxi \, r+ \frac{1}{2} \, \bxi^T \, f''(\bx) \, \bxi \, r^2 + 
			\frac{1}{6} \sum_{1 \leq i,j,k \leq d} \frac{\partial^3 f(\bx)}{\partial \bx_{i} \partial \bx_{j} \partial \bx_{k}} \xi_{i} \xi_{j} \xi_{k} r^3 \non
			&\quad
			+ \frac{1}{24} \sum_{1 \leq i,j,k, l \leq d} \frac{\partial^4 f(\bx + \theta r \bxi)}{\partial \bx_{i} \partial \bx_{j} \partial \bx_{k} \partial \bx_l } \xi_{i} \xi_{j} \xi_{k} \xi_l r^4
			\Big)\; \nu(\mathrm{d} \bxi)  ,
		\end{align}
		where $ 0 < \theta < 1 $. Note that the fourth-order partial derivatives of $f$ are bounded in some neiborghhood of $\bx$ by Condition \ref{cond2}, and 
		\begin{align}   \label{neweq.L8}
			\int_{\mathbb{S}^{d-1}}   \sum_{1 \leq i,j,k, l \leq d} |  \xi_{i} \xi_{j} \xi_{k} \xi_l |  \; \nu(\mathrm{d} \bxi)  
			& =  \int_{\mathbb{S}^{d-1}}  \Big( \sum_{i = 1}^d |  \xi_i  | \Big)^4  \; \nu(\mathrm{d} \bxi) \non
			& \leq   \int_{\mathbb{S}^{d-1}} d^2 \Big( \sum_{i = 1}^d  \xi_i^2 \Big)^2   \; \nu(\mathrm{d} \bxi) \non 
			& = d^2   \int_{\mathbb{S}^{d-1}}  \; \nu(\mathrm{d}  \bxi) = d^3 V_d. 
		\end{align}
		Thus, from \eqref{neweq.FL085}--\eqref{neweq.L7} and \eqref{neweq.L8} we can obtain 
		\begin{align}  \label{neweq.L9}
			\int_{\mathbb{S}^{d-1}} f(\bx + r  \bxi) \; \nu(\mathrm{d}  \bxi) & =   f(\bx)\, d \, V_d + \frac{1}{2} \, \tr(f''(\bx))  V_d \, r^2 +  R_1 (d, f , \bx)   \,  r ^4,
		\end{align}
		where the coefficient $ R_1 (d,  f, \bx ) $ in the remainder term is bounded and depends only on the fourth-order partial derivatives of $ f $ and  dimensionality $d$.

		For the numerator, it holds that 
		\begin{align} \label{neweq.FL090}
			& \int_{\mathbb{S}^{d-1}} \mu(\bx + r \, \bxi)f(\bx + r \, \bxi) \; \nu (\mathrm{d} \bxi) \non
			& =  \int_{\mathbb{S}^{d-1}} 
			\Big[\mu(\bx) + \mu'(\bx)^T\bxi \, r + \frac{1}{2} \, \bxi^T \mu''(\bx)\, \bxi \, r^2 \non
			&\quad+ \frac{1}{6} \sum_{1 \leq i,j,k \leq d} \frac{\partial^3 \mu(\bx)}{\partial \bx_{i} \partial \bx_{j} \partial \bx_{k}} \xi_{i} \xi_{j} \xi_{k} r^3
			+  \frac{1}{24} \sum_{1 \leq i,j,k,l \leq d} \frac{\partial^4 \mu(\bx + \theta_1 r \bxi )}{\partial \bx_{i} \partial \bx_{j} \partial \bx_{k} \partial \bx_{l}}  \xi_{i} \xi_{j} \xi_{k} \xi_l r^4 \Big] \non
			& \quad \times \Big[f(\bx) + f'(\bx)^T\bxi \, r + \frac{1}{2} \bxi^T f''(\bx) \, \bxi \, r^2  \non
			&\quad+ \frac{1}{6} \sum_{i,j,k} \frac{\partial^3 f(\bx)}{\partial \bx_{i} \partial \bx_{j} \partial \bx_{k}} \xi_{i} \xi_{j} \xi_{k} r^3
			+  \frac{1}{24} \sum_{1 \leq i,j,k,l \leq d} \frac{\partial^4 f (\bx + \theta_2 r \bxi )}{\partial \bx_{i} \partial \bx_{j} \partial \bx_{k} \partial \bx_{l}}  \xi_{i} \xi_{j} \xi_{k} \xi_l  r^4\Big]\; \nu (\mathrm{d} \bxi),
		\end{align}
		where $ 0 < \theta_1 < 1 $ and $ 0 < \theta_2 < 1 $.  In the same manner as deriving \eqref{neweq.L8}, we can bound the integrals associated with $r^4$ and the higher-orders $ r^5,r^6,  r^7$, and $ r^8 $ under Condition \ref{cond2} that the fourth-order partial derivatives of $f (\cdot)$ and $ \mu (\cdot)$ are bounded in a neighborhood of $\bx$. Hence, we can deduce that 
		\begin{align}  \label{neweq.L10}
			& \int_{\mathbb{S}^{d-1}} \mu(\bx + r \, \bxi)f(\bx + r \, \bxi) \; \nu (\mathrm{d} \bxi) \non
			& = \mu (\bx) f (\bx)  \int_{\mathbb{S}^{d-1}}   \; \nu (\mathrm{d} \bxi) + \frac { \mu (\bx) r^2  } {2}  \int_{\mathbb{S}^{d-1}} \bxi^T f'' (\bx) \bxi \; \nu (\mathrm{d} \bxi) \non
			& \quad  + r^2 \int_{\mathbb{S}^{d-1}} \bxi^T \mu' (\bx) f' (\bx) ^T \bxi \; \nu (\mathrm{d} \bxi) + \frac {f (\bx) r^2 } {2} \int_{\mathbb{S}^{d-1}} \bxi^T \mu'' (\bx) \bxi   \; \nu (\mathrm{d} \bxi)  \non
			& \quad + R_2 (d, f, \bx) r^4  + o (r^4) \non
			& =    \mu (\bx) f (\bx) d V_d +  \frac{1}{2} \, [f(\bx)\, \tr(\mu''(\bx)) + \mu(\bx)\, \, \tr(f''(\bx))] V_d \, r^2 \non
			& \quad + \mu'(\bx)^Tf'(\bx) V_d \, r^2 + R_2 (d, f, \bx) r^4 + o(r^4),
		\end{align}
		where the coefficient $ R_2 (d,  f, \bx ) $ in the remainder term is bounded and depends only on the fourth-order partial derivatives of $ f $ and  dimensionality $d$. The last equality in \eqref{neweq.L10} follows from \eqref{neweq.FL085}--\eqref{neweq.L7}. Therefore, substituting \eqref{neweq.L9} and \eqref{neweq.L10} into \eqref{eqn:mr-long-division} leads to 
		\begin{align*}
			m(r) 
			= \mu(\bx) + \frac{f(\bx)\tr(\mu''(\bx)) + 2 \, \mu'(\bx)^Tf'(\bx)}{2 \, d \, f(\bx)} \; r^2 + O_4 r^4
		\end{align*}
		as $ r \to 0 $, where $O_4 $ is a bounded quantity depending only on $d $ and the fourth-order partial derivatives of $ f (\cdot )$ and $ \mu (\cdot ) $. This completes the proof of Lemma \ref{lem2}. 
	}

	\subsection{Lemma \ref{new.lem.3} and its proof} \label{SecB.7}
	
	Lemma \ref{new.lem.3} below provides us with the order of the variance for the first-order  H{\'a}jek projection.  
	To simplify the technical presentation, we use $\bZ_i$ as a shorthand notation for $(\bX_i,Y_i)$. Given any fixed vector $\bx$, the projection of $\Phi(\bx; \bZ_1, \bZ_2, \ldots, \bZ_s)$ onto $\bZ_1$ is denoted as $\Phi_1(\bx; \bz_1)$ given by
	\begin{align} \label{neweq.FL012}
		\Phi_1(\bx; \bz_1) & = \mathbb{E} \, [\Phi(\bx; \bZ_1, \bZ_2, \ldots, \bZ_s) | \bZ_1 = \bz_1] \non
		&= \mathbb{E} \, [\Phi(\bx; \bz_1, \bZ_2, \ldots, \bZ_s)].
	\end{align}
	Denote by $\mathbb{E}_i$ and $\mathbb{E}_{i:s}$ the expectations with respect to $\bZ_i$ and $\{\bZ_i, \bZ_{i+1}, \ldots, \bZ_s\}$, respectively.
	
	
	\begin{lemma} \label{new.lem.3}
		For any fixed $ \bx $, the variance $\eta_1$ of $\Phi_1(\bx; \bZ_1)$ defined in (\ref{neweq.FL012}) satisfies that when $s \rightarrow \infty$ and $ s = o(n) $, 
		\begin{equation}
			\lim _{n \rightarrow \infty }\frac{\Var (\Phi)}{ n \eta_1 } = 0 .   \label{ratio1}
		\end{equation}
	\end{lemma}
	
	\noindent \textit{Proof}. A main ingredient of the proof is to decompose $ \Var (\Phi) $ and $ \eta_1 $ using the conditioning arguments. Denote by $ \zeta_{i, s} $  the indicator function for the event that  $ \bX_i $ is the 1NN of $ \bx $ among  $ \{\bX_1, \cdots, \bX_{s} \}$. 
	By symmetry, we can see that $ \zeta_{i, s} $ are identically distributed with mean 
	\[ \mathbb{E} \zeta_{i, s} = s^{-1}. \]
	In addition, observe that $ \Phi (\bx; \bZ_1, \bZ_2, \ldots, \bZ_s)  = \sum _{i =1 } ^{s } y_{i} \zeta_{i, s} $. Then we can obtain an upper bound of $ \Var \Phi $ as 
	\begin{align*}
		\Var (\Phi) & \leq \e [ \Phi^2 ] =  \e \Big[ \Big( \sum_{i = 1}^{s} y_i \zeta_{i, s} \Big)^2 \Big] = \sum_{i = 1} ^s \e [ y_i^2 \zeta_{i, s} ] \\
		&= s \e [ y_1^2 \zeta_{1, s} ],
	\end{align*}
	where we have used the fact that $\zeta_{i, s} \zeta_{j, s}  = 0 $ with probability one when $i \neq j$.
	
	Since $ \e [ \epsilon | {\bf X} ] = 0$ by assumption, it holds that 
	\begin{equation*}
		\begin{split}
			s \e [ y_1^2 \zeta_{1, s} ] & = s \e [ \mu^2 ({\bf X}_1) \zeta_{1, s} ] + \sigma_{\epsilon}^2 s \e [ \zeta_{1, s} ]  \\
			& =  \e_1 [ \mu^2 ({\bf X}_1) s\e_{2:s} [ \zeta_{1, s} ] ] + \sigma_{\epsilon}^2.
		\end{split}
	\end{equation*}
	A key observation is that $ \mathbb{E}_{2:s} [\zeta_{1, s}] = \{ 1 - \varphi(B(\bx, \| \bX_1 - \bx \|) )\}^{s-1} $ and $\mathbb{E}_1 [s \mathbb{E}_{2:s} [\zeta_{1, s}] ]  = 1 $. See Lemma \ref{lem:basic-prop-zeta} in Section \ref{sec:Supplemantary-Characteristic Function} for a list of properties for the indicator functions $ \zeta_{i, s} $. Thus, $s \mathbb{E}_{2:s} [\zeta_{1, s}] $ behaves like a Dirac measure at $ \bx $ as $ s \rightarrow \infty $. Such observation leads to Lemma \ref{lem:delta-simple} in Section \ref{SecC.3}, which entails that 
	\begin{equation} \label{neweq.FL091}
		\Var (\Phi) \leq  \mu^2 ({\bf x}) + \sigma_{\epsilon}^2 + o(1)
	\end{equation}
	as $s \to \infty$. 
	
	To derive a lower bound for $ \eta_1 $, we exploit the idea in Theorem 3 of \cite{PCM2019}. Let $ B $ be the event that $ {\bf X}_1 $ is the nearest neighbor of ${\bf x}$ among $ \{{\bf X}_1, \ldots, {\bf X}_s \}$. Denote by $\bX_1^*$ the nearest point to ${\bf x}$ and $y_1^*$ the corresponding response. Then we can deduce that 
	\begin{align*}
		\Phi_1 ({\bf x}; {\bf Z}_1)  & = \e [ y_1 \mathbbm{1}_B | {\bf Z}_1 ] + \e [ y_1^* \mathbbm{1}_{B^c} | {\bf Z}_1 ] \\
		& = y_1 \e [ \mathbbm{1}_B | {\bf Z}_1 ] +  \e [ y_1^* \mathbbm{1}_{B^c } | {\bf Z}_1 ]  \\
		& = \epsilon_1 \e [ \mathbbm{1}_B | {\bf X}_1 ] + \mu ({\bf X}_1 ) \e [ \mathbbm{1}_B | {\bf X}_1 ] + \e [ \mu ({\bf X}_{1}^*)  \mathbbm{1}_{B^c} | {\bf X}_1 ]  \\
		& = \epsilon_1 \e [ \mathbbm{1}_B | {\bf X}_1  ] + \e [ \mu( {\bf X}_1^*) | {\bf X}_1 ] .
	\end{align*}
	Since $ \epsilon $ is an independent model error term with $ \e [ \epsilon | {\bf X} ] = 0  $ by assumption, it holds that 
	\begin{align}
		\eta_1 & = \Var ( \Phi_1 ({\bf x}; {\bf Z}_1)  )  = \Var ( \epsilon_1 \e [ \mathbbm{1}_B | {\bf X}_1  ]  ) + \Var ( \e [ \mu( {\bf X}_1^*) | {\bf X}_1 ] ) \non
		& \geq \Var (  \epsilon_1 \e [ \mathbbm{1}_B | {\bf X}_1  ] ) = \sigma_{\epsilon}^2 \e \big[ \e^2 [ \mathbbm{1}_{B} | {\bf X}_1 ] \big] \non
		&= \frac {\sigma_{\epsilon}^2   } { 2 s - 1  }, \label{l-bdd}       
	\end{align}
	where we have used the fact that 
	\[ \e [ \e^2  [ \mathbbm{1}_B | {\bf X}_1 ] ]  = \e [ \mathbbm{1}_{B'} | {\bf X}_1 ] = \frac {1} {2 s - 1} \] 
	with $ B' $ representing the event that $ {\bf X}_1 $ is the nearest neighbor of ${\bf x}$ among the i.i.d. observations $ \{{\bf X}_1, {\bf X}_2, \ldots, {\bf X}_s, {\bf X}_2', \ldots, {\bf X}_{s}' \}$.
	
	We now turn to the upper bound for $\eta_1$. From the variance decomposition for $ \Var (\Phi) $ given in \eqref{decom-varphi}, we can obtain
	\begin{equation*}
		\begin{split}
			\Var (\Phi) & = \sum_{j = 1}^s {s \choose j } \Var ( g_j ( {\bf x} ; {\bf Z}_{1}, \ldots, {\bf Z}_j ) ) \\
			& = s \eta_1 + \sum_{j = 2} ^s {s \choose j } \Var ( g_j ( {\bf x} ; {\bf Z}_{1}, \ldots, {\bf Z}_j ) ),
		\end{split}     
	\end{equation*}
	which along with (\ref{neweq.FL091}) entails that 
	\begin{equation}
		s \eta_1   \leq  \Var (\Phi)  \leq \mu^2 ({\bf x}) + \sigma_{\epsilon}^2 + o(1). \label{u-bdd}
	\end{equation}
	Consequently, combining \eqref{l-bdd} and \eqref{u-bdd} leads to 
	\begin{equation} \label{neweq.FL017}
		\eta_1  \sim s^{-1},
	\end{equation}
	where $\sim$ denotes the asymptotic order. Finally, recall that it has been shown that $ \Var (\Phi) \leq C $ for some positive constant depending upon $ \mu (\bf x) $ and $ \sigma_{\epsilon} $. Therefore, we see that as long as $s \rightarrow \infty$ and $ s = o(n)$, 
	\[ \frac{\Var (\Phi)} {  n \eta_1  }   = O (\frac {s} {n} ) \to 0, \]
	which yields the desired conclusion in \eqref{ratio1}. This concludes the proof of Lemma \ref{new.lem.3}.

	\subsection{Lemma \ref{le-U-pre} and its proof} \label{Sec.B5}
	
	Assume that $s_1 < s_2$ for the two subsampling scales. Let us define 
	\begin{align} \label{neweq.FL092}
		\Phi^{(1)} ({\bf x}; {\bf Z}_{1}, {\bf Z}_2, \ldots, {\bf Z}_{s_2} ) = {s_2 \choose s_1}^{-1} \sum_{1 \leq i_1 < i_2 < \ldots <  i_{s_1} \leq s_2} \Phi ({\bf x}; {\bf Z}_{i_1}, {\bf Z}_{i_2}, \ldots, {\bf Z}_{i_{s_1}})      
	\end{align}
	and 
	\begin{align} \label{phistar}
		\Phi^{*} ( {\bf x}; {\bf Z}_1, {\bf Z}_2, \ldots,  {\bf Z}_{s_2} ) = w_1^*  \Phi^{(1)} ({\bf x}; {\bf Z}_{1}, {\bf Z}_2, \ldots, {\bf Z}_{s_2} ) +  w_2^* \Phi   ({\bf x}; {\bf Z}_{1}, {\bf Z}_2, \ldots, {\bf Z}_{s_2} ), 
	\end{align}
	where $ w_1^*  $ and $w_2^*$ are determined by the system of linear equations \eqref{wei1}--\eqref{wei2}.
	
	\begin{lemma} 
		\label{le-U-pre}
		The two-scale DNN estimator $D_n (s_1, s_2) ({\bf x})$ admits a U-statistic representation given by 
		\begin{align}
			D_n (s_1, s_2) ({\bf x}) = {n \choose s_2}^{-1} \sum_{1 \leq i_1 < i_2 < \ldots < i_{s_2} \leq n } \Phi^*({\bf x}; {\bf Z}_{1}, {\bf Z}_2, \ldots, {\bf Z}_{s_2} ),
		\end{align} 
		where the kernel function $\Phi^{*} ( {\bf x}; \cdot)$ is defined in (\ref{phistar}).
	\end{lemma}
	
	\noindent \textit{Proof}. From the definition of the two-scale DNN estimator $ D_n (s_1, s_2) ({\bf x}) $ introduced in \eqref{TDNN}, we have 
	\begin{align*}
		D_n (s_1, s_2) ({\bf x}) & = w_1^* {n \choose s_1}^{-1} \sum_{1 \leq i_1 < i_2 < \ldots < i_{s_1} \leq n } \Phi  ({\bf x}; {\bf Z}_{i_1}, \ldots, {\bf Z}_{i_{s_1}} ) \\
		& \quad + w_2^* {n \choose s_2}^{-1} \sum_{1 \leq i_1 < i_2 < \ldots <  i_{s_2} \leq n } \Phi  ({\bf x}; {\bf Z}_{i_1}, \ldots, {\bf Z}_{i_{s_2}} ).
	\end{align*}
	Thus, to establish the U-statistic representation for the two-scale DNN estimator $D_n (s_1, s_2) ({\bf x})$, it suffices to show that 
	\begin{align} \label{neweq.FL093}
		&  {n \choose s_1}^{-1}   \sum_{1 \leq i_1 < i_2 < \ldots <  i_{s_1} \leq n } \Phi  ({\bf x}; {\bf Z}_{i_1}, \ldots, {\bf Z}_{i_{s_1}} )  \non
		& = {n \choose {s_2}}^{-1} \sum_{1 \leq i_1 < i_2 < \ldots < i_{s_2} \leq n } \Phi^{(1)}({\bf x}; {\bf Z}_{1}, {\bf Z}_2, \ldots, {\bf Z}_{s_2} ) \non
		& = {n \choose s_2}^{-1} {s_2 \choose s_1}^{-1} \sum_{1 \leq i_1 < i_2 < \ldots < i_{s_2} \leq n } \sum_{1 \leq j_1 < j_2 < \ldots < j_{s_1} \leq s_2} \Phi ( {\bf x}; {\bf Z}_{i_{j_1}}, {\bf Z}_{i_{j_2}}, \ldots, {\bf Z}_{i_{j_{s_1}}} ).
	\end{align}
	
	Observe that for each given tuple $1 \leq u_1 < u_2 < \ldots < u_{s_1} \leq n$, it will appear a total of $ {n - s_1 \choose s_2 - s_1} $ times in the summation
	\begin{align*}
		\sum_{1 \leq i_1 < i_2 < \ldots < i_{s_2} \leq n } \sum_{1 \leq j_1 < j_2 < \ldots < j_{s_1} \leq s_2} \Phi ( {\bf x}; {\bf Z}_{i_{j_1}}, {\bf Z}_{i_{j_2}}, \ldots, {\bf Z}_{i_{j_{s_1}}} ).
	\end{align*} 
	Indeed, if $ ( i_{j_1}, i_{j_2}, \ldots,  i_{j_{s_1}} )  =  (u_1, u_2, \ldots, u_{s_1}) $ are fixed, then there exist $ {n - {s_1} \choose s_2 - s_1} $ options for the remaining $ s_2 - s_1 $ places in $ ( i_1, i_2, \ldots, i_{s_2} ) $. Consequently, it holds that 
	\begin{align*}
		&   {n \choose {s_2}}^{-1} {s_2 \choose s_1}^{-1} \sum_{1 \leq i_1 < i_2 < \ldots < i_{s_2} \leq n } \sum_{1 \leq j_1 < j_2 < \ldots < j_{s_1} \leq {s_2} } \Phi ( {\bf x}; {\bf Z}_{i_{j_1}}, {\bf Z}_{i_{j_2}}, \ldots, {\bf Z}_{i_{j_{s_1}}} ) \\
		& =  {n \choose s_2}^{-1} {s_2 \choose s_1}^{-1} {n - s_1 \choose s_2 - s_1}  \sum_{1 \leq i_1 < i_2 < \ldots <  i_{s_1} \leq n } \Phi  ({\bf x}; {\bf Z}_{i_1}, \ldots, {\bf Z}_{i_{s_1}} )  \\
		& = {n \choose s_1}^{-1}  \sum_{1 \leq i_1 < i_2 < \ldots <  i_{s_1} \leq n } \Phi  ({\bf x}; {\bf Z}_{i_1}, \ldots, {\bf Z}_{i_{s_1}} ),
	\end{align*}
	which establishes the desired claim in (\ref{neweq.FL093}). This completes the proof of Lemma \ref{le-U-pre}.

	\subsection{Lemma \ref{lemma-phi} and its proof} \label{SecB.9}
	
	We provide in Lemma \ref{lemma-phi} below the order of the variance of the kernel function $ \Phi^* $ defined in \eqref{phistar} for the two-scale DNN estimator $D_n (s_1, s_2) ({\bf x})$, which states that the variance of the kernel function is bounded from above by some positive constant depending upon the underlying distributions. Denote by $ \Var (\Phi^*) = \Var [ \Phi^* ( {\bf x}; {\bf Z}_1, \ldots, {\bf Z}_{s_2} ) ]  $ for simplicity. 
	
	\begin{lemma} \label{lemma-phi}
		Under the conditions of Theorem \ref{thm1}, there exists some positive constant $C$ depending upon $c_1$ and $c_2$ such that
		\begin{align}
			\Var (\Phi^*) \leq  C \big( \mu^2 ({\bf x})  + \sigma_{\epsilon}^2 + o(1) \big)  \label{phi-var}
		\end{align}
		as $s_1\to \infty$ and $ s_2 \to \infty$.
	\end{lemma}
	
	\noindent \textit{Proof}.  Since $ \Var ( \Phi^*  )  \leq \e [ (\Phi^* )^2 ] $, it suffices to bound $ \e [ ( \Phi^* )^2 ] $. It follows that 
	\begin{align} \label{neweq.FL094}
		\e [ ( \Phi^* )^2 ] & \leq 2 (w_1^*)^2 \e \big\{ [ \Phi^{(1)} ( {\bf x}; {\bf Z}_{1}, \ldots, {\bf Z}_{s_2} ) ]^2  \big\}\non
		& \quad+ 2 ( w_2^* )^2 \e [ \Phi^2 ({\bf x}; {\bf Z}_{1},  \ldots, {\bf Z}_{s_2}) ] \non
		& \leq 2 (w_1^*)^2 \e [ \Phi^2 ( {\bf x}; {\bf Z}_1,  \ldots, {\bf Z}_{s_1} ) ] \non
		& \quad+ 2 ( w_2^* )^2 \e [ \Phi^2 ({\bf x}; {\bf Z}_{1},  \ldots, {\bf Z}_{s_2}) ] ,
	\end{align} 
	where the last inequality holds since 
	\begin{equation*}
		\begin{split}
			& \e \big\{ [ \Phi^{(1)} ( {\bf x}; {\bf Z}_{1},   \ldots, {\bf Z}_{s_2} ) ]^2  \big\} \\
			& = {s_2 \choose s_1}^{-2}  \sum_{ \substack{1 \leq i_1 < \ldots < i_{s_1} \leq s_2 \\ 1 \leq j_1 < \ldots < j_{s_1} \leq s_2 } } \e \big\{ \Phi ( {\bf x}; {\bf Z}_{i_1}, \ldots, {\bf Z}_{i_{s_1}} )  \Phi ( {\bf x}; {\bf Z}_{j_1}, \ldots, {\bf Z}_{j_{s_1}} )  \big\} \\
			& \leq {s_2 \choose s_1}^{-2}  \sum_{ \substack{1 \leq i_1 < \ldots < i_{s_1} \leq s_2 \\ 1 \leq j_1 < \ldots < j_{s_1} \leq s_2 } }  \e [ \Phi^2 ( {\bf x}; {\bf Z}_{1}, \ldots, {\bf Z}_{s_1} ) ] \\
			& =  \e [ \Phi^2 ( {\bf x}; {\bf Z}_{1}, \ldots, {\bf Z}_{s_1} ) ] .
		\end{split}
	\end{equation*}
	
	Since $ \Phi ( {\bf x}; {\bf Z}_1, \ldots, {\bf Z}_{s_1} ) = \sum_{i = 1}^{s_1} y_i \zeta_{i, s_1} $ and $ \zeta_{i , s_1} \zeta_{j, s_1} = 0  $ with probability one when $ i \neq j $, we can deduce that 
	\begin{equation*}
		\begin{split}
			\e [ \Phi^2 ( {\bf x}; {\bf Z}_{1}, \ldots, {\bf Z}_{s_1} ) ]  & = \e \Big[ \Big( \sum_{i = 1}^{s_1} y_i \zeta_{i, s_1} \Big)^2 \Big] = \sum_{i= 1}^{s_1} \sum_{j = 1}^{s_1} y_i y_j \zeta_{i, s_1} \zeta_{j, s_1} \\
			& = \sum_{i = 1} ^{s_1} y_i^2 \zeta_{i, s_1} = s_1 \e [ y_1^2 \zeta_{1, s_1} ] \\
			& = s_1 \e [ \mu^2 ({\bf X}_1 ) \zeta_{1, s_1} ] +  \sigma_{\epsilon}^2  s_1 \e [ \zeta_{1, s_1} ]. 
		\end{split}      
	\end{equation*}
	Note that $ s_1 \e [ \zeta_{1, s_1} ]  = \sum_{i = 1}^n \zeta_{i, s_1} $. Furthermore, it follows from Lemma \ref{lem:delta-simple} in Section \ref{SecC.3} that 
	\[  s_1 \e [ \mu^2 ({\bf X}_1)  \zeta_{1, s_1}]  \to \mu^2 ({\bf x})  \] 
	as $ s_1 \to \infty $. Thus, we have that as $s_1 \to \infty $,
	\begin{equation} \label{neweq.FL095}
		\e [ \Phi ^2 (  {\bf x}; {\bf Z}_{1}, \ldots, {\bf Z}_{s_1} ) ] =  \mu^2 ({\bf x})  + \sigma_{\epsilon}^2 + o(1).
	\end{equation}
	Similarly, we can show that as $ s_2 \to \infty $, 
	\begin{equation} \label{neweq.FL096}
		\e   [ \Phi ^2 (  {\bf x}; {\bf Z}_{1}, \ldots, {\bf Z}_{s_2} ) ] = \mu^2 ({\bf x})  + \sigma_{\epsilon}^2 + o(1).
	\end{equation}
	
	Consequently, combining (\ref{neweq.FL094}), (\ref{neweq.FL095}), and (\ref{neweq.FL096}) results in 
	\begin{equation} \label{neweq.FL097}
		\e [ ( \Phi^* )^2 ] \leq  2  \big[( w_1^* )^2 + ( w_2^* )^2 \big]  \big[ \mu^2 ({\bf x})  + \sigma_{\epsilon}^2 + o(1) \big].
	\end{equation}
	Since $ c_1 \leq s_1 / s_2 \leq c_2 $ by assumption, it holds that 
	\[ (w_1^*)^2 \leq  C \ \text{ and } \ (w_2^*)^2 \leq C \]
	for some absolute positive constant $C$ depending upon $c_1$ and $c_2$, which together with (\ref{neweq.FL097}) entails the desired upper bound in \eqref{phi-var}. This concludes the proof of Lemma \ref{lemma-phi}.

	\subsection{Lemma \ref{lemma-eta} and its proof} \label{SecB.10}
	
	Lemma \ref{lemma-eta} below establishes the order of the variance for the first-order H\'ajek projection of the kernel function $\Phi^* $ defined in \eqref{phistar}. 
	Recall that in the proof of Theorem \ref{thm1} in Section \ref{SecA.3}, we have defined that for each $1 \leq i \leq s_2$, 
	\begin{equation*}
		\Phi^*_i ({\bf x}; {\bf z}_1, \ldots, {\bf z}_i )= \e [ \Phi ^*( {\bf x}; {\bf z}_1, \ldots, {\bf z}_{i}, {\bf Z}_{i + 1}, \ldots, {\bf Z}_{s_2} )  \, | \, {\bf z}_1, \ldots, {\bf z}_{i}],
	\end{equation*}
	\begin{align*}
		g_i^* ( {\bf z}_1, \ldots, {\bf z}_i ) & = \Phi_i^* ( {\bf x}; {\bf z}_1, \ldots, {\bf z}_i ) - \e \Phi^* ( {\bf x}; {\bf Z}_1, \ldots, {\bf Z}_i) \\
		&\quad- \sum_{j = 1}^{i - 1} \sum_{1 \leq \alpha_1 < \ldots < \alpha_{j} \leq i} g_j^* ( {\bf z}_{\alpha_1}, \ldots, {\bf z}_{\alpha_j} ),
	\end{align*}
	and $ \eta_1^* = \Var (\Phi_1^* ({\bf x}; {\bf Z}_1)) $. 
	
	\begin{lemma} \label{lemma-eta}
		Under the conditions of Theorem \ref{thm1}, it holds that
		\begin{align} \label{eta-var}
			\eta_1^*    \sim  s_2^{-1 } ,
		\end{align}
		where $\sim$ denotes the asymptotic order.
	\end{lemma}
	
	\noindent \textit{Proof}. We begin with the lower bound for $ \eta_1^* $. The proof follows the ideas used in the proof of Lemma \ref{new.lem.3} in Section \ref{SecB.7}. By definition, it holds that 
	\begin{align*}
		\Phi^*_1 ({\bf x}; {\bf Z}_1 ) & = w_1^* {s_2 \choose s_1}^{ - 1 } \sum_{1 \leq i_1 < \ldots < i_{s_1} \leq s_2} \e [  \Phi ( {\bf x}; {\bf Z}_{i_1}, \ldots, {\bf Z}_{i_{s_1}} ) | {\bf Z}_1 ] \\
		& \quad +  w_2^* \e [ \Phi ( {\bf x}; {\bf Z}_{i_1}, \ldots, {\bf Z}_{i_{s_2}} )  | {\bf Z}_1 ] \\
		& = w_1^* \frac { s_2 - s_1 } { s_2 } \e [ \Phi ( {\bf x}; {\bf Z}_1, \ldots, {\bf Z}_{s_1} )  ] + w_1^* \frac {s_1} { s_2 } \e [ \Phi ({\bf x}; {\bf Z}_1, \ldots, {\bf Z}_{s_1} ) | {\bf Z}_1 ] \\
		& \quad +  w_2^* \e [ \Phi ( {\bf x}; {\bf Z}_1, \ldots, {\bf Z}_{s_2}  ) | {\bf Z}_1 ].
	\end{align*}
	Since the first term on the right-hand side of the above equality is a constant, we have
	\begin{align*}
		\Var ( \Phi^*_1 ({\bf x}; {\bf Z}_1) ) 
		& = \Var \Big(  w_1^* \frac {s_1} { s_2 } \e [ \Phi ({\bf x}; {\bf Z}_1, \ldots, {\bf Z}_{s_1} ) | {\bf Z}_1 ] \\
		&\quad+  w_2^* \e [ \Phi ( {\bf x}; {\bf Z}_1, \ldots, {\bf Z}_{s_2}  ) | {\bf Z}_1 ] \Big).
	\end{align*}  
	
	Denote by $A_1$ the event that $ {\bf X}_1 $ is the nearest neighbor of ${\bf x}$ among $\{ {\bf X}_1, \ldots, {\bf X}_{s_1} \}$ and $A_2 $ the event that $ {\bf X}_1 $ is the nearest neighbor of ${\bf x} $ among $\{ {\bf X}_1, \ldots, {\bf X}_{s_2} \}$. Let $ {\bf X}_1^* $ be the nearest point to ${\bf x}$ among $ \{{\bf X}_1, \ldots, {\bf X}_{s_1} \}$ and $ y_1^* $ the corresponding value of the response. Similarly, we define $ \breve{\bf X}_1 $ as the nearest point to ${\bf x}$ among $ \{{\bf X}_1, \ldots, {\bf X}_{s_2} \}$ and $ \breve{y}_1 $ as the corresponding value of the response. Since $ \epsilon_i \indep {\bf X}_i $ and $ \e [ \epsilon_i ]  = 0$ by assumption, we can write 
	\begin{equation*}
		\begin{split}
			\e [ \Phi ( {\bf x}; {\bf Z}_1, \ldots, {\bf Z}_{s_1} ) | {\bf Z}_1 ] & =  \e [ y_1 \mathbbm{1}_{A_1} | {\bf Z}_1  ] + \e [ y_1^* \mathbbm{1}_{A_1^c} | {\bf Z}_1 ] \\
			&  = \epsilon_1 \e [ \mathbbm{1}_{A_1} | {\bf X}_1 ] + \e [ \mu ({\bf X}_1)  \mathbbm{1}_{A_1} | {\bf X}_1 ] +  \e [ \mu ({\bf X}_1^*) \mathbbm{1}_{A_1^c} | {\bf X}_1 ] \\
			& = \epsilon_1 \e [ \mathbbm{1}_{A_1} | {\bf X}_1 ] + \e [ \mu ({\bf X}_1^* ) | {\bf X}_1 ]  .
		\end{split}
	\end{equation*}
	Similarly, we can show that 
	\begin{equation*}
		\e [ \Phi ( {\bf x}; {\bf Z}_1, \ldots, {\bf Z}_{s_2} ) | {\bf Z}_1 ] = \epsilon_1 \e [ \mathbbm{1}_{A_2} | {\bf X}_1 ] + \e [ \mu ( \breve{\bf X}_1 ) | {\bf X}_1].
	\end{equation*}
	Thus, we can obtain
	\begin{align*}
		\Var ( \Phi_1^* ( {\bf x}; {\bf Z}_{1} ) )  
		& = \Var \Big\{ \epsilon_1 \Big( w_1^* \frac {s_1} {s_2}  \e [ \mathbbm{1}_{A_1} | {\bf X}_1 ] + w_2^* \e [ \mathbbm{1}_{A_2} | {\bf X}_1 ]   \Big) \\
		& \quad + w_1^* \frac {s_1} {s_2} \e [ \mu ({\bf X}_1^* ) | {\bf X}_1 ] + w_2^* \e [ \mu ( \breve{\bf X}_1 ) | {\bf X}_1 ] \Big\},
	\end{align*}
	which along with the assumption of $ \epsilon_1 \indep {\bf X}_1 $ and $ \e [ \epsilon_1   ]  = 0 $ yields 
	\begin{align*}
		\Var ( \Phi_1^* ( {\bf x}; {\bf Z}_{1} ) ) & =     \Var  \Big\{ \epsilon_1 \Big( w_1^* \frac {s_1} {s_2} \e [ \mathbbm{1}_{A_1} | {\bf X}_1  ] +  w_2^* \e [ \mathbbm{1}_{A_2} | {\bf X}_1 ]  \Big) \Big\} \\
		& \quad+ \Var \Big\{  w_1^* \frac {s_1} { s_2} \e [ \mu ({\bf X}_1^* ) | {\bf X}_1 ] + w_2^* \e [ \mu ( \breve{\bf X}_1 ) | {\bf X}_1 ]  \Big\} \\
		& \geq   \Var  \Big\{ \epsilon_1 \Big( w_1^* \frac {s_1} {s_2} \e [ \mathbbm{1}_{A_1} | {\bf X}_1  ] +  w_2^* \e [ \mathbbm{1}_{A_2} | {\bf X}_1 ]  \Big) \Big\}.
	\end{align*}
	
	Furthermore, we can deduce that 
	\begin{align*}
		& \Var  \Big\{ \epsilon_1 \Big( w_1^* \frac {s_1} {s_2} \e [ \mathbbm{1}_{A_1} | {\bf X}_1  ] +  w_2^* \e [ \mathbbm{1}_{A_2} | {\bf X}_1 ]  \Big) \Big\}  \\
		& = \sigma_{\epsilon}^2 \e \Big\{ \Big( w_1^* \frac {s_1} {s_2} \e [ \mathbbm{1}_{A_1} | {\bf X}_1 ] + w_2^* \e [ \mathbbm{1}_{A_2} | {\bf X}_1 ] \Big)^2 \Big\} \\
		& = \sigma_{\epsilon}^2 \Big\{ \Big( w_1^* \frac {s_1} {s_2} \Big)^2 \e \big[ \e^2 [ \mathbbm{1}_{A_1} | {\bf X}_1 ]  \big] +  2 w_1^* w_2^* \frac {s_1} {s_2} \e \big[ \e [ \mathbbm{1}_{A_1} | {\bf X}_1 ] \e  [ \mathbbm{1}_{A_2} | {\bf X}_1 ] \big] \\
		& \quad + (w_2^* )^2  \e \big[ \e^2 [ \mathbbm{1}_{A_2} | {\bf X}_1 ]  \big]
		\Big\}.
	\end{align*}
	Let us make use of the following basic facts
	\begin{align*}  
		& \e \big[ \e^2 [ \mathbbm{1}_{A_1} | {\bf X}_1 ]  \big]  = \frac {1} {2 s_1 - 1},  \\
		& \e \big[ \e [ \mathbbm{1}_{A_1} | {\bf X}_1 ] \e  [ \mathbbm{1}_{A_2} | {\bf X}_1 ] \big]  = \frac {1} {s_1 + s_2 - 1} ,  \\ 
		&  \e \big[ \e^2 [ \mathbbm{1}_{A_2} | {\bf X}_1 ]  \big] = \frac  { 1 } {2 s_2 - 1 }.
	\end{align*}
	Then it follows that
	\begin{align}
		\Var ( \Phi_1^* ( {\bf x}; {\bf Z}_1 ) ) 
		& \geq \sigma_{\epsilon}^2 \Big\{ \Big( w_1^* \frac {s_1} {s_2} \Big)^2 \frac {1} {2 s_1 -1  } + 2 w_1^* w_2^* \frac {s_1} {s_2}  \frac {1} {s_1 + s_2 - 1} \non
		&\quad+  (w_2^*)^2 \frac {1} {2 s_2 - 1}
		\Big\}.
		\label{lower-bdd}
	\end{align}	
	By (\ref{lower-bdd}) and the assumption of $ c_1 \leq s_1 / s_2 \leq c_2  $, we can obtain
	\begin{equation} \label{neweq.FL098}
		\Var ( \Phi_1^* ( {\bf x}; {\bf Z}_1 ) ) \geq C \sigma_{\epsilon}^2 s_2^{-1} 
	\end{equation}
	for some positive constant $ C  $ depending upon $c_1$ and $c_2$.
	
	We next proceed to show the upper bound for $ \Var ( \Phi_1^* ( {\bf x}; {\bf Z}_1 ) )  $. Since  
	\begin{align*}
		\Phi^* ({\bf x}; {\bf Z}_1, \ldots, {\bf Z}_{s_2} ) - \e \Phi^* ({\bf x}; {\bf Z}_1, \ldots, {\bf Z}_{s_2} ) = \sum_{j = 1}^{s_2} \sum_{1 \leq \alpha_1 < \ldots < \alpha_j \leq s_2} g_j^* ( {\bf Z}_{\alpha_1}, \ldots, {\bf Z}_{\alpha_j} ),
	\end{align*}
	we see that 
	\begin{align*}
		\Var ( \Phi^* ( {\bf x}; {\bf Z}_1, \ldots, {\bf Z}_{s_2} )  ) = \sum_{j = 1}^{s_2} {s_2 \choose j } \Var ( g_j^* ( {\bf Z }_{1}, \ldots, {\bf Z}_{j} ) ).
	\end{align*}
	Then it follows that 
	\begin{align*}
		\Var ( \Phi^* ( {\bf x}; {\bf Z}_1, \ldots, {\bf Z}_{s_2} )  ) \geq  s_2 \Var ( \Phi_1^* ( {\bf x}; {\bf Z}_1 ) ).
	\end{align*}
	Recall that it has been shown in Lemma \ref{lemma-phi} in Section \ref{SecB.9} that 
	\[ \Var ( \Phi^*  ( {\bf x}; {\bf Z}_1, \ldots, {\bf Z}_{s_2} )   ) \leq C, \] 
	where $ C  $ is some positive constant depending upon $ c_1$, $c_2 $, and the underlying distributions. Therefore, we can deduce that 
	\begin{equation}
		\Var ( \Phi_1^* ( {\bf x}; {\bf Z}_1 )  ) \leq C s_2^{ - 1 }, \label{var-ub}
	\end{equation}	
	which together with \eqref{neweq.FL098} entails the desired asymptotic order in \eqref{eta-var}. This completes the proof of Lemma \ref{lemma-eta}.

	\renewcommand{\thesubsection}{F.\arabic{subsection}}
	\section{Additional technical details} \label{SecC}
	
	\subsection{Lemma \ref{lemma:spherical-integration} and its proof} \label{sec:Supplemantary-SphericalIntegrationFormulas}
	We present in Lemma \ref{lemma:spherical-integration} below some useful spherical integration formulas.

	\begin{lemma} \label{lemma:spherical-integration}
		Let $\mathbb{S}^{d-1}$ be the unit sphere in $\mathbb{R}^d$, {\color{black} $\nu$ some measure constructed specifically on the unit sphere $\mathbb{S}^{d-1}$}, and $\bxi = (\xi_i) \in \mathbb{S}^{d-1}$ an arbitrary point on the unit sphere. 
		Then for any $d \times d$ symmetric matrix $A$, it holds that 
		\begin{eqnarray}
			\label{eqn:spherical-constant}	
			\int_{\mathbb{S}^{d-1}} \; \nu (\mathrm{d} \bxi) & = & d \, V_d,	\\
			\label{eqn:spherical-linear}
			\int_{\mathbb{S}^{d-1}} \bxi \; \nu(\mathrm{d} \bxi) & = & \bzero,	\\
			\label{eqn:spherical-quadratic}
			\int_{\mathbb{S}^{d-1}} \bxi^T A \, \bxi \; \nu (\mathrm{d} \bxi) & = & \tr (A) \, V_d,	\\
			\label{eqn:spherical-third-order}
			\int_{\mathbb{S}^{d-1}}   \xi_{i} \xi_{j} \xi_{k} \nu (\mathrm{d} \bxi) & = & 0  \quad \mbox{for any} ~ 1 \leq i, j, k \leq d, 
		\end{eqnarray}
		where $V_d = \frac{\pi^{d/2}}{\Gamma(1 + d/2)} $ denotes the volume of the unit ball in $\mathbb{R}^d$.
	\end{lemma}
	
	\noindent \textit{Proof}. It is easy to see that identities \eqref{eqn:spherical-linear} and \eqref{eqn:spherical-third-order} hold. This is because for each of them, the integrand is an odd function of variable $ \bxi $, which entails that the integral is zero.
	Identity \eqref{eqn:spherical-constant} can be derived using the iterated integral
	\begin{align*}
		V_d & = \int_0^1 \int _{\mathbb{S}^{d-1}}  \rho^{d-1} \nu( \mathrm{d} \bxi) \, \mathrm{d}\rho  
		= \left(\int_0^1 \rho^{d-1} \, \mathrm{d} \rho \right) \left(\int _{\mathbb{S}^{d-1}}   \nu( \mathrm{d} \bxi) \right) \\
		&= \frac{1}{d} \int _{\mathbb{S}^{d-1}}   \nu( \mathrm{d}\bxi).
	\end{align*}
	
	{\color{black}
		To prove \eqref{eqn:spherical-quadratic}, we first represent the integral in \eqref{eqn:spherical-quadratic} as a sum of integrals by expanding the quadratic expression in the integrand
		\begin{align}
			\int_{\mathbb{S}^{d-1}} \bxi^T A \, \bxi \; \nu (\mathrm{d}\bxi) & = \sum_{1 \leq i, j \leq d} A_{ij}  \int_{\mathbb{S}^{d-1}} \xi_i  \xi_j  \; \nu (\mathrm{d} \bxi).
		\end{align}
		For $i \neq j$, we have by symmetry that 
		\begin{align}
			\int_{\mathbb{S}^{d-1}} \xi_i  \xi_j  \; \nu (\mathrm{d} \bxi)  = \int_{\mathbb{S}^{d-1}}  - \xi_i  \xi_j  \; \nu (\mathrm{d} \bxi) =  0 .
		\end{align}
		Thus, it holds that 
		\begin{align}   \label{neweq.L11}
			\int_{\mathbb{S}^{d-1}} \bxi^T A \, \bxi \; \nu (\mathrm{d} \bxi) & = \sum_{i = 1}^d A_{ii} \int_{\mathbb{S}^{d-1}} \xi_i^2 \; \nu (\mathrm{d} \bxi) \non
			& = \tr (A) \int_{\mathbb{S}^{d-1}} \xi_1^2 \; \nu (\mathrm{d} \bxi) .
		\end{align}
		When $ d = 1 $, $ \mathbb{S}^{d-1} $ reduces to the trivial case of two points, $1$ and $-1$. Then we can obtain that for $d = 1$, 
		\begin{align} \label{neweq.L12}
			\int_{\mathbb{S}^{d-1}} \bxi^T A \, \bxi \; \nu (\mathrm{d} \bxi) & = 2 \tr (A) = \tr (A) V_d,  
		\end{align}
		where the last equality comes from the fact that $V_d = 2$ for $d = 1$.  
		When $ d \geq 2 $, we now use the spherical coordinates: $ \xi_1 = \cos(\phi_1) $, $ \xi_k = \cos(\phi_k) \prod_{i=1}^{k-1} \sin(\phi_i) $ for $ 1  \leq k \leq d-1 $, and $ \xi_d = \prod_{i=1}^{d - 1 } \sin(\phi_i) $, where  $ 0 \leq \phi_{d-1} < 2 \pi$ and  $ 0 \leq \phi_i  < \pi$ for $ 1 \leq i \leq d-2 $. Then the volume element becomes 
		\[ \nu (\mathrm{d} \bxi) = \left(\prod_{i=1}^{d-2} \sin^{d-1-i}(\phi_{i}) \right) \prod_{i=1}^{d} \, \mathrm{d}\phi_{i}. \] 
		It follows that 
		\begin{align}  \label{neweq.L13}
			\int_{\mathbb{S}^{d-1}} \xi_1^2 \; \nu (\mathrm{d} \bxi)  & =  \int_{ 0 }^{2 \pi} \int_{0}^{\pi} \cdots \int_{0}^{\pi}  \cos^2 (\phi_1 ) \left(\prod_{i=1}^{d-2} \sin^{d-1-i}(\phi_{i}) \right) \prod_{i=1}^{d} \, \mathrm{d} \phi_{i} \non
			&  = \frac {   \int_{0}^{\pi}  \cos^2 (\phi_1 )  \sin^{d - 2} (\phi_1)  \, \mathrm{d}\phi_{i}  }  {   \int_{0}^{\pi}     \sin^{d - 2} (\phi_1)  \, \mathrm{d}  \phi_{i}   }    \int_{ 0 }^{2 \pi} \int_{0}^{\pi} \cdots \int_{0}^{\pi}   \left(\prod_{i=1}^{d-2} \sin^{d-1-i}(\phi_{i}) \right) \prod_{i=1}^{d} \,  \mathrm{d}\phi_{i} \non
			& =  \frac {   \int_{0}^{\pi}  \cos^2 (\phi_1 )  \sin^{d - 2} (\phi_1)  \,  \mathrm{d} \phi_{i}  }  {   \int_{0}^{\pi}     \sin^{d - 2} (\phi_1)   \, \mathrm{d} \phi_{i}   }     \int_{\mathbb{S}^{d-1}}  \; \nu (\mathrm{d} \bxi)  \non
			& =  \frac {   \int_{0}^{\pi}  \cos^2 (\phi_1 )  \sin^{d - 2} (\phi_1)   \, \mathrm{d} \phi_{i}  }  {   \int_{0}^{\pi}     \sin^{d - 2} (\phi_1)  \, \mathrm{d} \phi_{i}   }  d V_d.
		\end{align}
		
		By applying the integration by parts twice to the numerator from the above expression, we can obtain 
		\begin{align*}
			\int_{0}^{\pi} \cos^{2}(\phi_{1}) \sin^{d-2}(\phi_{1})  \, \mathrm{d} \phi_{1} 
			& = \frac{1}{ d - 1 } \int_{0}^{\pi} \ \sin^{d}(\phi_{1})   \, \mathrm{d} \phi_{1}.
		\end{align*}
		In addition, using the trigonometric integration formulas, we can show that 
		\[ 
		\frac{\int_{0}^{\pi} \ \sin^{d}(\phi_{1})   \, \mathrm{d} \phi_{1}}{\int_{0}^{\pi} \ \sin^{d-2}(\phi_{1})   \, \mathrm{d}\phi_{1}} =  \frac { d - 1} {d},
		\] 
		which along with \eqref{neweq.L11} and \eqref{neweq.L13} leads to 
		\begin{align*}
			\int_{\mathbb{S}^{d-1}} \bxi^T A \, \bxi \; \nu (\mathrm{d} \bxi) & = \tr (A) V_d 
		\end{align*}
		for the case of $d = 2$. Also, it is easy to see that the same formula holds for the case of $d = 1$ by \eqref{neweq.L12}. This concludes the proof of Lemma \ref{lemma:spherical-integration}.
		
	}

	\subsection{Lemma \ref{lem:basic-prop-zeta} and its proof} \label{sec:Supplemantary-Characteristic Function}
	
	Let us define $ \zeta_{i, s} $ as the indicator function for the event that  $ \bX_i $ is the 1NN of $ \bx $ among  $ \{\bX_1, \cdots, \bX_{s} \}$. We provide in Lemma \ref{lem:basic-prop-zeta} below a list of properties for these indicator functions $ \zeta_{i, s} $.  
	
	\begin{lemma} \label{lem:basic-prop-zeta}
		The indicator functions $ \zeta_{i, s} $ satisfy that 
		\begin{enumerate}
			\item[1)] For any $ i \neq j $, we have $ \zeta_{i, s} \zeta_{j, s} = 0 $ with probability one;
			\item[2)] $\sum_{i=1}^{s} \zeta_{i, s} = 1 $;
			\item[3)] $\Expected [ \zeta_{i, s} ] = s^{-1} $;
			\item[4)] $\Expected_{2:s} [\zeta_{1, s}] = \{1 - \varphi(B(\bx, \|\bX_1 - \bx\|))\}^{s-1} $, where $\mathbb{E}_{i:s}$ denotes the expectation with respect to $\{\bZ_i, \bZ_{i+1}, \ldots, \bZ_s\}$.
		\end{enumerate}
	\end{lemma}
	
	The proof of Lemma \ref{lem:basic-prop-zeta} involves some standard calculations and thus we omit it here for simplicity. Let us make some remarks on $\Expected_{2:s} [\zeta_{1, s}] $ that can be regarded as a function of $ \bX_{1}$. The last property in Lemma \ref{lem:basic-prop-zeta} above shows that $\Expected_{2:s} [\zeta_{1, s}] $ vanishes asymptotically as $ s $ tends to infinity, unless $ \bX_1 $ is equal to $\bx$. Moreover, we see that 
	\[ \Expected _1 [ \Expected_{2:s} [\zeta_{1, s}]]  = s ^{-1}. \] 
	These two facts suggest that $\Expected_{2:s} [\zeta_{1, s}] $ tends to approximate the Dirac delta function at $ \bx $, which will be established formally in Lemma \ref{lem:delta-simple} in Section \ref{SecC.3}.

	\subsection{Lemma \ref{lem:delta-simple} and its proof} \label{SecC.3}
	
	\begin{lemma} \label{lem:delta-simple}
		For any $L^{1}$ function $ f $ that is continuous at $\bx$, it holds that 
		\begin{equation} \label{neweq.FL103}
			\lim_{s \rightarrow \infty} \Expected_1[ f(\bX_{1}) s \Expected_{2:s}[\zeta_{1, s}] ] = f(\bx).
		\end{equation}
	\end{lemma}
	
	\noindent \textit{Proof}. We will show that the absolute difference $ | \Expected_1[ f(\bX_{1}) s \Expected_{2:s}[\zeta_{1, s}] ] - f(\bx) | $ converges to zero as $s \rightarrow \infty$. By property 3) in Lemma \ref{lem:basic-prop-zeta} in Section \ref{sec:Supplemantary-Characteristic Function}, we have \[ \Expected _1 [ s \Expected_{2:s} [\zeta_{1, s}]]  = 1. \] Thus, we can deduce that 
	\begin{align} \label{neweq.FL102}
		| \Expected_1[ f(\bX_{1}) s \Expected_{2:s}[\zeta_{1, s}] ]  -  f(\bx) | 
		& = | \Expected_1[ \left( f(\bX_{1}) -  f(\bx) \right) s \Expected_{2:s}[\zeta_{1, s}] ]   |  \non
		& \leq \Expected_1[ | f(\bX_{1}) -  f(\bx) | s \Expected_{2:s}[\zeta_{1, s}] ] .
	\end{align}
	Let $ \epsilon >0 $ be arbitrarily given. By the continuity of function $ f $ at point $\bx$, there exists a neighborhood $ B(\bx, \delta) $ of $\bx$ with some $\delta > 0$ such that 
	\[ | f(\bX_1) - f(\bx) | < \epsilon \] 
	for all $ \bX_1 \in B(\bx, \delta) $. We will decompose the above expectation in (\ref{neweq.FL102}) into two parts: one inside and the other outside of $ B(\bx, \delta) $ as 
	\begin{align} \label{neweq.FL099}
		\Expected_1 & [ | f(\bX_{1}) -  f(\bx) | s \Expected_{2:s}[\zeta_{1, s}] ] 
		= \Expected_1[ | f(\bX_{1}) -  f(\bx) | s \Expected_{2:s}[\zeta_{1, s}] \Char_{B(\bx, \delta)}(\bX_1)]  \non
		& \quad + \Expected_1[ | f(\bX_{1}) -  f(\bx) | s \Expected_{2:s}[\zeta_{1, s}] \Char_{B^{c}(\bx, \delta)}(\bX_1)],
	\end{align}
	where the superscript $c$ stands for set complement in $\mathbb{R}^d$. 
	
	The first term on the right-hand side of (\ref{neweq.FL099}) is bounded by $ \epsilon$ since 
	\begin{align} \label{neweq.FL100}
		\Expected_1[ | f(\bX_{1}) -  f(\bx) | s \Expected_{2:s}[\zeta_{1, s}] \Char_{B(\bx, \delta)}(\bX_1)] 
		& \leq \Expected_1[ \epsilon s \Expected_{2:s}[\zeta_{1, s}] \Char_{B(\bx, \delta)}(\bX_1)] \non
		& \leq \Expected_1[ \epsilon s \Expected_{2:s}[\zeta_{1, s}] ] = \epsilon.
	\end{align}
	To bound the second term on the right-hand side of (\ref{neweq.FL099}), observe that 
	\[ B(\bx, \delta) \subset B(\bx, \|\bX_1 - \bx\|) \]
	when $ \bX_{1} \in B^{c}(\bx, \delta) $. Then an application of Lemma \ref{lem:basic-prop-zeta} gives 
	\[
	\Expected_{2:s}[\zeta_{1, s}] \leq (1 - \varphi ( B( \bx, \delta))) ^{s-1} \] 
	when $ \bX_{1} \in B^{c}(\bx, \delta) $. Thus, we can deduce that 
	\begin{align} \label{neweq.FL101}
		& \Expected_1[ | f(\bX_{1}) -  f(\bx) | s \Expected_{2:s}[\zeta_{1, s}] \Char_{B^{c}(\bx, \delta)}(\bX_1)] \non
		& \leq \Expected_1[ | f(\bX_{1}) -  f(\bx) | s (1 - \varphi ( B( \bx, \delta))) ^{s-1} \Char_{B^{c}(\bx, \delta)}(\bX_1)] \non
		& \leq s (1 - \varphi ( B( \bx, \delta))) ^{s-1} \Expected_1[ | f(\bX_{1}) -  f(\bx) |] \non
		& \leq s (1 - \varphi ( B( \bx, \delta))) ^{s-1} \left( \| f \|_{L^{1}} +   f(\bx) \right),
	\end{align}
	where $\| \cdot \|_{L^{1}}$ denotes the $L^{1}$-norm of a given function.
	
	Finally, we see that the right-hand side of the last equation in (\ref{neweq.FL101}) tends to 0 as $ s \rightarrow \infty $. Therefore, for large enough $ s $, the quantity
	\[ \Expected_1[ | f(\bX_{1}) -  f(\bx) | s \Expected_{2:s}[\zeta_{1, s}] \Char_{B^c(\bx, \delta)}(\bX_1)] \] 
	can be bounded from above by $ 2 \epsilon $. Since the choice of $ \epsilon >0 $ is arbitrary, combining such upper bound, (\ref{neweq.FL102}), (\ref{neweq.FL099}), and (\ref{neweq.FL100}) yields the desired limit in (\ref{neweq.FL103}) as $s \rightarrow \infty$. This completes the proof of Lemma \ref{lem:delta-simple}.

\end{document}